\documentclass[nodate]{lucid-twocol}

\usepackage[T1]{fontenc}
\usepackage{graphicx}
\usepackage{amsmath}
\usepackage{amssymb}
\usepackage{booktabs}
\usepackage{multirow}
\usepackage{tabularx}
\usepackage{xspace}
\usepackage{caption}
\usepackage{pifont}
\usepackage{colortbl}
\definecolor{oursrow}{HTML}{EEF3F9}
\definecolor{sdgray}{gray}{0.42}
\definecolor{gainblue}{HTML}{6F9BD8}\definecolor{gainbluedark}{HTML}{2F5FA8}
\definecolor{gaingray}{HTML}{C3C9D1}\definecolor{gaingraydark}{HTML}{6B7480}
\newcommand{\cmark}{\ding{51}}
\newcommand{\nmark}{--}
\newcommand{\synsar}{SynSAR\xspace}
\newcommand{\Sfive}{S@5\,px\xspace}
\newcommand{\Sten}{S@10\,px\xspace}
\newcommand{\miou}{mIoU$_8$\xspace}
\newcommand{\hd}[1]{\begin{tabular}[b]{@{}c@{}}#1\end{tabular}}

\definecolor{cvprblue}{rgb}{0.21,0.49,0.74}
\definecolor{chrxcolor}{RGB}{1, 140, 116}

\usepackage[pagebackref,breaklinks,colorlinks,allcolors=cvprblue]{hyperref}

\title[SynSAR: A Scene-Scale Physics-Based SAR Simulator and Dataset for Remote Sensing]%
  {SynSAR: A Scene-Scale Physics-Based SAR Simulator\\and Dataset for Remote Sensing}

\author[1,2]{Jian Song}\email{jian.song@riken.jp}
\author[1,2]{Hongruixuan Chen}
\author[1]{Hao Zheng}
\author[1,2]{Pengyu Dai}
\author[1]{Yafei Ou}
\author[2]{Yuzeng Chen}
\author[3]{Songxiao Yang}
\author[2]{Junjue Wang}
\author[1]{Junshi Xia}
\author[2]{Ryo Natsuaki}
\author[1,2]{Naoto Yokoya}\email{yokoya@k.u-tokyo.ac.jp}

\affil[1]{RIKEN}
\affil[2]{The University of Tokyo}
\affil[3]{Institute of Science Tokyo}

\logo*{\includegraphics[height=8.4mm]{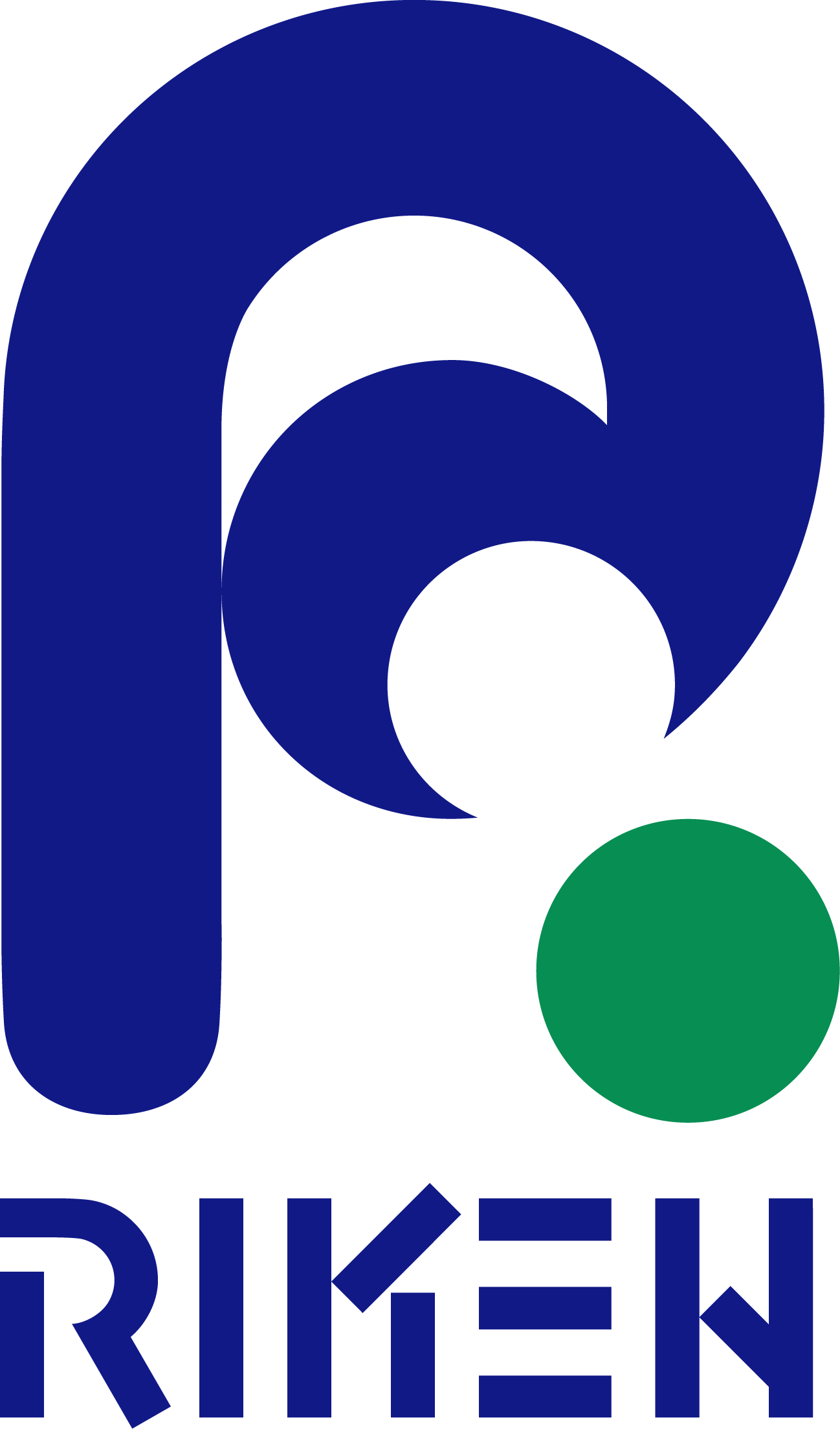}}
\logo*{\includegraphics[height=6.2mm]{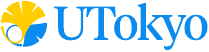}}

\makeatletter
\newcommand\SuppTocStart{}
\newcount\preprint@seccount
\newtoks\preprint@lefttoks
\newtoks\preprint@righttoks
\def\preprint@split{5}
\def\preprint@sectionword{section}
\def\preprint@collect#1#2#3#4{%
  \def\preprint@kind{#1}%
  \ifx\preprint@kind\preprint@sectionword \global\advance\preprint@seccount by 1 \fi
  \ifnum\preprint@seccount<\preprint@split\relax
    \global\preprint@lefttoks\expandafter{\the\preprint@lefttoks\preprint@contentsline{#1}{#2}{#3}{#4}}%
  \else
    \global\preprint@righttoks\expandafter{\the\preprint@righttoks\preprint@contentsline{#1}{#2}{#3}{#4}}%
  \fi}
\newcommand*\preprint@lsection[2]{%
  \ifnum \c@tocdepth >\z@
    \addpenalty\@secpenalty
    \addvspace{1.0em \@plus\p@}%
    \setlength\@tempdima{1.5em}%
    \begingroup
      \parindent \z@ \rightskip \@pnumwidth \@plus 3em
      \parfillskip -\@pnumwidth
      \hyphenpenalty\@M \exhyphenpenalty\@M
      \leavevmode \bfseries
      \advance\leftskip\@tempdima
      \hskip -\leftskip
      #1\nobreak\hfil
      \nobreak\hb@xt@\@pnumwidth{\hss #2\kern-\p@\kern\p@}\par
    \endgroup
  \fi}
\newlength\preprintsuppskip 
\newlength\preprintsuppafter 
\newcommand\preprintsupphead{%
  \newpage
  \twocolumn[{%
    \supplementhead
    \vskip \preprintsuppskip
    \edef\preprint@tocwd{\the\dimexpr\columnwidth-8pt\relax}%
    \begin{labpanel}[Contents]%
      \normalsize\normalfont\raggedright
      \let\l@section\preprint@lsection
      \setcounter{tocdepth}{2}%
      \renewcommand\@pnumwidth{2.2em}%
      \renewcommand\@tocrmarg{\dimexpr 2.55em-8pt\relax plus 1fil}%
      \renewcommand\@dotsep{6}%
      \hypersetup{linktoc=page}%
      \patchcmd\@dottedtocline{\leaders\hbox}{\color{labmuted!55!white}\leaders\hbox}{}{}%
      \patchcmd\@dottedtocline{\hfill \nobreak}{\hfill \normalcolor\nobreak}{}{}%
      \let\preprint@contentsline\contentsline
      \global\preprint@seccount=0 \global\preprint@lefttoks{}\global\preprint@righttoks{}%
      \def\contentsline##1##2##3##4{}%
      \def\SuppTocStart{\let\contentsline\preprint@collect}%
      \@starttoc{toc}%
      \noindent
      \begin{minipage}[t]{\preprint@tocwd}\the\preprint@lefttoks\end{minipage}\hfill
      \begin{minipage}[t]{\preprint@tocwd}\the\preprint@righttoks\end{minipage}\par
    \end{labpanel}%
    \vspace{\preprintsuppafter}%
  }]}

\newcount\preprint@head \newcount\preprint@par \newcount\preprint@total \newcount\preprint@enl
\newcount\preprint@fhd \newcount\preprint@tol \newdimen\preprint@ems
\newif\ifpreprint@marks \newif\ifpreprint@loosened \newif\ifpreprint@main \newif\ifpreprint@stale \newif\ifpreprint@wasstale
\newbox\preprint@box
\newcommand\preprintparmarks{\preprint@markstrue}
\newcommand\preprintloose[2][1000]{\@namedef{preprint@loose@#2}{#1}}
\newcommand\preprintat[2]{\@namedef{preprint@at@#1}{#2}}
\newcommand\preprintfitfor[2]{\def\preprint@fitN{#1}\def\preprint@fitH{#2}}
\newcommand\preprintenlarge[2]{\@namedef{preprint@enl@#1}{#2}}
\let\preprintpageheight\enlargethispage
\renewcommand\enlargethispage[1]{\global\advance\preprint@enl\@ne
  \@ifundefined{preprint@enl@\the\preprint@enl}{\preprintpageheight{#1}}%
    {\expandafter\ifx\csname preprint@enl@\the\preprint@enl\endcsname\@empty\else
       \preprintpageheight{\@nameuse{preprint@enl@\the\preprint@enl}}\fi}}
\def\preprint@name{\the\preprint@head.\the\preprint@par}
\gdef\preprint@thisname{0.0}
\gdef\preprint@collist{}
\def\preprint@bib{thebibliography}
\let\preprint@logname\@gobble
\newcommand\preprintfitstale{\global\preprint@staletrue \global\preprint@wasstaletrue}
\@ifl@t@r\fmtversion{2021/06/01}{%
  \AddToHook{cmd/@startsection/before}{\global\advance\preprint@head\@ne \global\preprint@par\z@}
  \AddToHook{para/before}{\ifinner \global\preprint@mainfalse \else \global\preprint@maintrue \fi}
  \AddToHook{para/begin}{%
    \@ifundefined{@captype}{%
      \global\advance\preprint@par\@ne
      \ifpreprint@main \xdef\preprint@thisname{\preprint@name}\fi
      \preprint@logname{\preprint@name}%
      \ifpreprint@marks \rlap{\normalfont\tiny\color{red}[[\preprint@name]]}\fi
      \ifpreprint@stale\else
        \@ifundefined{preprint@loose@\preprint@name}{}%
          {\looseness=\@ne
           \global\preprint@loosenedtrue \preprint@fhd\finalhyphendemerits \preprint@tol\tolerance \preprint@ems\emergencystretch
           \finalhyphendemerits=1000000 \tolerance=\@nameuse{preprint@loose@\preprint@name}\relax \emergencystretch\z@}%
        \@ifundefined{preprint@at@\preprint@name}{}{\@nameuse{preprint@at@\preprint@name}}%
      \fi
    }{\preprint@logname{-}}}
  \AddToHook{para/after}{%
    \ifpreprint@loosened \global\preprint@loosenedfalse
      \finalhyphendemerits\preprint@fhd \tolerance\preprint@tol \emergencystretch\preprint@ems \fi
    \ifinner\else \@ifundefined{@captype}{%
      \global\advance\preprint@total\@ne
      \ifx\@currenvir\preprint@bib\else
        \xdef\preprint@collist{\preprint@collist\space\preprint@thisname:\the\prevgraf}\fi}{}\fi}
  \AtEndDocument{%
    \@ifundefined{preprint@fitN}{}{%
      \@tempswafalse
      \ifnum\preprint@total=\preprint@fitN\relax\else \@tempswatrue \fi
      \ifnum\preprint@head=\preprint@fitH\relax\else \@tempswatrue \fi
      \if@tempswa
        \immediate\write\@auxout{\string\providecommand\string\preprintfitstale{}\string\preprintfitstale}%
        \ifpreprint@wasstale
          \@latex@warning@no@line{preprint_fit.tex was made for another text (now \the\preprint@total\space paragraphs,
            \the\preprint@head\space headings) and is not applied: run scripts/preprint_fit.py}%
        \else \@latex@warning@no@line{The fit list does not belong to this text. Rerun to get the column feet without it}\fi
      \else
        \ifpreprint@wasstale \@latex@warning@no@line{The fit list belongs to this text again. Rerun to get the column feet right}\fi
      \fi}}
}{}
\@ifundefined{preprintfitrecords}{}{%
  \newwrite\preprint@cols \immediate\openout\preprint@cols=\jobname.cols
  \def\preprint@logname#1{\immediate\write\m@ne{PPFIT par #1}}%
  \AtBeginDocument{\hbadness=100 }%
  \let\preprint@makecol\@makecol
  \def\@makecol{%
    \setbox\preprint@box\vbox{\unvcopy\@cclv}%
    \dimen@\vsize \advance\dimen@-\ht\preprint@box
    \immediate\write\preprint@cols{col abs=\the\numexpr\ReadonlyShipoutCounter+1\relax\space
      first=\if@firstcolumn 1\else 0\fi\space gap=\the\dimen@\space pen=\the\outputpenalty\space
      special=\ifvbox\@kludgeins 1\else 0\fi\space two=\if@twocolumn 1\else 0\fi\space pars=\preprint@collist}%
    \gdef\preprint@collist{}%
    \preprint@makecol}
  \AtEndDocument{\immediate\write\preprint@cols{end paragraphs=\the\preprint@total\space headings=\the\preprint@head}}}
\makeatother
\preprintfitfor{236}{93}
\preprintloose[5000]{10.1}
\preprintloose[5000]{14.2}
\preprintloose[5000]{13.1}
\preprintloose[5000]{28.1}
\preprintloose[5000]{27.1}
\preprintloose[5000]{34.2}
\preprintloose[5000]{75.1}

\preprintat{91.2}{\ifnum\value{page}=36 \preprintpageheight{12pt}\fi}
\preprintat{92.3}{\ifnum\value{page}=36 \preprintpageheight{12pt}\fi}
\preprintat{93.2}{\looseness=-1 }

\begin{document}
\twocolumn[{%
\renewcommand\twocolumn[1][]{#1}%
\maketitle
\vskip -22pt\vskip 11pt\relax
\begin{center}
\includegraphics[width=\textwidth]{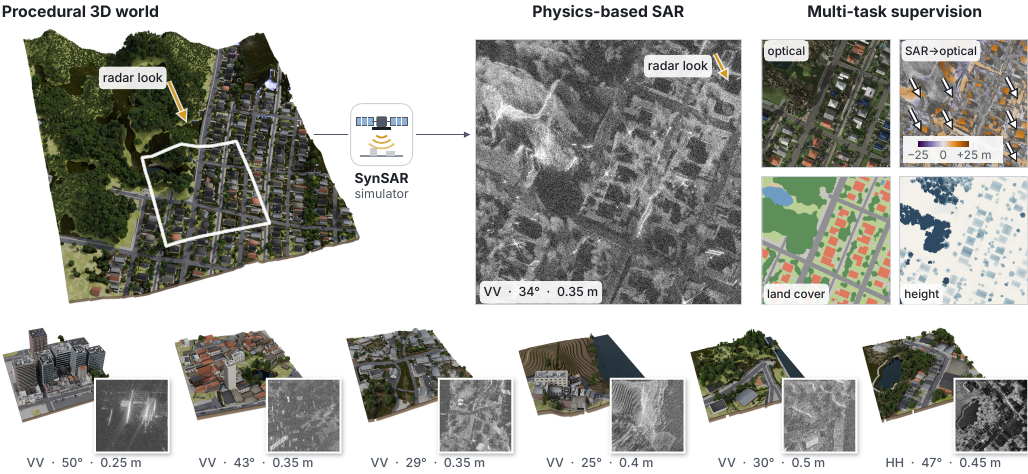}
\captionof{figure}{\textbf{SynSAR at a glance.}
A procedural 3D world (left) is rendered into a physics-based SAR image over the outlined region (center; VV, $34^\circ$ incidence, 0.35\,m resolution).
The same world provides a paired optical image from another date, land cover labels, height, and dense SAR--optical correspondences with validity filtering (right).
Bottom: additional 3D worlds and their simulated SAR observations under different acquisition settings.}
\label{fig:hero}
\end{center}
\vskip -3.5pt\hrule height 0pt\relax
\begin{abstract}
Synthetic aperture radar (SAR) images the Earth by day and night and through clouds, but very-high-resolution (VHR) SAR analysis remains limited by scarce reliable supervision.
Synthetic data offer a remedy, but SAR appearance is determined by 3D structure, acquisition geometry, and microwave scattering: optical-to-SAR translation typically models none of these explicitly, and many physics-based simulators focus on isolated targets or predefined scenes.
We present SynSAR, a scene-scale physics-based SAR simulator for generating deep-learning training data.
It builds procedural 3D worlds, square synthetic scenes of terrain, buildings, roads, and vegetation, and forms X-band SAR images through GPU ray tracing, explicit coherent multi-bounce returns, statistical sub-resolution scattering, and band-limited image formation.
It agrees with analytic and statistical references for layover, radar cross sections, and speckle.
Each world provides paired optical imagery, land cover maps, height, and validity-aware dense SAR--optical correspondences.
Our dataset comprises 11,847 worlds of $154\times154$\,m for land cover mapping and height estimation and 198 of $512\times512$\,m for registration, with typical resolutions of 0.25--1.4\,m.
Trained only on SynSAR and tested on real benchmarks, five SAR--optical matchers outperform their counterparts trained on co-registered real pairs by at least 3.5 points in success rate within 10 pixels, and height models outperform cross-city real-trained baselines by 0.74--1.32\,m in MAE.
Adding SynSAR to limited real supervision improves all three tasks.
Controlled experiments reveal how imaging mechanisms affect each task, and digital twins show that SynSAR, unlike a translation baseline, follows the real acquisition geometry.
We will release the synthetic dataset and the training and evaluation code.
\end{abstract}
\vspace{-16pt}%
}]
\section{Introduction}
\label{sec:intro}
\begin{table*}[t]
\caption{\textbf{Comparison of SAR simulation frameworks and learning-ready outputs.}
Model: RT, ray tracing; SBR, shooting-and-bouncing rays; PO, physical optics; SE, semi-empirical scattering; VRT, vector radiative transfer; IEM, integral-equation model; img/raw, image-/raw-signal formation.
Public: released synthetic dataset (SynSAR: planned).
Opt./LC/Hgt./Corr.: optical imagery, land cover, height, and correspondences paired with the simulated SAR images.
\cmark: explicitly reported; \nmark: not reported.}
\label{tab:related}
\centering\footnotesize
\setlength{\tabcolsep}{3.7pt}
\renewcommand{\arraystretch}{1.0}
\begin{tabular}{@{}l lll l ccccc@{}}
\toprule
System & Content & Source & Scale & Model & Public & Opt. & LC & Hgt. & Corr. \\
\midrule
SAMPLE~\cite{lewis2019sample} & targets & CAD & 10 targets $\times$ views & RT, raw & \cmark & \nmark & \nmark & \nmark & \nmark \\
SARCASTIC v2~\cite{woollard2022sarcastic} & targets & CAD & 1 target $\times$ views & RT+PO, raw & \cmark & \nmark & \nmark & \nmark & \nmark \\
Salsa~\cite{camus2026salsa} & targets & CAD & 10 targets $\times$ views & SBR+PO, img & \nmark & \nmark & \nmark & \nmark & \nmark \\
MPA~\cite{xu2006mpa} & heterogeneous & hand-built & 1 scene & VRT+IEM, img & \nmark & \nmark & \nmark & \nmark & \nmark \\
FERMAT / OKTAL~\cite{mametsa2002imaging,latger2005fermat} & heterogeneous & geodata & few sites & SBR+PO, img & \nmark & \cmark & \nmark & \nmark & \nmark \\
RaySAR~\cite{auer2010raytracing,auer2016raysar} & buildings & CAD, DSM & single objects & RT, img & \nmark & \nmark & \nmark & \nmark & \nmark \\
SimGeoI~\cite{auer2017simgeoi} & urban & DSM & 2 cities & RT, geometry only & \nmark & \nmark & \nmark & \nmark & \nmark \\
RT-SBRAS~\cite{jiao2024sarbusim} & buildings & CAD & 8 buildings $\times$ views & RT, raw & \cmark & \nmark & \nmark & \nmark & \nmark \\
\addlinespace[3pt]
\multicolumn{1}{@{}>{\columncolor{oursrow}[0pt][\tabcolsep]}l}{\textbf{SynSAR (ours)}} & \cellcolor{oursrow}heterogeneous & \cellcolor{oursrow}procedural & \cellcolor{oursrow}\textbf{12k worlds} & \cellcolor{oursrow}SBR+PO+SE, img & \cellcolor{oursrow}\cmark & \cellcolor{oursrow}\cmark & \cellcolor{oursrow}\cmark & \cellcolor{oursrow}\cmark & \multicolumn{1}{>{\columncolor{oursrow}[\tabcolsep][0pt]}c@{}}{\cmark} \\
\bottomrule
\end{tabular}
\end{table*}

Synthetic aperture radar (SAR) observes the Earth regardless of daylight and cloud cover, yet deep learning on very-high-resolution (VHR) SAR remains limited by supervision.
SAR measures range from the side, so elevated structures are displaced toward the sensor and followed by shadow: at $35^\circ$ incidence, a 30\,m roof is imaged about 43\,m from its footprint.
Labels drawn on maps or optical images thus do not align with the radar image at such structures, and SAR is hard to annotate by eye.
Existing benchmarks therefore derive labels from optical imagery, laser scanning, or image-level co-registration~\cite{chen2025bright,hansch2024spacenet9,shermeyer2020spacenet6,xia2025oemsar}, each covering a limited range of sites and acquisition conditions.

Synthetic data are a natural remedy~\cite{kong2020synthinel,raistrick2023infinigen,song2024synrs3d,song2024syntheworld}, but SAR cannot be synthesized the way optical images are: a SAR pixel records the coherent sum of all returns that share the same range and Doppler, so where a roof appears, which facade is bright, and what falls in shadow are fixed by 3D structure and acquisition geometry.
Generative models can synthesize convincing SAR-like imagery from optical images, semantic maps, or other inputs~\cite{debuysere2025spaceborne,jakubik2025terramind,liu2026huiyanearth,zhang2025phigan}, but they typically observe neither the 3D scene nor the acquisition and are themselves trained on real SAR.
Physics-based simulators derive radar effects from known geometry~\cite{auer2010raytracing,auer2016raysar,camus2026salsa,xu2006mpa}, yet many focus on isolated targets or predefined scenes.
Whether physics-based simulation can scale to a source of training data, and stay faithful enough to transfer to real SAR, has remained open.

We present \textbf{SynSAR}, a scene-scale physics-based SAR simulator that simulates image formation rather than imitating appearance, at dataset scale (\cref{fig:hero}).
Three design choices make this tractable:
(i)~each scene is a procedural 3D world of 154\,m or 512\,m that carries both the geometry and the physical properties governing X-band scattering;
(ii)~scattering is hybrid: GPU ray tracing computes coherent multi-bounce returns from resolved structures, while sub-resolution roughness and foliage enter through semi-empirical models and are sampled as speckle;
(iii)~images are formed directly in the image domain and band-limited by the sensor response, without synthesizing raw echoes.
Since all outputs derive from one 3D world, every SAR image comes with a paired optical image, land cover, height, and dense SAR--optical correspondences that mark where layover and shadow invalidate a match.

Our main contributions are:
\begin{itemize}
\item \textbf{Simulator and dataset.} A scalable physics-based SAR simulator and 11,847 small and 198 large worlds with paired optical imagery, land cover, height, and dense correspondences; to our knowledge, the first large-scale physics-based VHR SAR dataset supporting registration, land cover mapping, and height estimation.
\item \textbf{Validation.} SynSAR agrees with analytic and statistical references for geometry, radar cross sections, and speckle, and its digital-twin renderings correlate with real acquisitions only under the matching acquisition geometry, unlike a learned translation baseline.
\item \textbf{Transfer and analysis.} Models trained on SynSAR transfer to real SAR across the three tasks, alone or with limited real supervision, and controlled experiments reveal how imaging mechanisms affect each task.
\end{itemize}

\section{Related work}
\label{sec:related}

\paragraph{Learning-based SAR generation.}
GANs and diffusion models synthesize SAR images from optical imagery, semantic maps, and other inputs~\cite{debuysere2025spaceborne,jakubik2025terramind,liu2026huiyanearth,zhang2025phigan}, and recent work adds 3D guidance for finer control~\cite{wu2026geodiffsar}.
These models learn an appearance prior from real SAR, which does not guarantee that layover, shadow, and scattering agree with the scene geometry.
We compare with such a pipeline in \cref{sec:translation}.

\paragraph{Physics-based SAR simulation.}
Forward SAR simulation predicts radar returns with shooting-and-bouncing rays, physical optics, and surface-scattering models~\cite{bhalla1995image,ling1989sbr,oh2004retrieval}, while neural and differentiable renderers address the inverse problem of recovering a scene from radar data~\cite{fu2022dsr,kung2025radarsplat,lei2024sarnerf,li2025sargs}.
Target-oriented simulators render CAD targets from many views for recognition~\cite{camus2025combining,camus2026salsa,lewis2019sample,woollard2022sarcastic}.
Scene-oriented simulators such as RaySAR model buildings and terrain~\cite{auer2010raytracing,latger2005fermat,mametsa2002imaging,xu2006mpa}, but each scene is built by hand or derived from site-specific geodata, which limits them to a few objects or sites and to partial supervision (\cref{tab:related}).
SynSAR removes this dependence: it generates worlds procedurally, with the physical properties that scattering depends on, and renders more than ten thousand of them with paired optical imagery, land cover, height, and correspondences.

\paragraph{Synthetic data and geometric supervision.}
Synthetic data provide optical vision with supervision that is costly to annotate by hand~\cite{raistrick2023infinigen}, including land cover and height in remote sensing~\cite{kong2020synthinel,song2024synrs3d,song2024syntheworld}.
For SAR, they have mainly served target recognition~\cite{camus2025combining,kusk2016synthetic,lewis2019sample,malmgren2017transfer}.
Scene-level SAR supervision must respect the side-looking geometry: DSM-based methods derive SAR--optical correspondences and joint visibility for existing sites~\cite{auer2017simgeoi,hughes2018joint}, simulated datasets and environments provide building annotations, scattering labels, or paired observations~\cite{chen2022cvcmff,guo2025omnisar,jiao2024sarbusim}, and cross-modal matchers depend on such correspondence supervision~\cite{edstedt2024roma,he2025matchanything,ren2025minima,tuzcuoglu2024xoftr,zhang2022osmnet}.
SynSAR generates SAR, optical imagery, land cover, height, and validity-aware correspondences from the same procedural worlds.

\section{SynSAR}
\label{sec:method}

\begin{figure*}[t]
\centering
\includegraphics[width=\textwidth]{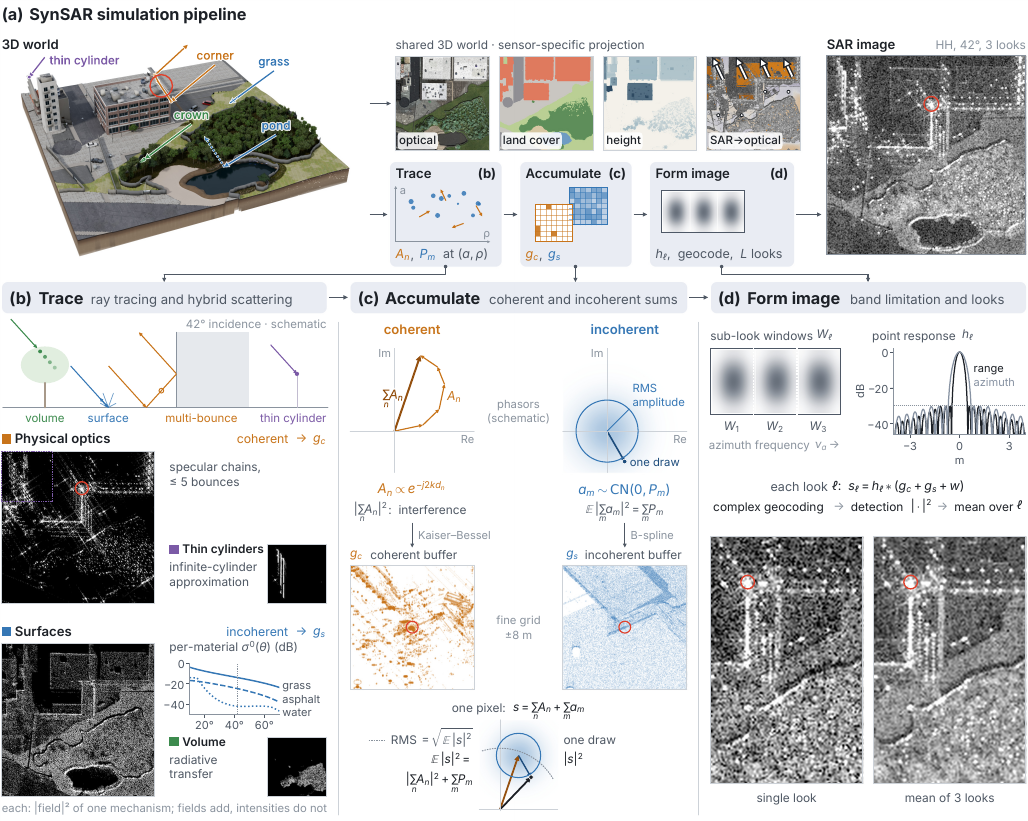}
\caption{\textbf{SynSAR simulation pipeline} for a 108\,m window of one world (HH, $42^\circ$ incidence, three looks). Red circles track the same corner flash.
\textbf{(a)}~Ray tracing produces coherent complex amplitudes $A_n$ and incoherent powers $P_m$ at zero-Doppler coordinates $(a,\rho)$, which form the coherent and incoherent buffers $g_c$ and $g_s$. The upper row shows supervision derived from the same 3D world.
\textbf{(b)}~Isolated intensity contributions from four scattering mechanisms, obtained by differencing complex-valued simulation passes.
\textbf{(c)}~Coherent returns interfere according to their propagation phase ($d_n$: half the path length; $k$: wavenumber), while circular Gaussian incoherent returns add in expected power. Buffers are shown before band limitation ($\pm$8\,m).
\textbf{(d)}~Azimuth sub-look windows $W_\ell$ determine the point responses $h_\ell$; sub-looks are geocoded in the complex domain, detected, and averaged.
Drawings are schematic; images and curves are simulated.}
\label{fig:method}
\end{figure*}

A SAR image is the band-limited coherent sum of the echoes of all scatterers, each placed at its along-track position and slant range.
SynSAR therefore models the single-look complex image as
\begin{equation}
\begin{aligned}
s(\mathbf{r}) ={}& \sum_{n} A_n\, h(\mathbf{r}-\mathbf{r}_n) + \sum_{m} a_m\, h(\mathbf{r}-\mathbf{r}_m) + w(\mathbf{r}),\\
& a_m \sim \mathcal{CN}(0, P_m),
\end{aligned}
\label{eq:image}
\end{equation}
where $A_n$ are deterministic complex amplitudes of explicitly traced structures, $P_m$ are expected powers of sub-resolution scatterers, $\mathbf{r}_n$ and $\mathbf{r}_m$ are their radar coordinates, $h$ is the point-spread function, and $w$ is thermal noise of power $N_0$.
Coherent returns interfere, whereas incoherent returns add in expected power: $\mathbb{E}|s|^2 = |\sum_n A_n h|^2 + \sum_m P_m |h|^2 + N_0$.
\Cref{eq:image} organizes the pipeline (\cref{fig:method}): given a procedural world and an acquisition (\cref{sec:worlds}), SynSAR finds where each return lands and how strong it is (\cref{sec:scatter}), applies the sensor response $h$ (\cref{sec:formation}), and derives supervision from the same world and returns (\cref{sec:supervision}).

\subsection{Procedural worlds and acquisitions}
\label{sec:worlds}

Dataset scale rules out hand-built scenes, so SynSAR generates 154\,m and 512\,m square worlds procedurally from 2D layouts of land cover, building footprints, roads, and vegetation, on terrain that ranges from coastal plains and valley towns to terraced and forested hills.
Buildings have detailed roofs and facades, trees have trunks, branches, and volumetric crowns, and vehicles, fences, poles, and power lines add finer structure.
Beyond geometry, ground parcels carry the physical state that X-band scattering depends on: soil moisture, surface roughness, vegetation height and water content, pavement type, and wind over water.

Acquisitions follow the product configurations of Umbra and Capella~\cite{umbra_opendata,capella_opendata}, commercial X-band VHR constellations (the band of all our evaluation data), with typical resolutions of 0.25--1.4\,m.
They preserve the documented relations among incidence angle, squint, slant range, resolution, and polarization (HH or VV); the number of looks follows catalog metadata where available, and look direction and noise floor vary independently.
The same world can be imaged again under other acquisitions (\cref{sec:s-div}).

\subsection{Radar geometry and hybrid scattering}
\label{sec:scatter}

\paragraph{Radar geometry.}
SynSAR works in zero-Doppler coordinates $(a,\rho)$: a return is placed at its azimuth $a$, the position along the straight sensor track, and at its slant range $\rho$, the shortest distance to the track.
Layover and foreshortening follow from this projection, shadow from occlusion along the line of sight, and a multi-bounce return lands midway in azimuth between its first and last interactions, at half its total path length~\cite{auer2010raytracing}.

\paragraph{Resolve structure, sample roughness.}
At X band the wavelength is about 3\,cm: buildings, terrain, and tree trunks are far larger and can be traced explicitly, whereas surface roughness and foliage vary at its scale and cannot be modeled as geometry.
A resolution cell, however, contains many such scatterers, so their summed return is circular complex Gaussian and is described by its expected power alone~\cite{goodman1976speckle}.
A GPU ray tracer thus follows each incident ray through up to five specular bounces, propagating the polarized field through Fresnel reflection and roughness attenuation~\cite{bhalla1995image,ling1989sbr}, and evaluates two terms at every interaction visible to the sensor (\cref{fig:method}b).
The \textbf{coherent} term $A_n$ is a physical-optics integral over the ray tube that keeps the two-way propagation phase~\cite{bhalla1995image}; it produces specular glints, corner reflections, and wall--ground returns.
The \textbf{incoherent} term $P_m$ is the expected backscatter of the material, from small-perturbation theory for building surfaces and pavements~\cite{rice1951reflection}, semi-empirical models for soil~\cite{oh2004retrieval}, and the water-cloud model for grass and crops~\cite{attema1978water}.
Tree crowns are traced as volumes with extinction and first-order volume scattering~\cite{ulaby1990mimics}, and wires, rails, and masts use an infinite-cylinder approximation~\cite{karam1988cylinders}.
A 154\,m world yields about $3\times10^8$ contributions and a 512\,m world about $3\times10^9$ (expressions in \cref{sec:s-scattering}).

\begin{figure}[t]
\centering
\includegraphics[width=\linewidth]{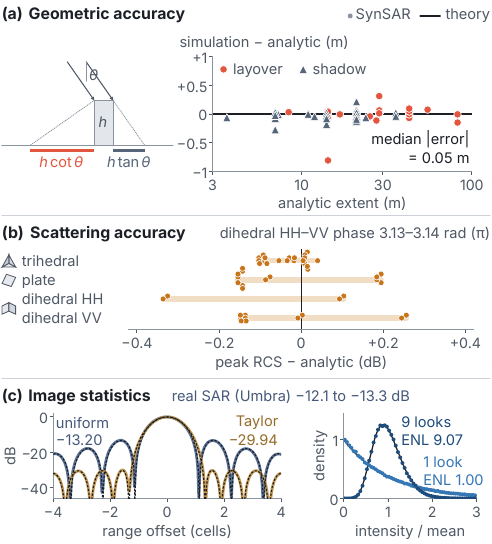}
\caption{\textbf{Physical and statistical validation of SynSAR.}
\textbf{(a)}~Layover and shadow extent errors against analytic predictions across 14 acquisition configurations (median absolute error: 0.05\,m).
\textbf{(b)}~Peak RCS errors of canonical reflectors and dihedral HH--VV phase difference.
\textbf{(c)}~Simulated point responses and speckle distributions compared with theory and with point targets in real SAR images (Umbra, uniform window).
Details in \cref{tab:s-validation}.}
\label{fig:validation}
\end{figure}

\subsection{Image-domain formation}
\label{sec:formation}

Learning tasks use the focused image, not the raw echoes.
For an ideal processor, focusing leaves each return as a copy of $h$ at its radar coordinates, so SynSAR evaluates \cref{eq:image} directly, without synthesizing or focusing raw echoes.
Resolution, sidelobes, and looks are thus set directly, and ray tracing dominates the cost.
Coherent amplitudes are accumulated as phasors, and each incoherent power is drawn as a circular complex Gaussian amplitude, which yields speckle with the correct mean power (\cref{fig:method}c).
The field is then band-limited by the spectral window of the acquisition.
For multi-look products, non-overlapping azimuth sub-bands form sub-looks, each at the specified resolution, which are geocoded in the complex domain onto a reference plane or a 30\,m DEM, detected, and averaged (\cref{fig:method}d).
Geocoding keeps the height-dependent displacement of everything above the reference surface (\cref{sec:s-formation}).

\subsection{Geometrically consistent supervision}
\label{sec:supervision}

All supervision derives from the 3D world and sensor geometry that form the image (\cref{fig:method}a).
Land cover, in the eight classes of OEM-SAR~\cite{xia2025oemsar}, and surface height above ground are rasterized in map coordinates, consistent with OEM-SAR, whose labels are annotated at optical image positions; all land cover mapping and height experiments train on these labels.
Because a SAR pixel can collect returns from several objects, we also record for each radar pixel the class with the strongest expected backscatter after the point-spread function, with a separate label below the noise floor; these radar-pixel classes are used only to construct the L1 and GS images of \cref{sec:mechanisms}.
Optical images are rendered from the same world at another date, so that static geometry is shared while appearance changes.

\paragraph{Dense correspondences with validity.}
Every surface point visible in the optical image is projected into the SAR image through the zero-Doppler geometry.
The simulator also knows where such a match is unreliable: a correspondence is kept only if the surface is visible to the radar, a single object dominates the return, and the contributing surfaces agree in class and height.
Ambiguous layover and mixed returns are excluded, and below the noise floor validity is decided from geometry alone.
Forward and inverse correspondence fields agree to a median round-trip discrepancy of 0.36 pixels over all 12,045 worlds.
Since all products share one world, individual scattering mechanisms can be switched off with the world and labels fixed (\cref{sec:mechanisms}).

\subsection{Physical and statistical validation}
\label{sec:validation}

We validate imaging geometry, canonical scattering responses, and image formation against analytic and statistical references (\cref{fig:validation}; \cref{tab:s-validation}).
Simulated layover and shadow extents agree with closed-form predictions across 14 acquisition configurations, with a median absolute error of 0.05\,m.
Peak radar cross sections of canonical reflectors differ from analytic solutions by at most 0.34\,dB, while dihedral reflections reproduce the expected HH--VV phase difference of $\pi$.
Simulated point responses match theoretical sidelobe levels, and single- and nine-look speckle follows the expected Gamma distributions~\cite{goodman1976speckle} (ENL 1.00 and 9.07).
The simulated uniform-window sidelobes also agree with those of point targets measured in real Umbra images~\cite{umbra_opendata}.

\section{Experimental setup}
\label{sec:setup}

\begin{table*}[t]
\caption{\textbf{Setup of the main experiments} (\cref{tab:registration,tab:segmentation,tab:height}): training on SynSAR, real data, or both; testing on real data. Sizes in pixels.}
\label{tab:setup}
\centering\footnotesize
\setlength{\tabcolsep}{3.5pt}
\newcommand{\setupsep}{\arrayrulecolor{black!18}\specialrule{0.3pt}{2.5pt}{2.5pt}\arrayrulecolor{black}}
\newcommand{\sub}[1]{{\color{sdgray}#1}}
\begin{tabularx}{\textwidth}{@{}>{\raggedright\arraybackslash}p{0.085\textwidth}>{\raggedright\arraybackslash}p{0.175\textwidth}>{\raggedright\arraybackslash}p{0.115\textwidth}>{\raggedright\arraybackslash}X>{\raggedright\arraybackslash}X>{\raggedright\arraybackslash}p{0.115\textwidth}@{}}
\toprule
Task & Model & \cellcolor{oursrow}SynSAR training & Real training & Real test & Metric \\
\midrule
\textbf{Registration} & five matchers\newline\sub{(\cref{tab:registration})} & \cellcolor{oursrow}191 image pairs\newline\sub{$1024^2$ or $1707^2$} & BRIGHT~\cite{chen2025bright}: 681 image pairs\newline\sub{$1024^2$, four scenes} & BRIGHT + SpaceNet~9~\cite{hansch2024spacenet9}\newline\sub{6 scenes, 1,106 control points} & \Sten (\%)\newline\sub{within 10 pixels} \\
\setupsep
\textbf{Land cover mapping} & DINOv3-L + DPT~\cite{simeoni2025dinov3,ranftl2021dpt}\newline\sub{and U-Net~\cite{ronneberger2015unet}} & \cellcolor{oursrow}9,251 images\newline\sub{$307^2$ or $512^2$} & OEM-SAR~\cite{xia2025oemsar}: 4,333 images\newline\sub{$1024^2$, optical pseudo labels} & OEM-SAR: 490 images\newline\sub{$1024^2$, human labels} & \miou (\%)\newline\sub{eight classes} \\
\setupsep
\textbf{Height estimation} & DINOv3-L + DPT & \cellcolor{oursrow}9,251 images\newline\sub{$307^2$ or $512^2$} & SN6~\cite{shermeyer2020spacenet6}, AMS~\cite{umbra_opendata,capella_opendata}\newline\sub{8,461 and 836 images, $512^2$} & SN6, AMS test areas\newline\sub{3,575 and 787 images, $512^2$} & MAE (m)\newline\sub{building pixels} \\
\bottomrule
\end{tabularx}
\end{table*}

We evaluate SynSAR on three real-world tasks: SAR--optical registration, land cover mapping, and building height estimation (\cref{tab:setup}).
We examine transfer from synthetic training alone and with limited real supervision (\cref{sec:registration,sec:segmentation,sec:height,sec:limited}), which imaging mechanisms matter for each task (\cref{sec:mechanisms}), and how physics-based simulation compares with learned optical-to-SAR translation (\cref{sec:translation}).

\paragraph{SynSAR data.}
SynSAR contains one image per world: 11,847 images of 154\,m worlds, with $307^2$ or $512^2$ pixels at a ground sampling distance (GSD) of 0.5 or 0.3\,m, and 198 of 512\,m worlds, with $1024^2$ or $1707^2$ pixels; the small worlds provide scene diversity for land cover mapping and height estimation, and the large worlds give registration wider spatial context.
Training uses the 9,251 small-world images in the Umbra spotlight mode of OEM-SAR (VV, 24--50$^\circ$ incidence, 0.25--1.0\,m resolution) and 191 large-world pairs, with seven held out (HH/VV, spotlight and stripmap, 20--54$^\circ$, 0.3--1.4\,m); ranges are 5th--95th percentiles.
On a single NVIDIA GB200 GPU, SynSAR generates about 200 small or 7 large worlds per GPU-hour.

\paragraph{Training settings.}
We compare training on SynSAR alone, real data alone, and combinations of both.
We also evaluate unsupervised domain adaptation (UDA) by teacher--student self-training~\cite{zou2018cbst,xie2020noisystudent}: a SynSAR-trained teacher generates confidence-filtered pseudo labels on unlabeled real SAR images, which train a student.
Results are averaged over three seeds unless otherwise noted; protocols and statistical analyses are given in \cref{sec:s-protocols}.

\section{Results}
\label{sec:experiments}

\begin{table}[t]
\caption{\textbf{SAR--optical registration on six real evaluation scenes.} \Sten (\%) averaged across scenes. $\Delta$: SynSAR only minus Real only; Wins: scenes favoring SynSAR only.}
\label{tab:registration}
\centering\small
\setlength{\tabcolsep}{1.75pt}
\begin{tabular}{@{}lccccc@{}}
\toprule
& \multicolumn{3}{c}{\Sten (\%)\,$\uparrow$} & & \\
\cmidrule(lr){2-4}
Matcher & \hd{Zero-\\shot} & \hd{Real\\only} & \cellcolor{oursrow}\hd{SynSAR\\only} & $\Delta$ & Wins \\
\midrule
\multicolumn{6}{@{}l}{\emph{Dense, RoMa-based}} \\
\quad MINIMA-RoMa & 51.7 & 65.5{\scriptsize\color{sdgray}$\,\pm$1.4} & \cellcolor{oursrow}69.6{\scriptsize\color{sdgray}$\,\pm$0.4} & $+$4.0 & 4/6 \\
\quad MatchAnything-RoMa & 57.8 & 63.7{\scriptsize\color{sdgray}$\,\pm$1.0} & \cellcolor{oursrow}68.7{\scriptsize\color{sdgray}$\,\pm$0.2} & $+$5.0 & 5/6 \\
\quad RoMa & 18.3 & 64.4{\scriptsize\color{sdgray}$\,\pm$0.2} & \cellcolor{oursrow}67.9{\scriptsize\color{sdgray}$\,\pm$1.1} & $+$3.5 & 4/6 \\
\addlinespace[2pt]
\multicolumn{6}{@{}l}{\emph{Semi-dense and sparse}} \\
\quad XoFTR & 9.5 & 18.7{\scriptsize\color{sdgray}$\,\pm$2.2} & \cellcolor{oursrow}54.3{\scriptsize\color{sdgray}$\,\pm$2.0} & $+$35.6 & 6/6 \\
\quad SuperPoint+LightGlue & 11.9 & 21.2{\scriptsize\color{sdgray}$\,\pm$1.3} & \cellcolor{oursrow}48.8{\scriptsize\color{sdgray}$\,\pm$1.9} & $+$27.5 & 6/6 \\
\bottomrule
\end{tabular}

\end{table}

\begin{table}[t]
\caption{\textbf{Land cover mapping on 490 human-labeled OEM-SAR test images.} Optical pseudo: labels derived from optical imagery; human: manually labeled validation images.}
\label{tab:segmentation}
\centering\small
\begin{tabularx}{\linewidth}{@{}X>{\centering\arraybackslash}p{15mm}>{\centering\arraybackslash}p{15mm}@{}}
\toprule
& \multicolumn{2}{c}{\miou (\%)\,$\uparrow$} \\
\cmidrule(l){2-3}
Training & DINOv3 & U-Net \\
\midrule
\multicolumn{3}{@{}l}{\emph{SynSAR-based}} \\
\multicolumn{1}{@{}>{\columncolor{oursrow}[0pt][\tabcolsep]}l}{\quad SynSAR} & \cellcolor{oursrow}23.4{\scriptsize\color{sdgray}$\,\pm$0.2} & \multicolumn{1}{>{\columncolor{oursrow}[\tabcolsep][0pt]\centering\arraybackslash}p{15mm}@{}}{20.2{\scriptsize\color{sdgray}$\,\pm$0.6}} \\
\multicolumn{1}{@{}>{\columncolor{oursrow}[0pt][\tabcolsep]}l}{\quad SynSAR + UDA} & \cellcolor{oursrow}27.5{\scriptsize\color{sdgray}$\,\pm$1.9} & \multicolumn{1}{>{\columncolor{oursrow}[\tabcolsep][0pt]\centering\arraybackslash}p{15mm}@{}}{26.9{\scriptsize\color{sdgray}$\,\pm$1.3}} \\
\addlinespace[3pt]
\multicolumn{3}{@{}l}{\emph{Real SAR training}} \\
\quad BRIGHT {\footnotesize\color{sdgray}(optical pseudo)} & 31.4{\scriptsize\color{sdgray}$\,\pm$0.6} & 24.0{\scriptsize\color{sdgray}$\,\pm$0.7} \\
\quad OEM-SAR train {\footnotesize\color{sdgray}(optical pseudo)} & 36.7{\scriptsize\color{sdgray}$\,\pm$0.7} & \textbf{33.8}{\scriptsize\color{sdgray}$\,\pm$0.6} \\
\quad OEM-SAR val {\footnotesize\color{sdgray}(human)} & \textbf{37.3}{\scriptsize\color{sdgray}$\,\pm$1.0} & 33.7{\scriptsize\color{sdgray}$\,\pm$0.3} \\
\bottomrule
\end{tabularx}

\vspace{8pt}
\caption{\textbf{Building height estimation on real SN6 and AMS test sets.} Real SAR training: in-domain on its own city or cross-city.}
\label{tab:height}
\begin{tabularx}{\linewidth}{@{}X>{\centering\arraybackslash}p{15mm}>{\centering\arraybackslash}p{15mm}@{}}
\toprule
& \multicolumn{2}{c}{Building MAE (m)\,$\downarrow$} \\
\cmidrule(l){2-3}
Training & SN6 & AMS \\
\midrule
\multicolumn{3}{@{}l}{\emph{SynSAR-based}} \\
\multicolumn{1}{@{}>{\columncolor{oursrow}[0pt][\tabcolsep]}l}{\quad SynSAR} & \cellcolor{oursrow}7.73{\scriptsize\color{sdgray}$\,\pm$0.21} & \multicolumn{1}{>{\columncolor{oursrow}[\tabcolsep][0pt]\centering\arraybackslash}p{15mm}@{}}{7.50{\scriptsize\color{sdgray}$\,\pm$0.09}} \\
\multicolumn{1}{@{}>{\columncolor{oursrow}[0pt][\tabcolsep]}l}{\quad SynSAR + UDA} & \cellcolor{oursrow}7.51{\scriptsize\color{sdgray}$\,\pm$0.15} & \multicolumn{1}{>{\columncolor{oursrow}[\tabcolsep][0pt]\centering\arraybackslash}p{15mm}@{}}{7.26{\scriptsize\color{sdgray}$\,\pm$0.11}} \\
\addlinespace[3pt]
\multicolumn{3}{@{}l}{\emph{Real SAR training}} \\
\quad SN6 real training & \textbf{6.66}{\scriptsize\color{sdgray}$\,\pm$0.02} & 8.82{\scriptsize\color{sdgray}$\,\pm$0.02} \\
\quad AMS real training & 8.47{\scriptsize\color{sdgray}$\,\pm$0.11} & \textbf{5.38}{\scriptsize\color{sdgray}$\,\pm$0.07} \\
\bottomrule
\end{tabularx}

\end{table}

\subsection{SAR--optical registration}
\label{sec:registration}

We fine-tune five matchers for 12,000 steps on SynSAR or real pairs and compare them with their released weights (\cref{tab:registration}).
The co-registered real pairs cover more ground than SynSAR pairs (73.0 vs.\ 50.1\,km$^2$), on scenes disjoint from the evaluation scenes; both use their own valid correspondences under random warps, as is common~\cite{detone2018superpoint,lindenberger2023lightglue,wu2026soma1m}.
\Sten is measured at native resolution after a coarse 1\,m pass, as protocols strongly affect accuracy~\cite{corley2026pretrained}.

SynSAR-only training outperforms real-pair training for all five matchers, by 3.5--35.6 points.
The dense RoMa-based matchers~\cite{edstedt2024roma,ren2025minima,he2025matchanything} gain 3.5--5.0 points, and RoMa, pretrained on optical pairs only, reaches the level of the two with cross-modal pretraining.
XoFTR~\cite{tuzcuoglu2024xoftr} and SuperPoint+LightGlue, which reach only 18.7 and 21.2 with real pairs, gain 35.6 and 27.5 points.
All five outperform their released weights and lead on 4--6 of six scenes (\cref{sec:s-reg}).
The two sources differ in correspondence supervision and scene diversity (\cref{tab:s-reg-data}), not only in appearance.

\begin{figure*}[t]
\centering
\includegraphics[width=\textwidth]{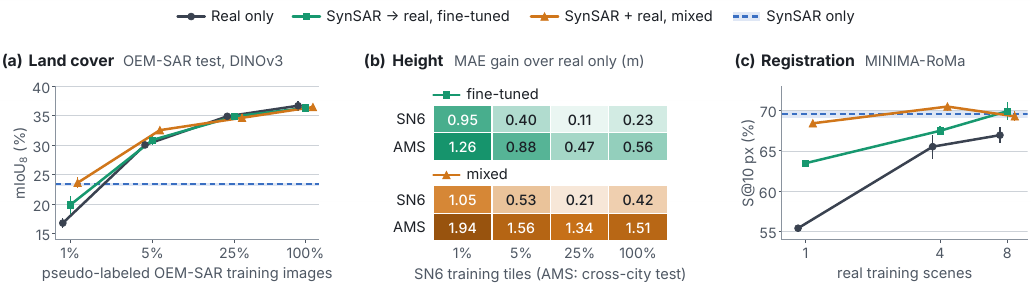}
\caption{\textbf{SynSAR with increasing real supervision.} \textbf{(a)}~Land cover \miou on OEM-SAR. \textbf{(b)}~Building MAE reduction relative to real-only training on SN6 and cross-city AMS (positive is better). \textbf{(c)}~Registration \Sten on six evaluation scenes (one scene: mean of four runs, each on a different scene). Dashed lines indicate SynSAR-only performance. Markers are offset horizontally for visibility.}
\label{fig:limited}
\end{figure*}

\subsection{Land cover mapping}
\label{sec:segmentation}

Training on SynSAR alone reaches 23.4 \miou with DINOv3 and 20.2 with U-Net (\cref{tab:segmentation}), and teacher--student UDA on unlabeled real SAR raises these to 27.5 and 26.9.
Models trained on real SAR remain more accurate (31.4--37.3 with DINOv3), whether on optical-derived pseudo labels from BRIGHT~\cite{chen2025bright} or OEM-SAR or on 210 human-labeled OEM-SAR images.
The gap is uneven across classes: SynSAR comes within 3.6--5.9 IoU points of OEM-SAR pseudo-label training on trees, roads, and developed areas, but trails it widely on agriculture and water (\cref{sec:s-landcover}).
\Cref{sec:limited} adds real supervision.

\subsection{Building height estimation}
\label{sec:height}

On SN6 (Rotterdam) and AMS (Amsterdam), with heights from airborne laser data~\cite{ahn_ahn4}, real-trained models are most accurate in their own city (6.66\,m and 5.38\,m; \cref{tab:height}) but degrade in the other (8.47\,m and 8.82\,m).
Without training on either city, SynSAR reaches 7.73\,m on SN6 and 7.50\,m on AMS, 0.74\,m and 1.32\,m better than these cross-city models, mainly on buildings of 10\,m and taller (\cref{sec:s-height}).
UDA further lowers these to 7.51\,m and 7.26\,m.
SynSAR thus generalizes across cities better than real training in another city, although in-domain real training remains best.

\subsection{Learning with limited real supervision}
\label{sec:limited}

We next vary the amount of real supervision (\cref{fig:limited}): the fraction of pseudo-labeled OEM-SAR training images for land cover mapping, the fraction of SN6 training tiles for height estimation, and the number of real training scenes for registration (one, four, or eight, the last adding four further BRIGHT scenes).
We compare real-only training with fine-tuning from a SynSAR model and with mixed training on equal shares of synthetic and real samples.

For land cover mapping, mixing improves \miou by 6.8 points with 1\% of the real training images and by 2.5 with 5\%, with no detectable gain at 25\% or above.
For height estimation, mixing reduces MAE at every fraction on both test sets, by 1.51\,m on AMS even with all SN6 tiles.
For registration, real-only training with one scene depends strongly on the scene (38.4--63.4 \Sten), whereas fine-tuning and mixing narrow the range to 54.4--69.5 and 63.4--70.8; with eight scenes, real-only training (66.9) stays below SynSAR only (69.6), and mixing helps with four or eight.
Overall, mixing is more consistent than fine-tuning, particularly with scarce real supervision or across cities.

\subsection{Which simulation mechanisms matter?}
\label{sec:mechanisms}

\begin{figure*}[t]
\centering
\includegraphics[width=\textwidth]{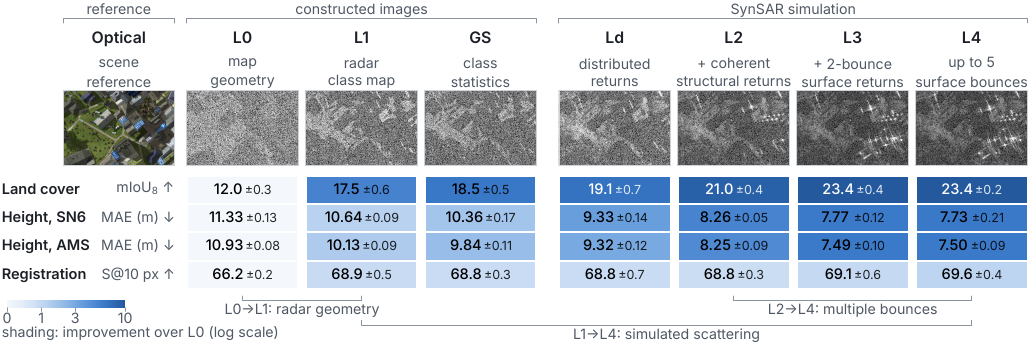}
\caption{\textbf{Task-dependent effects of SAR imaging mechanisms.} Top: one scene at each level. Bottom: real-world transfer; shading: gain over L0 in units of 1 \miou, 5\% of L0 building MAE, or 3 \Sten. For registration, L0$\to$L1 also changes correspondence supervision.}
\label{fig:matrix}
\end{figure*}

We generate the SynSAR training set of each task at seven controlled SAR simulation levels (\cref{fig:matrix}).
Starting from class-mean intensities with speckle in map geometry (L0), we introduce radar geometry (L1), class-wise intensity statistics (GS), traced distributed scattering (Ld), coherent structural returns (L2), and additional surface bounces (L3/L4).
Scene content, dense-prediction labels, and training settings remain fixed.

\paragraph{Dense prediction benefits substantially from imaging physics.}
Radar geometry improves land cover mapping from 12.0 to 17.5 \miou (L0$\to$L1), while class-wise intensity statistics add little.
Scattering and multiple bounces further raise \miou to 23.4 and reduce building height errors.
On SN6, MAE decreases from 11.33\,m (L0) to 7.73\,m (L4).
The observed differences between L3 and L4 are small.

\paragraph{Registration changes most from L0 to L1.}
L0 already matches real-pair training (66.2 vs.\ 65.5).
From L0 to L1, where image geometry and correspondence supervision both change, \Sten rises by 2.6 points (95\%~CI [1.5, 3.8]; \cref{tab:s-register-c}); L1 to L4 adds little ($+0.7$ [$-0.5$, 1.8]).

\subsection{Physics-based simulation vs.\ learned optical-to-SAR translation}
\label{sec:translation}

\begin{figure*}[t]
\centering
\includegraphics[width=\textwidth]{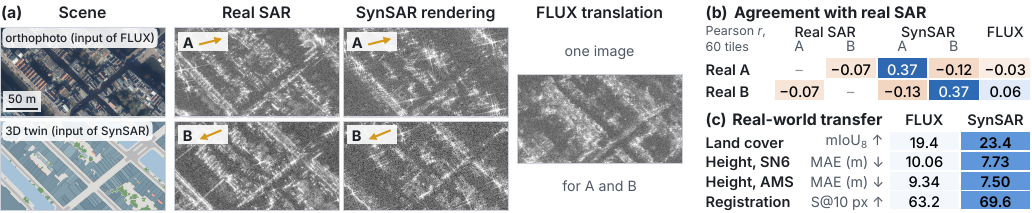}
\caption{\textbf{Physics-based simulation vs.\ learned translation.} \textbf{(a)}~An Amsterdam tile under two real acquisitions, A and B, with opposite look directions (arrows). SynSAR renders its 3D twin under each geometry; FLUX translates the orthophoto~\cite{beeldmateriaal_ortho} once. \textbf{(b)}~Mean $r$ with the real images (2\,m blocks). \textbf{(c)}~Real-world transfer from the worlds of \cref{fig:matrix} with FLUX translations or SynSAR images (L4).}
\label{fig:translation}
\end{figure*}

We compare SynSAR with learned optical-to-SAR translation (\cref{fig:translation}; \cref{sec:s-cmp}): FLUX~\cite{bfl2024flux}, with its pretrained weights frozen, is adapted by LoRA on 4,543 real optical--SAR pairs of OEM-SAR.

\paragraph{Downstream transfer.}
FLUX translates the same training worlds as in \cref{sec:mechanisms}, and the translations replace the SynSAR images, with labels and training unchanged.
For land cover mapping and height, the input of FLUX is an optical image that Changen2~\cite{zheng2025changen2} generates from the land cover map of the world.
For registration, it is the rendered optical image of each pair, which keeps the translation aligned with that image and gives the highest accuracy among the inputs we tested; the correspondences then follow map geometry, as the translation lies on the optical grid.
Models trained on the translations trail those trained on the SynSAR images of the same worlds (L4) on every task (\cref{fig:translation}c).
With real optical images of BRIGHT as input, FLUX also remains below SynSAR in land cover mapping and registration (\cref{tab:s-flux-all}).

\paragraph{Agreement with real acquisitions.}
For 60 Amsterdam tiles observed by two Umbra acquisitions~\cite{umbra_opendata} from opposite directions, we build digital twins from public 3D building models, topographic maps, and laser scans~\cite{peters2022bag3d,bgt_pdok,ahn_ahn4}.
SynSAR renders each twin under both geometries; FLUX translates the orthophoto into a single image.
On 2\,m blocks, the two real images of a tile are nearly uncorrelated ($r=-0.07$; \cref{fig:translation}b).
SynSAR renderings correlate with the real image of the same geometry ($r=0.37$) but not with the opposite one ($-0.12$, $-0.13$); FLUX translations stay near zero ($-0.03$, 0.06).
SynSAR's agreement is moderate, but it follows the acquisition geometry.
As the twin is identical in both renderings, this agreement comes from the simulated acquisition geometry, despite approximate materials.
FLUX starts from the real orthophoto but, without geometry input, agrees with neither acquisition; translators conditioned on acquisition geometry may differ.

\section{Discussion and conclusion}
\label{sec:discussion}

SynSAR shows that physics-based SAR simulation can scale to dataset level and transfer to real VHR acquisitions.
What the simulation must model depends on the downstream task.
Registration relies mainly on where returns land: with radar geometry and validity-aware correspondences, simplified backscatter is nearly as effective as the full simulation.
Land cover mapping and height estimation also depend on how the scene scatters: distributed scattering, coherent structural returns, and multiple bounces narrow the gap to real training.
These findings suggest that synthetic SAR datasets should be designed around task-specific learning requirements rather than uniformly maximizing physical fidelity.
Synthetic data also complement real supervision, particularly when annotations are scarce.
Rendered from 3D worlds under explicit acquisition geometry, SynSAR transfers better than the tested optical-to-SAR translation and follows look-direction effects that it misses.

To reach dataset scale, SynSAR trades raw-echo synthesis for image-domain formation, and its procedural worlds and scattering models necessarily approximate real geographic and material diversity.
Within this scope, SynSAR provides both multi-task training supervision and a controlled framework for understanding which physical mechanisms drive real-world SAR transfer.

\begingroup
\let\clearpage\relax
\clearpage
\section*{LLM Usage Statement}
Claude Opus 5.5, Claude Fable 5.2, and GPT-6 Astra assisted with the development and review of the simulator code, literature screening, and draft revision and language editing of the paper.
The authors verified the scientific content and take full responsibility for the work.

\endgroup
\begingroup
\small
\bibliography{references}
\endgroup

\clearpage
\preprintsupphead
\addtocontents{toc}{\protect\SuppTocStart}
\setcounter{section}{0}\setcounter{subsection}{0}\setcounter{table}{0}\setcounter{figure}{0}
\renewcommand{\thesection}{S\arabic{section}}
\renewcommand{\thetable}{S\arabic{table}}
\renewcommand{\thefigure}{S\arabic{figure}}
\renewcommand{\theHsection}{S\arabic{section}}
\renewcommand{\theHtable}{S\arabic{table}}
\renewcommand{\theHfigure}{S\arabic{figure}}
\renewcommand{\topfraction}{0.9}\renewcommand{\dbltopfraction}{0.9}\renewcommand{\textfraction}{0.07}\renewcommand{\floatpagefraction}{0.75}\renewcommand{\dblfloatpagefraction}{0.75}\setcounter{topnumber}{3}\setcounter{dbltopnumber}{3}\setcounter{totalnumber}{4}%
\section{SynSAR dataset and supervision}
\label{sec:s-dataset}

This section supplements Secs.~3.1 and 3.4 of the main paper with details of the dataset and its supervision.

\subsection{Dataset overview and examples}
\label{sec:s-overview-data}

SynSAR comprises 11,847 independently generated 154\,m worlds and 198 larger 512\,m worlds (\cref{tab:s-dataset}).
The smaller worlds support land cover mapping and building height estimation, while the larger worlds provide broader spatial context for SAR--optical registration.
Each world provides SAR imagery under sampled acquisition settings, optical imagery rendered at another date, land cover and height maps, and geometry-derived SAR--optical correspondences (\cref{sec:s-supervision}).

\begin{figure*}[p]
\centering
\includegraphics[width=\textwidth]{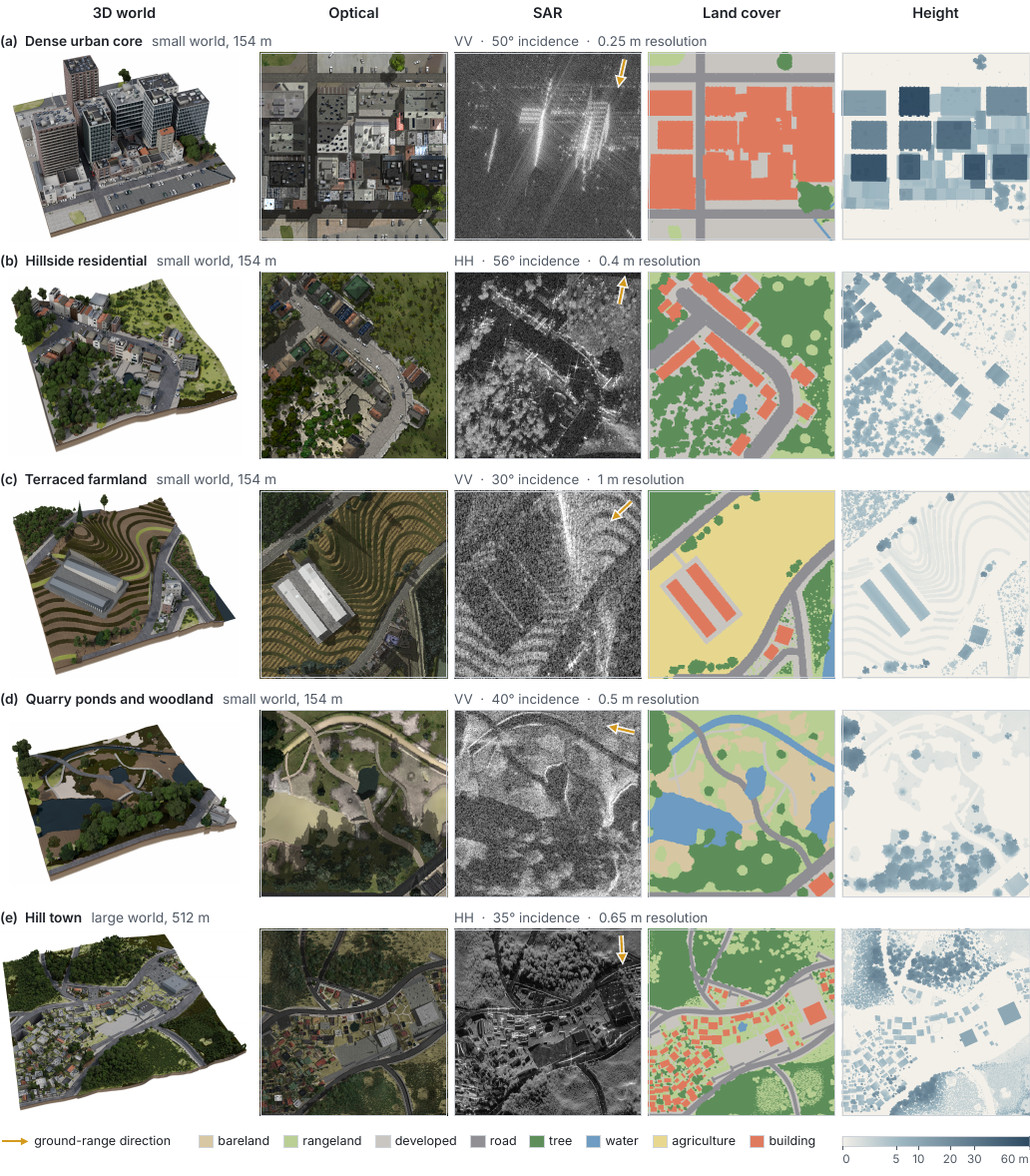}
\caption{\textbf{Examples of SynSAR worlds.}
Arrows point along ground range, away from the radar.
SAR images are individually stretched between the 1st and 99.6th percentiles in dB; height maps use a square-root scale.}
\label{fig:s-gallery}
\end{figure*}

\begin{table}[t]
\caption{\textbf{SynSAR dataset composition and acquisition characteristics.}
Incidence-angle and resolution ranges are the 5th--95th percentiles.
Across all small worlds, acquisitions span spotlight and stripmap modes, VV and HH polarizations, and incidence angles of 6--70$^\circ$; the training subset is selected as in \cref{sec:s-training}.
Seven large worlds were set aside at generation time and are not used in training or evaluation.}
\label{tab:s-dataset}
\centering\footnotesize
\setlength{\tabcolsep}{3pt}\renewcommand{\arraystretch}{0.93}
\begin{tabularx}{\linewidth}{@{}>{\raggedright\arraybackslash}p{0.26\linewidth}>{\raggedright\arraybackslash}X>{\raggedright\arraybackslash}X@{}}
\toprule
& Small worlds & Large worlds \\
\midrule
World size & 154\,m & 512\,m \\
Worlds & 11,847 & 198 \\
\quad with buildings & 9,746 (82\%) & 195 (98\%) \\
SAR geocoding & 30\,m DEM: 8,171\newline plane: 3,676 & 30\,m DEM: 131\newline plane: 67 \\
\multicolumn{3}{@{}l@{}}{\setlength{\tabcolsep}{2pt}\begin{tabular}{@{}l*{8}{c}@{}}
\emph{Land cover (\%)} & Bare & Range & Devel. & Road & Tree & Water & Agri. & Bldg. \\
Small worlds & 1.6 & 20.1 & 17.3 & 9.0 & 18.0 & 4.9 & 10.6 & 18.4 \\
Large worlds & 1.5 & 22.7 & 16.4 & 6.4 & 20.3 & 3.1 & 15.1 & 14.5 \\
\end{tabular}} \\
\addlinespace[2pt]
\multicolumn{3}{@{}l}{\emph{Used in experiments}} \\
Training samples & 9,251 worlds & 191 worlds \\
Mode & one spotlight mode & two spotlight modes (180), stripmap (11) \\
Polarization & VV & VV (143), HH (48) \\
Incidence angle & 24--50$^\circ$ & 20--54$^\circ$ \\
Resolution & 0.25--1.0\,m & 0.3--1.4\,m \\
Pixel spacing (GSD) & 0.5 / 0.3\,m (6,526 / 2,725) & 0.5 / 0.3\,m (159 / 32) \\
\bottomrule
\end{tabularx}
\end{table}

The world generator builds on the procedural approach of SynRS3D~\cite{song2024synrs3d}, extended for SAR simulation; its worlds combine terrain, roads, buildings, vegetation, water bodies, and material properties across diverse landscapes.
\Cref{fig:s-gallery} presents five examples, ranging from dense urban blocks and hillside neighborhoods to terraced farmland, wooded ponds, and a hill town.
The optical, SAR, land cover, and height views show the scene content, the radar appearance, and the supervision derived from the same 3D geometry.

Acquisition settings are sampled from the X-band product configurations of Umbra and Capella~\cite{umbra_opendata,capella_opendata}, preserving documented relationships among incidence angle, squint, slant range, resolution, and polarization.
The number of looks follows catalog metadata when available and is otherwise sampled separately.
Look direction and noise level vary independently.
The same world can be observed under different acquisition settings without changing its underlying geometry (\cref{sec:s-div-protocol}).

\subsection{Geometrically consistent supervision}
\label{sec:s-supervision}

\begin{figure*}[p]
\centering
\includegraphics[width=\textwidth]{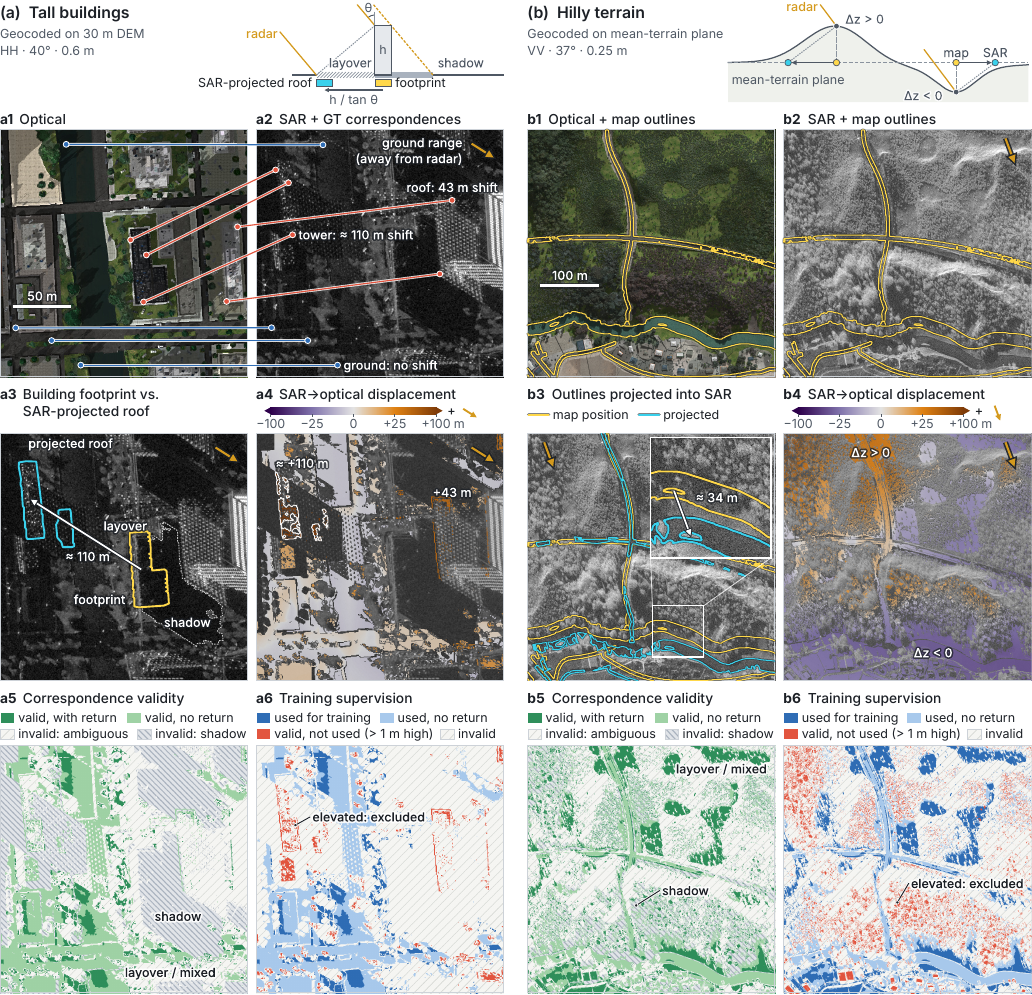}
\caption{\textbf{Geometrically consistent SAR--optical supervision.}
Two large SynSAR worlds illustrate height-dependent displacement and correspondence validity.
Side views are schematic; image panels share a north-up map grid.
(a)~A 91\,m tower appears approximately 110\,m toward the radar.
(b)~Terrain above and below the reference plane shifts in opposite directions.
Arrows in the side views and in a3/b3 point from the map position to the SAR position; the displacement in a4/b4 runs from SAR to optical along ground range (positive away from the radar; square-root scale).
In a5/b5, a correspondence can be valid where the SAR image has no return above the noise floor; a6/b6 keep the valid correspondences on surfaces at most 1\,m high.}
\label{fig:s-supervision}
\end{figure*}

\begin{table*}[t]
\caption{\textbf{SynSAR data products and their roles in downstream learning.}
All raster products use a north-up reference grid, but SAR-derived layers retain radar-induced displacement.
Small and large worlds store the same products.}
\label{tab:s-products}
\centering\small
\setlength{\tabcolsep}{5pt}
\begin{tabularx}{\textwidth}{@{}>{\raggedright\arraybackslash}p{0.20\textwidth}>{\raggedright\arraybackslash}X>{\raggedright\arraybackslash}p{0.28\textwidth}@{}}
\toprule
Data product & Spatial reference / meaning & Use in experiments \\
\midrule
SAR intensity & detected, look-averaged intensity; DEM-based geocoding on a 30\,m DEM or plane-based geocoding at the mean terrain height, with the radar displacement of everything off this surface retained & input for all three tasks \\
Optical image & rendered at another date and orthorectified on the 30\,m DEM; elevated objects lean in off-nadir views & registration input; also translated by FLUX~\cite{bfl2024flux} (\cref{sec:s-flux}) \\
\addlinespace[2pt]
Land cover map & class of the top static surface at each map position, eight classes (\cref{tab:s-landcover}) & land cover mapping target; input of L0 and of the Changen2~\cite{zheng2025changen2}--FLUX baseline (\cref{sec:s-levels,sec:s-flux}) \\
Height map & top static-surface height above the terrain at each map position & height target; used in the registration training mask \\
Radar-geometry class map & class with the strongest expected return in each SAR pixel, or no return & not a training target; L1 and GS use the map of the single-bounce render (Ld; \cref{sec:s-levels}) \\
\addlinespace[2pt]
SAR$\to$optical correspondence & per SAR pixel: optical position of the same surface, with validity & supervision for all matchers \\
Optical$\to$SAR correspondence & per optical pixel: SAR position of the same surface, with validity & additional supervision for XoFTR~\cite{tuzcuoglu2024xoftr} and SuperPoint+LightGlue~\cite{detone2018superpoint,lindenberger2023lightglue}, whose losses match points in both directions \\
Registration training mask & derived: valid correspondences whose matched optical pixels have a map-grid height of at most 1\,m & pixels that supervise the matchers (\cref{fig:s-supervision}) \\
\addlinespace[2pt]
No-return map & SAR pixels without a return above the noise floor; correspondence validity rests on geometry alone & not a training input; no-return pixels can still be supervised (\cref{fig:s-supervision}) \\
Auxiliary annotations & layover/mixed-return, radar-shadow, hidden-from-optical, and transient-object masks; optical-view labels; per-object tables & not used in training; used for validity filtering and figures \\
\bottomrule
\end{tabularx}
\end{table*}

SynSAR derives all supervision from the shared 3D scene and acquisition geometry (\cref{fig:s-supervision}; \cref{tab:s-products}).
SAR images are geocoded onto a north-up grid using a reference DEM or plane, as in terrain- and ellipsoid-corrected commercial products~\cite{capella_products,umbra_gec}, retaining height-dependent imaging displacement.
Land cover and height labels remain at their map positions.
Elevated buildings can thus appear far from their footprints (\cref{fig:s-supervision}a), while terrain above or below a reference plane is displaced in opposite directions (\cref{fig:s-supervision}b).
Labels and heights describe the static scene on the map grid. Each pixel takes the class of the first surface met by a vertical ray at its center, with vehicles, free-standing walls, and fences labeled as the ground below. The height target is the height above the terrain of the top surface at the pixel center, counting walls, fences, tree crowns, and other vegetation canopies but not vehicles.
The optical images are rendered at another date from the same static geometry.
They are orthorectified on the 30\,m reference DEM, so that elevated objects in off-nadir views are displaced as in real orthoimages.

\paragraph{Correspondence construction.}
SAR--optical correspondences are obtained by projecting visible 3D scene points into both views.
A correspondence is retained when visibility and the contributing radar response support an unambiguous association with the same scene element.
Ambiguous layover and mixed returns without a consistent surface association are excluded.
So are pixels compressed from a wider area, as under strong foreshortening, and pixels affected by vehicles or other transient objects at either date.
Valid correspondences can also exist on elevated surfaces, including tree crowns, even when these surfaces are not used for training.
The forward and inverse correspondence fields have a median round-trip raster-coordinate discrepancy of 0.36 pixels across 12,045 worlds (optical to SAR and back through the nearest SAR pixel; median over worlds of the per-world medians).
This discrepancy primarily reflects pixel discretization and measures internal consistency, rather than absolute geometric accuracy.

\paragraph{Validity criteria.}
Power is attributed per return: each traced return, single- or multi-bounce, is assigned to the object and class of the surface of its last reflection. An object's power in a SAR pixel is the expected power of its own returns after the point-spread function; interference between returns of different objects is not attributed to any object.
At a SAR pixel with a return above the noise floor, a valid correspondence requires that one object holds at least 90\% of this power, that the power-weighted scatterer height agrees with the optically visible surface height within 1.5\,m plus 15\% of that height, and that optical surface points mapping to the pixel spread by at most 1\,m in height and one optical pixel (rms) in position.
A pixel with expected scene power below the thermal noise floor is judged by the geometric criteria alone.
Validity therefore means that an unambiguous geometric correspondence exists, not that the SAR image contains a detectable return there; valid pixels without a return above the noise floor are marked separately in \cref{fig:s-supervision}.

\paragraph{Supervision used in learning.}
Land cover mapping and height estimation use map-grid labels and heights, respectively.
The radar-geometry class labels instead support the controlled simulation levels in \cref{sec:s-mech}.
Registration training uses valid correspondences whose matched optical pixels have a map-grid height of at most 1\,m (\cref{fig:s-supervision}, a5--a6 and b5--b6).
This selects predominantly near-ground correspondences.
Across the 191 large training worlds, 38.1\% of SAR pixels meet these training criteria; 37.0\% of these supervised pixels are below the noise floor.

\section{Physics-based simulation and validation}
\label{sec:s-simulation}

This section supplements Secs.~3.2, 3.3, and 3.5 of the main paper with details of the simulation and its validation.

\begin{table}[t]
\caption{\textbf{Main simulation settings.} $k_f$: bandwidth factor of the window (\cref{sec:s-formation}).}
\label{tab:s-simparams}
\centering\footnotesize
\setlength{\tabcolsep}{4pt}
\begin{tabularx}{\linewidth}{@{}>{\raggedright\arraybackslash}p{0.38\linewidth}X@{}}
\toprule
Setting & Value \\
\midrule
\multicolumn{2}{@{}l}{\emph{Scene and sensor}} \\
Carrier, channel & 9.6\,GHz (X band, $\lambda=3.1$\,cm), HH or VV \\
Ground mesh & 0.5\,m (water 2\,m) \\
Curved surfaces & chords of at most 3.3\,mm ($\approx\lambda/10$) \\
Noise-equivalent $\sigma^0$ & drawn uniformly per mode, within $-26.7$ to $-10$\,dB \\
\addlinespace[2pt]
\multicolumn{2}{@{}l}{\emph{Tracing and scattering}} \\
Ray lattice & 1.2\,cm spacing, $\sim\!10^8$ rays per 154\,m world \\
Specular bounces & up to five \\
Volume integration & 0.15\,m steps \\
Crowns & extinction $\kappa_e$ of 0.35--0.40\,Np/m, volume backscatter $\sigma_v = 0.3\,\kappa_e$ \\
\addlinespace[2pt]
\multicolumn{2}{@{}l}{\emph{Image formation}} \\
Slant-plane pixel & 0.6 of the full-band resolution on each axis, at most 0.15\,m \\
Kernels & coherent: Kaiser--Bessel; incoherent: cubic B-spline \\
Spectral window & uniform / $-30$\,dB Taylor ($\bar n = 3$); $k_f = 0.886$ / 1.118 \\
Looks & non-overlapping azimuth sub-bands; catalog value when available, otherwise drawn per mode (1 or 2 for Umbra spotlight) \\
\addlinespace[2pt]
\multicolumn{2}{@{}l}{\emph{Products}} \\
Pixel spacing (GSD) & 0.3\,m for ground resolution $\le$ 0.35\,m, 0.5\,m otherwise \\
Product surface & 30\,m DEM, or plane at the mean terrain height \\
Optical image & 15--1,099 days before the SAR date; nadir or up to 15$^\circ$ off nadir \\
\bottomrule
\end{tabularx}
\end{table}

\subsection{Ray tracing and hybrid scattering}
\label{sec:s-scattering}

\synsar traces specular paths~\cite{ling1989sbr} through a world of surfaces, closed vegetation volumes, and thin linear members (settings in \cref{tab:s-simparams}).
At each reflection, the vector field is multiplied by the Fresnel coefficients of the dielectric or perfectly conducting surface and by the roughness factor $\exp[-2(k\sigma_h\cos\theta_i)^2]$, where $k=2\pi/\lambda$ is the wavenumber, $\sigma_h$ the rms surface height, and $\theta_i$ the local incidence angle.
Each interaction with an unobstructed return path to the sensor contributes two separately evaluated terms: a coherent amplitude $A$ from physical-optics integration over the ray tube~\cite{lee1988raytube,bhalla1995image} and an incoherent power $P$ from the scattering model of the material.
Both refer to the surface element $\Delta S=\Delta S_\perp/\cos\theta_i$ illuminated by a tube of cross-section $\Delta S_\perp$:
\begin{align*}
A &= \frac{\sqrt{4\pi}}{\lambda}\,\Delta S\,(\mathbf{F}\cdot\hat{\mathbf{p}})\,\operatorname{sinc}x_1\,\operatorname{sinc}x_2\,\sqrt{T}\,e^{-j2k(d-R_0)},\\
P &= c_b\,\sigma^0\,\Delta S\,|\mathbf{E}_i|^2\,T,\\
\mathbf{F} &= \tfrac12\bigl[\hat{\mathbf{o}}\times(\hat{\mathbf{o}}\times\mathbf{J})+\hat{\mathbf{o}}\times\mathbf{M}\bigr].
\end{align*}
Here $\mathbf{F}$ is the far field toward the sensor direction $\hat{\mathbf{o}}$ of the tangent-plane currents $\mathbf{J}=\hat{\mathbf{n}}\times(\mathbf{H}_i+\mathbf{H}_r)$ and $\mathbf{M}=-\hat{\mathbf{n}}\times(\mathbf{E}_i+\mathbf{E}_r)$ on a surface with normal $\hat{\mathbf{n}}$, for a unit incident field of polarization $\hat{\mathbf{p}}$ and magnetic fields in units of the electric field.
The sinc factors ($\operatorname{sinc}x=\sin x/x$) integrate the linear phase over $\Delta S$, with $x_1$ and $x_2$ half the two-way phase differences across the tube, and equal one for retro-directive paths such as corner reflectors.
Finally, $T$ is the two-way power transmission through vegetation, $d$ half the total path length, $R_0$ the range to the scene center, $|\mathbf{E}_i|^2$ the squared incident field magnitude after the preceding reflections, and $\sigma^0$ the incoherent scattering coefficient.

Both terms use an RCS-equivalent normalization without range or system factors: the tubes of a smooth conducting plate of area $S$ at normal incidence sum to $|\sum A|^2=4\pi S^2/\lambda^2$, its radar cross section, and a directly illuminated element contributes the mean cross section $P=\sigma^0\Delta S$.
They are the terms $A_n$ and $P_m$ of Eq.~(1) of the main paper and are computed independently, not by dividing a common energy budget: roughness attenuates the coherent field, while $\sigma^0$ sets the incoherent power directly.
Reciprocity fixes $c_b$: an incoherent return after specular reflections has a reversed path of equal length and amplitude that adds in phase, and as only one of the two is traced, $c_b=4$ ($c_b=1$ for single scattering).
Coherent double bounces likewise keep, with doubled amplitude, the ordering that first meets the horizontal or more extended surface, where geometric reflection is most accurate.

The coefficient $\sigma^0$ depends on the surface type and on the ground state of each world and acquisition date.
Building surfaces, pavements, and tree trunks use first-order small-perturbation theory~\cite{rice1951reflection} in bistatic full-vector form.
Natural ground and water use co-polarized backscatter curves $\sigma^0(\theta)$, evaluated at the heuristic effective angle $\theta=\arccos\sqrt{\cos\theta_i\cos\theta_s}$ for scattering angle $\theta_s$: the semi-empirical model of Oh~\cite{oh2004retrieval} for bare soil, an empirical bare-surface model for gravel and sand, and the XMOD2 wind model~\cite{li2014xmod2} with a polarization ratio~\cite{thompson1998polarization} for water, including inland water.
Short grass and crops add a vegetation layer through the water-cloud model~\cite{attema1978water}, $\sigma^0=C\cos\theta\,(1-T_v)+T_v\,\sigma^0_\mathrm{soil}$ with $T_v=\exp(-2BV/\cos\theta)$, where $V$ is the vegetation water content and $B$ and $C$ are vegetation-specific coefficients.

Taller vegetation is traced as a volume with extinction coefficient $\kappa_e$ and volume backscatter coefficient $\sigma_v$, in the first-order approximation of radiative-transfer canopy models~\cite{ulaby1990mimics}.
Along each primary ray, the tracer accumulates $\mathrm{d}P=\sigma_v\,\eta\,e^{-2\tau}\,\Delta S_\perp\,\mathrm{d}\ell$, with relative foliage density $\eta$ and optical depth $\tau$ up to the path element $\mathrm{d}\ell$; all rays are attenuated by the volumes they cross, which yields $T$ and partial shadows behind tree crowns.
A uniform layer of nadir optical thickness $\tau_0$ thus returns $\sigma^0=\tfrac{\sigma_v}{2\kappa_e}\cos\theta\,(1-e^{-2\tau_0/\cos\theta})$, the vegetation term of the water-cloud model for $\sigma_v=2\kappa_eC$ and $\tau_0=BV$; grass and crops traced as volumes use these values, and tree crowns use fixed coefficients (\cref{tab:s-simparams}).
Members too thin for physical optics, such as wires and railings, follow the infinite-cylinder approximation~\cite{karam1988cylinders}: each carries, over its length, the surface current of the infinite conducting cylinder~\cite{ruck1970rcs}, and its direct and surface-reflected returns add to the coherent term.

\subsection{SAR image formation}
\label{sec:s-formation}

The coherent amplitudes $A_n$ and incoherent powers $P_m$ of \cref{sec:s-scattering} are accumulated at their zero-Doppler coordinates on an oversampled complex grid, using a Kaiser--Bessel kernel~\cite{beatty2005gridding} for coherent and a cubic B-spline for incoherent contributions, both compensated in the Fourier domain.
Coherent amplitudes add as phasors and interfere before detection.
Each $P_m$ is an expected power and enters as a zero-mean circular complex Gaussian amplitude of variance $P_m$~\cite{goodman1976speckle}, producing speckle with the mean power of Eq.~(1) of the main paper.
White thermal noise, set by the noise-equivalent $\sigma^0$, is added before band limitation.

Band limitation multiplies the spectrum of this field by a uniform or Taylor window defined along and across the line of sight, and thus rotated with the squint in the image plane; its inverse transform is the point-spread function $h$.
Bandwidths equal $k_f$ divided by the resolution (\cref{tab:s-simparams}), so the $-3$\,dB width of $h$ matches the specified resolution.
The window has unit energy, so a distributed target has mean intensity $\beta^0$ and a point target peaks at its cross section divided by the equivalent resolution-cell area $1/|h(0)|^2$.
Multi-look products split the band across the line of sight into $L$ non-overlapping azimuth sub-bands, each with its own unit-energy window and the specified azimuth resolution; the sub-looks share one complex field, and their non-overlapping bands yield decorrelated speckle for distributed targets.

Each complex look is then geocoded: every pixel of the north-up map grid is placed on a 30\,m DEM block-averaged from the scene terrain or on a plane at the mean terrain height, and the look is interpolated at its zero-Doppler position with cubic splines.
Detection and averaging over looks come last, so the order is complex accumulation, band limitation, complex geocoding, detection, and multi-look averaging.
Neither surface removes the displacement and layover of objects above it, such as buildings and tree crowns (\cref{fig:s-supervision}).
The product is a linear, $\beta^0$-like intensity in 32-bit floating point without radiometric terrain correction; its conversion to network inputs is given in \cref{sec:s-training}.

\subsection{Physical and statistical validation}
\label{sec:s-validation}

\synsar was checked against analytical references for radar geometry and canonical scattering (\cref{tab:s-validation}; Fig.~3 of the main paper).
The geometry fixtures are boxes 10--30\,m tall on flat ground, rendered in 14 acquisitions that vary incidence angle, resolution, look direction, polarization, and mode.
Their layover and shadow extents follow the zero-Doppler predictions, with the largest deviations (0.15\,m, and 0.81\,m at 1\,m resolution) below the resolution cell, and the wall--ground double-bounce line of a concrete box lies within 4.6\,mm of the wall foot in every acquisition.
The scattering references are isolated conducting trihedrals, plates, and dihedrals, whose peak radar cross sections and HH--VV phases agree with the analytic values; the largest deviation, $-0.34$\,dB, is that of a 0.2\,m dihedral at HH ($-0.14$\,dB at VV), and at the 1.2\,cm ray spacing of the dataset the errors of plates and dihedrals are $+0.09$ to $+0.26$\,dB.
For a distributed target, the mean backscatter of a uniform rough surface reproduces its model value within 0.033\,dB.

The image-formation chain was assessed through point-target responses and distributed-scatterer statistics on slant-plane images.
The $-3$\,dB width of the point response is within 1.0\% of the specified resolution, and its median peak sidelobe ratio is within 0.25\,dB of the theoretical value for both windows.
Five- and nine-look images reach equivalent numbers of looks of 4.98 and 9.07, consistent with decorrelated sub-looks, and the speckle grain width matches the value expected for each window.
After geocoding, 16 simulated point targets show a mean peak-position error below 13\,mm relative to their known map coordinates.

For comparison with observed SAR, we measured the sidelobes of isolated point targets in ten Umbra complex images~\cite{umbra_opendata}.
Their per-scene median peak sidelobe ratios of $-12.1$ to $-13.3$\,dB are consistent with the uniform window adopted for this mode, which supports the simulated point-spread function rather than the scattering models.
The checks above confirm that the implementation reproduces the model of \cref{sec:s-scattering,sec:s-formation} on canonical cases; the modeling scope is described in \cref{sec:s-cost}.

\begin{table}[t]
\caption{\textbf{Checks of geometry, scattering, and image formation.} RCS: radar cross section; PSLR: peak sidelobe ratio; ENL: equivalent number of looks. $\Delta$ and the RCS rows give deviations from the reference; ranges are over acquisitions, windows, and polarizations. Plates and dihedrals are traced with 5\,mm rays unless noted. Point-target rows give the median of 16 targets per acquisition (the mean for the geocoded position); the Umbra row gives per-scene medians of observed images.}
\label{tab:s-validation}
\centering\footnotesize
\setlength{\tabcolsep}{3pt}
\begin{tabularx}{\linewidth}{@{}>{\raggedright\arraybackslash}X>{\raggedright\arraybackslash}p{0.21\linewidth}>{\raggedright\arraybackslash}p{0.34\linewidth}@{}}
\toprule
Check & Reference & Measured \\
\midrule
\multicolumn{3}{@{}l}{\emph{Canonical geometry and scattering}} \\
Layover and shadow extent, 112 box edges & zero-Doppler prediction & median $|\Delta|$ 0.05\,m; max.\ 0.15\,m, or 0.81\,m at 1\,m resolution \\
Wall--ground double-bounce line & wall foot & $|\Delta|\le 4.6$\,mm \\
Peak RCS, trihedrals (0.5, 1\,m) & analytic & $-0.10$ to $+0.04$\,dB \\
Peak RCS, plates and dihedrals (0.2--0.3\,m) & analytic & $-0.34$ to $0.00$\,dB; $+0.09$ to $+0.26$\,dB at the 1.2\,cm ray spacing of the dataset \\
HH--VV phase: dihedral / plate, trihedral & $\pi$ / 0 & 3.133--3.135 / $\le 0.011$\,rad \\
Mean level, uniform rough surface & model $\sigma^0(\theta)$ & $|\Delta|\le 0.033$\,dB \\
\addlinespace[2pt]
\multicolumn{3}{@{}l}{\emph{Image formation}} \\
$-3$\,dB width of the point response & specified resolution & within 1.0\% \\
PSLR, uniform / Taylor window & $-13.26$ / $-30.18$\,dB & $-13.13$ to $-13.31$ / $-29.94$ to $-30.21$\,dB \\
ENL, 1 / 5 / 9 looks & 1 / 5 / 9 & 0.993--1.009 / 4.98 / 9.07 \\
Speckle grain width over resolution, uniform / Taylor & 1.00 / 1.25 & 0.994--1.006 / 1.244--1.259 \\
Geocoded position of point targets & true map position & mean error $\le 12.9$\,mm \\
\addlinespace[2pt]
\multicolumn{3}{@{}l}{\emph{Observed SAR}} \\
PSLR of point targets, ten Umbra scenes & $-13.26$\,dB (uniform) & $-12.1$ to $-13.3$\,dB \\
\bottomrule
\end{tabularx}
\end{table}

\subsection{Computational efficiency and modeling scope}
\label{sec:s-cost}

\synsar generates scene-scale SAR imagery by GPU ray tracing followed by image-domain formation, and ray tracing dominates the cost of SAR generation.
On one GPU of an NVIDIA GB200 system (189\,GB memory), SAR generation runs at approximately 200 small or 7 large worlds per occupied GPU-hour, including concurrent CPU-side processing.
Tracing takes a median of 17\,s for a small world and 369\,s for a large world, and reaches 2,735\,s for the most demanding, forest-dominated world.

\synsar targets geocoded VHR X-band SAR intensity imagery for image-based learning.
Scattering is evaluated once, at the carrier frequency and from the aperture center, for one co-polarized channel, HH or VV, per acquisition.
Cross-polarized channels (HV, VH), which arise mainly from depolarization in multiple and volume scattering, are not simulated.
The dataset therefore contains no dual-polarized pairs, and the correlation of their speckle and phase is not modeled; the HH--VV phase of traced returns is checked on canonical reflectors (\cref{tab:s-validation}).
The sensor follows a straight, horizontal track in a local Cartesian scene frame, and an ideal processor forms the image; acquisition modes differ only in sampled geometry, resolution, looks, and noise level, with the spectral window set by the provider's product.
Geocoding uses the exact simulated geometry and reference terrain, without additional orbit or terrain errors.
The released products are detected intensities: the phase of distributed returns is drawn at random, so the data do not support interferometry.

Distributed and vegetation scattering rely on semi-empirical parameterizations, in places beyond their published validity ranges.
Fresnel reflection along specular chains, attenuation by vegetation, and surface-reflected returns of thin members are modeled (\cref{sec:s-scattering}), whereas explicit edge diffraction, transmission of rays through solid materials, and higher-order coupling, such as multiple diffuse scattering or canopy--ground double bounces, are not.
Trees are always in leaf, and there is no snow.
These approximations leave differences from observed SAR in material-dependent backscatter and fine-scale image texture.
The resulting dataset is evaluated on real-world learning tasks in \cref{sec:s-transfer}.

\section{Datasets and experimental protocols}
\label{sec:s-protocols}

This section supplements Sec.~4 of the main paper with details of the datasets and experimental protocols.

\subsection{Real-world datasets and splits}
\label{sec:s-real-data}

\begin{table*}[t]
\caption{\textbf{Real datasets, their supervision, and their splits.} Numbers are images or tiles unless noted; height tiles measure $512\times512$ pixels at 0.5\,m.}
\label{tab:s-real-data}
\centering\small
\setlength{\tabcolsep}{5pt}
\begin{tabularx}{\textwidth}{@{}ll>{\raggedright\arraybackslash}X>{\raggedright\arraybackslash}Xccc@{}}
\toprule
Task & Dataset & SAR data & Supervision & Train & Val. & Test \\
\midrule
Land cover mapping & OEM-SAR~\cite{xia2025oemsar} & VHR SAR; $1024\times1024$ images & eight classes; pseudo labels from optical imagery (train), human labels (val., test) & 4,333 & 210 & 490 \\
& BRIGHT~\cite{chen2025bright} & post-event SAR; 11 disaster events & land cover pseudo labels from co-registered optical images; building annotations & 1,837 & --- & --- \\
\addlinespace[2pt]
Height & SN6~\cite{shermeyer2020spacenet6} & airborne X-band, VV; Rotterdam & heights above ground from airborne laser data~\cite{ahn_ahn4} & 8,461 & 4,963 & 3,575 \\
& AMS & Umbra and Capella X-band~\cite{umbra_opendata,capella_opendata}, HH and VV; Amsterdam & as for SN6 & 836 & 287 & 787 \\
\addlinespace[2pt]
Registration & BRIGHT~\cite{chen2025bright} & disaster scenes; $1024\times1024$ tiles & official co-registration (train); manual control points (val., test) & \begin{tabular}[t]{@{}c@{}}681 tiles\\(4 scenes)\end{tabular} & 4 scenes & 4 scenes \\
& SN9~\cite{hansch2024spacenet9} & evaluation only & manual control points & --- & --- & 2 scenes \\
\bottomrule
\end{tabularx}
\end{table*}

\paragraph{Land cover mapping.}
All models are tested on the 490 human-labeled test images of OEM-SAR~\cite{xia2025oemsar} (\cref{tab:s-real-data}).
Its 4,333 training images carry pseudo labels derived from optical imagery; they supervise real-only, fine-tuned, and mixed training, with nested random subsets for the label fractions (43, 217, and 1,083 images for 1\%, 5\%, and 25\%), whereas unsupervised domain adaptation (UDA) uses the same images without their labels.
The 210 human-labeled validation images train a separate real baseline.
The human labels are drawn on optical images, so human and pseudo labels alike lie at map positions, as \synsar labels do (\cref{sec:s-supervision}).
A further baseline trains on 1,837 post-event SAR tiles from 11 BRIGHT disaster events~\cite{chen2025bright}, with land cover pseudo labels predicted from the co-registered optical images and building labels from the dataset's annotations. Training ignores damaged buildings with a 50\,m margin, pixels with missing optical data, and pixels whose predictions change between pre- and post-event images.

\paragraph{Building height estimation.}
We use two X-band datasets whose heights above ground come from airborne laser data~\cite{ahn_ahn4}: SpaceNet~6 (SN6)~\cite{shermeyer2020spacenet6}, the VV channel of airborne imagery over Rotterdam, and AMS, six Umbra and Capella acquisitions over Amsterdam in HH and VV~\cite{umbra_opendata,capella_opendata}.
Building pixels follow the SN6 annotations and, for AMS, the 3DBAG footprints~\cite{peters2022bag3d}; each acquisition is aligned to these footprints by a single translation, so heights and footprints also lie at map positions.
Training, validation, and test tiles cover separate areas: SN6 keeps its official extents, and AMS is split into blocks of about 1\,km.
UDA uses the training tiles of both datasets without labels.
SN6 images each 256\,m ground cell in up to 24 passes, and its label-fraction subsets prioritize distinct cells before repeated observations (\cref{sec:s-limited}).

\paragraph{SAR--optical registration.}
Real-pair training uses the official co-registered SAR--optical pairs of BRIGHT disaster scenes, with the co-registration as ground truth; damaged buildings with a 10\,m margin and pixels without data are not supervised (\cref{tab:s-reg-data}).
Evaluation uses manual SAR--optical control points on six scenes, four from BRIGHT and two from SpaceNet~9 (SN9)~\cite{hansch2024spacenet9}, on which no model trains (\cref{tab:s-reg-scenes}).
The control points of the BRIGHT scenes were placed manually by three annotators experienced in SAR interpretation and cross-checked, with points added in further rounds for some scenes; SpaceNet~9 provides its own tie points.
Models that use real pairs train on the 681 tiles of four other BRIGHT scenes, whose co-registered rasters, including areas without data, measure between 13,312\,$\times$\,12,288 and 23,635\,$\times$\,23,552 pixels.
To study the amount of real supervision, we train on one, four, or eight BRIGHT scenes: the one-scene setting consists of four independent runs, one per scene of the standard training set, and the eight-scene setting adds four further BRIGHT scenes.
These four scenes serve as validation scenes; \cref{tab:s-reg-scenes} also reports them for the models that do not train on them.

\subsection{Training configurations}
\label{sec:s-training}

\begin{table*}[t]
\caption{\textbf{Training settings.} Two-stage settings (fine-tuning and the UDA students) split the step budget into a SynSAR stage and a real-data stage. The other matchers are described in the text.}
\label{tab:s-train}
\centering\small
\setlength{\tabcolsep}{4pt}
\begin{tabularx}{\textwidth}{@{}>{\raggedright\arraybackslash}p{0.105\textwidth}>{\raggedright\arraybackslash}X>{\raggedright\arraybackslash}X>{\raggedright\arraybackslash}X@{}}
\toprule
& DINOv3-L + DPT & U-Net & MINIMA-RoMa \\
\midrule
Tasks & land cover mapping, height & land cover mapping, height & registration \\
Weights & DINOv3-L/16~\cite{simeoni2025dinov3}, DPT head~\cite{ranftl2021dpt} & U-Net~\cite{ronneberger2015unet} with ImageNet~\cite{deng2009imagenet} EfficientNet-B4 encoder~\cite{tan2019efficientnet} & RoMa~\cite{edstedt2024roma} with MINIMA weights~\cite{ren2025minima}; DINOv2~\cite{oquab2024dinov2} features frozen \\
Input & 8-bit SAR scaled to [0, 1], repeated to three channels with ImageNet normalization; random $512\times512$ crops & 8-bit SAR scaled to [0, 1], one channel; random $512\times512$ crops & 8-bit SAR and optical crops of $560\times560$ pixels \\
Loss & cross-entropy (land cover); smooth L1 with $\beta = 1$\,m on valid pixels (height) & as for DINOv3 & RoMa's training losses on the correspondence field \\
Optimizer, learning rate & AdamW, weight decay 0.01; $5\times10^{-6}$ backbone, $5\times10^{-5}$ head & AdamW, no weight decay; $10^{-4}$ & AdamW, weight decay 0.01, gradient clipping; $2.5\times10^{-6}$ encoder, $5\times10^{-5}$ decoder \\
Schedule & warm-up over 5\% of the steps, then polynomial decay (power 0.9) to zero & as for DINOv3 & cosine decay to zero, no warm-up \\
Steps $\times$ batch & 16,000 $\times$ 8; fine-tuning and UDA: 8,000 + 8,000 & 50,000 $\times$ 8; UDA: 25,000 + 25,000 & 12,000 $\times$ 4 in every setting; fine-tuning: 6,000 + 6,000 \\
\bottomrule
\end{tabularx}
\end{table*}

\paragraph{Training strategies.}
Models are trained on SynSAR or on real data alone, fine-tuned on real data from a SynSAR model, or trained on both sources with equal sampling (\cref{tab:s-train}).
All settings of a model train for the same number of steps: 16,000 for DINOv3 and 50,000 for the U-Net in land cover mapping and height estimation, and 12,000 for every matcher in registration, including the one- and eight-scene settings and the simulation levels (\cref{sec:s-levels}).
Fine-tuning trains on SynSAR for the first half of the budget and continues on real data for the second half with a new schedule.
Mixed training samples the two sources equally, with one loss and no source weights: in land cover mapping and height estimation, four of the eight samples of each batch come from each source; in registration, each sample comes from either source with equal probability.
Models are used after their last step without selection on validation data.

\paragraph{Land cover mapping and building height estimation.}
Land cover mapping and height estimation train on the small worlds acquired in one spotlight mode with VV polarization at incidence angles between 22$^\circ$ and 50$^\circ$ (5th--95th percentiles: 24--50$^\circ$; \cref{tab:s-dataset}).
The other 2,596 small worlds comprise 107 worlds in the same mode that are held out to monitor training and 2,489 worlds with other acquisitions: 1,740 in other spotlight and stripmap modes, 133 with HH polarization, and 616 at incidence angles outside 22--50$^\circ$.
Images are drawn with probability proportional to their number of pixels, and images smaller than the crop are padded with an ignored label.
A synthetic amplitude $x$, normalized by its median, is converted to an 8-bit digital number as $\mathrm{DN} = \mathrm{clip}_{[0,255]}(38.9 + 61.9\ln x)$.
SAR-specific augmentation randomizes this conversion with one draw per image that stays fixed across epochs: the speckle is either smoothed, as by multi-looking, or strengthened by correlated speckle; the image is blurred with probability 0.3; the logarithmic mapping and its black point vary; and Gaussian noise of up to 3\,DN is added.
For land cover mapping, it is applied only to synthetic images, and real images receive flips and 90$^\circ$ rotations; the human-label baseline ignores unlabeled pixels.
For height estimation, it is applied to synthetic images and real tiles alike, and no flips or rotations are used as augmentation, because layover and shadow depend on the look direction.
SynSAR-only models use no other augmentation, and the settings without SAR-specific augmentation omit it.

\paragraph{Unsupervised domain adaptation.}
SynSAR + UDA uses the real training images without their labels, with the first-stage (8,000-step) DINOv3 SynSAR model of the same seed as teacher.
The DINOv3 student starts from the teacher and the U-Net student from a 25,000-step U-Net SynSAR model; both train on the teacher's pseudo labels without augmentation for the second half of their budget.
For land cover mapping, the teacher averages its softmax over the identity and three flips of each OEM-SAR training image, and a pixel keeps its predicted class only if the confidence reaches the median confidence of that class; all other pixels are ignored.
For height estimation, the teacher averages its predictions of the SN6 and AMS training tiles over the identity and three flips and uses the spread of the four predictions as uncertainty; within each predicted-height band (0--3, 3--10, 10--20, 20--40, and above 40\,m), a pixel is kept if its spread is at most the median of the band, and each batch draws equally from both cities.

\paragraph{SAR--optical registration.}
With \synsar and with the real pairs alike, the matchers are supervised by a known correspondence composed with a random warp: for \synsar, the correspondence derived from the scene geometry (\cref{sec:s-supervision}); for the real pairs, the official co-registration, which is based on manual control points.
Synthetic samples map the SAR image to 8 bits with a random logarithmic display mapping and warp the optical crop by a random shift (up to 48 pixels along each axis for half of the samples and up to 128 pixels for the others), a small random affine transform, and a smooth elastic field; the simulator's correspondence, composed with this warp, supervises valid pixels whose matched optical pixels have a map-grid height of at most 1\,m.
Real samples are windows of the official co-registered pairs, from which two independent random homographies with up to 45$^\circ$ of rotation crop the SAR and optical views, so that the homography between them is the ground truth.
For MINIMA-RoMa, \cref{tab:s-reg-data} also gives the four combinations of training data and warp: with the homographies, real pairs reach 65.5 \Sten and \synsar 68.3; with the \synsar warp, 58.6 and 69.6.
The other matchers (RoMa from its MegaDepth~\cite{li2018megadepth} weights, XoFTR~\cite{tuzcuoglu2024xoftr}, SuperPoint+LightGlue~\cite{detone2018superpoint,lindenberger2023lightglue}, and MatchAnything-RoMa~\cite{he2025matchanything}) train on the same samples with three seeds for 12,000 steps, with their own batch sizes and cosine schedules decaying to zero; MatchAnything-RoMa, which has no released training code, uses RoMa's losses and learning rates.
The three RoMa-based matchers train 111\,M parameters on frozen DINOv2 ViT-L/14 features (304\,M parameters); XoFTR trains 11\,M, and LightGlue trains 12\,M on frozen SuperPoint features (1.3\,M).
Every matcher is evaluated with its zero-shot settings.

\subsection{Evaluation metrics and statistical reporting}
\label{sec:s-metrics}

\paragraph{Land cover mapping.}
Each OEM-SAR test image is predicted whole, as in the official evaluation.
With $n_{ij}$ the number of test pixels of class $i$ predicted as class $j$, accumulated over all test images with unlabeled pixels ignored, we report the IoU of each class and its mean over the eight classes,
\begin{gather*}
\mathrm{IoU}_c=\frac{n_{cc}}{\sum_j n_{cj}+\sum_j n_{jc}-n_{cc}},\\
\mathrm{mIoU}_8=\frac{1}{8}\sum_{c=1}^{8}\mathrm{IoU}_c .
\end{gather*}
All land cover and height results use a single forward pass.

\paragraph{Building height estimation.}
Each test tile is predicted whole, and predictions are clamped at 0\,m.
With $\hat h_p$ and $h_p$ the predicted and reference heights and $\mathcal{B}$ the set of all building pixels of a test set, pooled rather than averaged over buildings or tiles,
\begin{equation*}
\mathrm{MAE}=\frac{1}{|\mathcal{B}|}\sum_{p\in\mathcal{B}}\bigl|\hat h_p-h_p\bigr|,\qquad
\mathrm{bias}=\frac{1}{|\mathcal{B}|}\sum_{p\in\mathcal{B}}\bigl(\hat h_p-h_p\bigr).
\end{equation*}
The height groups split these pixels by their reference height (0--3, 3--10, 10--20, 20--40, and above 40\,m; \cref{tab:s-height-groups}).
A constant reference predicts, at every pixel, the mean of the synthetic training height targets over their valid pixels (\cref{tab:s-mech-all}).

\paragraph{SAR--optical registration.}
Accuracy is measured at the manual control points.
For each point, the matcher runs on a SAR crop and an optical crop: the dense matchers sample correspondences from their predicted warp (4,096 for RoMa; the defaults of 10,000 for MINIMA-RoMa and 5,000 for MatchAnything-RoMa), SuperPoint+LightGlue matches up to 4,096 detected keypoints per image, XoFTR keeps the matches above its confidence threshold, and OSMNet~\cite{zhang2022osmnet} estimates one translation in each of 25 windows of the crop, each giving one correspondence.
All correspondences whose endpoints lie on valid data in both crops enter an affine RANSAC fit (3-pixel threshold), and an inlier-weighted local affine transform (Gaussian weights, $\sigma = 32$\,m) maps the SAR point to the optical image; where the weights of the inliers sum to less than four, the global affine transform is used instead.
Points with fewer than four RANSAC inliers or a fitted affine scale outside $[1/2, 2]$ are failures.
Each point is first matched on 512\,m crops at 1\,m spacing; the result comes from a pass on 256\,m crops at native resolution, for which the optical crop is re-centered on the first estimate if that estimate moves the point by more than 10\,m.
With $e_i$ the distance between the registered position of control point $i$ and its manual optical position, in native SAR pixels, and $\mathcal{P}_s$ the control points of scene $s$, the success rate at a threshold of $\tau$ pixels is
\begin{equation*}
\mathrm{S@}\tau=\frac{1}{6}\sum_{s=1}^{6}\frac{1}{|\mathcal{P}_s|}\sum_{i\in\mathcal{P}_s}\mathbf{1}[\,e_i\le\tau\,],
\end{equation*}
with $\tau=5$ and 10 for \Sfive and \Sten; S@5\,m is the same rate with $e_i$ in meters and a threshold of 5\,m.
Failures count as misses ($e_i=\infty$). The mean error (Err.) averages $e_i$ over the control points of a scene, with failures assigned their error without registration.
Scores are thus computed within each scene, averaged over the six evaluation scenes with equal weights, and then over seeds.

\begin{table}[t]
\caption{\textbf{Registration training data and supervision.} Scene and world counts are not directly comparable. Bottom: \Sten of MINIMA-RoMa on the six evaluation scenes (mean{\color{sdgray}\,$\pm$\,SD} over three seeds) for the four combinations of training data and geometric augmentation.}
\label{tab:s-reg-data}
\centering\footnotesize
\setlength{\tabcolsep}{3pt}
\begin{tabularx}{\linewidth}{@{}>{\raggedright\arraybackslash}p{0.23\linewidth}>{\raggedright\arraybackslash}X>{\raggedright\arraybackslash}X@{}}
\toprule
& Real pairs & SynSAR \\
\midrule
Source & four BRIGHT scenes & 191 synthetic worlds, one acquisition each \\
Extent & 681 tiles, 73.0\,km$^2$; 11.7--23.3\,km$^2$ per scene & 512\,m worlds, 50.1\,km$^2$; 0.26\,km$^2$ per world \\
Correspondence & official co-registration & dense, from the scene geometry \\
Supervised pixels & all pixels except damaged buildings (10\,m margin) and missing data & valid correspondences whose matched optical pixel has a map-grid height $\le$ 1\,m (38.1\% of SAR pixels) \\
Geometric augmentation & two random homographies, with rotation up to 45$^\circ$ & shift, affine, and elastic warp of the optical crop \\
Training steps & 12,000 & 12,000 \\
\bottomrule
\end{tabularx}

\vspace{6pt}

\setlength{\tabcolsep}{4pt}
\begin{tabular}{@{}lcc@{}}
\toprule
\Sten\,$\uparrow$ & Homographies & \synsar warp \\
\midrule
Real pairs & 65.5{\scriptsize\color{sdgray}$\,\pm$1.4} & 58.6{\scriptsize\color{sdgray}$\,\pm$0.6} \\
\synsar & 68.3{\scriptsize\color{sdgray}$\,\pm$1.3} & 69.6{\scriptsize\color{sdgray}$\,\pm$0.4} \\
\bottomrule
\end{tabular}

\end{table}

\paragraph{Statistical reporting.}
IoU values and success rates are given in percent. Trained settings use three seeds, and we report the mean and standard deviation over seeds; zero-shot matchers run once, and the four runs of the one-scene registration setting train on different scenes, so their spread also reflects the choice of scene.
Differences are computed from the unrounded means and given with 95\% intervals: Welch's two-sample $t$ interval between two trained settings, and a one-sample $t$ interval of the trained setting against a zero-shot matcher.
These intervals describe training variability on the evaluated images and scenes, not uncertainty over other scenes; no correction for multiple comparisons is applied.
Results reported without intervals are descriptive.

\section{Real-world transfer results}
\label{sec:s-transfer}

This section supplements Secs.~5.1--5.4 of the main paper with additional results and details.

\subsection{SAR--optical registration}
\label{sec:s-reg}

\begin{table*}[t]
\caption{\textbf{Registration with five matchers on the six evaluation scenes.} \Sten, \Sfive, the success rate within 5\,m (S@5\,m), and the mean error in native pixels (Err.); trained models: mean{\color{sdgray}\,$\pm$\,SD} over three seeds; zero-shot: released weights, one run. $\Delta$: SynSAR only minus Real only with its 95\% Welch interval {\color{sdgray}[in brackets]}. OSMNet: zero-shot only.}
\label{tab:s-reg-matchers}
\centering\footnotesize
\setlength{\tabcolsep}{3.4pt}
\begin{tabular}{@{}lccccccccccc@{}}
\toprule
& Zero-shot & \multicolumn{4}{c}{Real only} & \multicolumn{4}{c}{SynSAR only} & SynSAR $-$ real \\
\cmidrule(lr){2-2}\cmidrule(lr){3-6}\cmidrule(lr){7-10}\cmidrule(l){11-11}
Matcher & \Sten\,$\uparrow$ & \Sten\,$\uparrow$ & \Sfive\,$\uparrow$ & S@5\,m\,$\uparrow$ & Err.\,$\downarrow$ & \Sten\,$\uparrow$ & \Sfive\,$\uparrow$ & S@5\,m\,$\uparrow$ & Err.\,$\downarrow$ & $\Delta$\Sten\,$\uparrow$ \\
\midrule
MINIMA-RoMa & 51.7 & 65.5{\scriptsize\color{sdgray}$\,\pm$1.4} & 34.2 & 70.1 & 11.6 & 69.6{\scriptsize\color{sdgray}$\,\pm$0.4} & 40.1 & 73.0 & 11.0 & $+$4.0\,{\color{sdgray}[0.9, 7.2]} \\
MatchAnything-RoMa & 57.8 & 63.7{\scriptsize\color{sdgray}$\,\pm$1.0} & 32.7 & 70.6 & 11.2 & 68.7{\scriptsize\color{sdgray}$\,\pm$0.2} & 40.1 & 73.7 & 11.2 & $+$5.0\,{\color{sdgray}[2.8, 7.3]} \\
RoMa & 18.3 & 64.4{\scriptsize\color{sdgray}$\,\pm$0.2} & 32.4 & 68.8 & 11.7 & 67.9{\scriptsize\color{sdgray}$\,\pm$1.1} & 40.2 & 72.3 & 11.2 & $+$3.5\,{\color{sdgray}[0.9, 6.1]} \\
XoFTR & 9.5 & 18.7{\scriptsize\color{sdgray}$\,\pm$2.2} & 9.1 & 20.8 & 32.0 & 54.3{\scriptsize\color{sdgray}$\,\pm$2.0} & 32.0 & 56.8 & 27.7 & $+$35.6\,{\color{sdgray}[30.8, 40.5]} \\
SuperPoint+LightGlue & 11.9 & 21.2{\scriptsize\color{sdgray}$\,\pm$1.3} & 9.4 & 28.2 & 31.1 & 48.8{\scriptsize\color{sdgray}$\,\pm$1.9} & 25.4 & 55.6 & 22.3 & $+$27.5\,{\color{sdgray}[23.7, 31.4]} \\
OSMNet & 37.9 & --- & --- & --- & --- & --- & --- & --- & --- & --- \\
\bottomrule
\end{tabular}

\end{table*}

\begin{table*}[t]
\caption{\textbf{Registration per scene} (\Sten; mean of three seeds; zero-shot: one run). Top: the six evaluation scenes. Bottom: the four validation scenes, on which none of the listed models trains. All manual control points are scored; the last column gives their total and the unweighted mean of the scene scores. Pixel size: native SAR pixel. No registration: the products' georeference alone. Real only, fine-tuned, and mixed use the four real training scenes.}
\label{tab:s-reg-scenes}
\centering\small
\setlength{\tabcolsep}{3.9pt}
\begin{tabular}{@{}lccccccc@{}}
\toprule
Training & Hawaii & Calif. & Spain & Morocco & SN9-1 & SN9-2 & Mean \\
\midrule
\# Control points & 75 & 221 & 308 & 247 & 151 & 104 & 1,106 \\
Pixel size (m) & 0.35 & 0.96 & 0.35 & 0.41 & 0.41 & 0.41 & \\
\midrule
No registration, \Sten\,$\uparrow$ & 34.7 & 97.7 & 16.9 & 26.3 & 0.0 & 0.0 & 29.3 \\
\addlinespace[3pt]
\multicolumn{8}{@{}l}{\emph{MINIMA-RoMa, \Sten\,$\uparrow$}} \\
\quad Zero-shot & 53.3 & 14.5 & 56.5 & 82.2 & 61.6 & 42.3 & 51.7 \\
\quad SynSAR only & 61.8 & 97.6 & 53.6 & 80.0 & 70.2 & 54.2 & 69.6 \\
\quad Real only & 56.4 & 82.5 & 59.4 & 82.2 & 62.3 & 50.3 & 65.5 \\
\quad Fine-tuned & 58.2 & 90.6 & 58.4 & 82.6 & 64.7 & 50.3 & 67.5 \\
\quad Mixed & 65.3 & 93.7 & 58.1 & 83.1 & 69.8 & 52.9 & 70.5 \\
\addlinespace[3pt]
\multicolumn{8}{@{}l}{\emph{SynSAR only $-$ Real only, \Sten\,$\uparrow$}} \\
\quad MINIMA-RoMa & $+$5.3 & $+$15.1 & $-$5.8 & $-$2.2 & $+$7.9 & $+$3.8 & $+$4.0 \\
\quad MatchAnything-RoMa & $+$12.0 & $+$2.7 & $-$0.9 & $+$4.7 & $+$8.2 & $+$3.5 & $+$5.0 \\
\quad RoMa & $+$4.4 & $+$18.9 & $-$6.5 & $-$4.6 & $+$7.5 & $+$1.3 & $+$3.5 \\
\quad XoFTR & $+$36.4 & $+$84.9 & $+$34.2 & $+$19.7 & $+$20.3 & $+$18.3 & $+$35.6 \\
\quad SuperPoint+LightGlue & $+$28.9 & $+$46.3 & $+$31.6 & $+$31.8 & $+$12.4 & $+$14.1 & $+$27.5 \\
\bottomrule
\end{tabular}

\vspace{6pt}

\begin{tabular}{@{}lccccc@{}}
\toprule
Training & Turkey & Noto & Jamaica-1 & Jamaica-2 & Mean \\
\midrule
\# Control points & 124 & 64 & 28 & 68 & 284 \\
Pixel size (m) & 0.31 & 0.34 & 0.31 & 0.31 & \\
\midrule
No registration, \Sten\,$\uparrow$ & 37.1 & 1.6 & 0.0 & 0.0 & 9.7 \\
\addlinespace[3pt]
\multicolumn{6}{@{}l}{\emph{MINIMA-RoMa, \Sten\,$\uparrow$}} \\
\quad Zero-shot & 71.0 & 67.2 & 3.6 & 2.9 & 36.2 \\
\quad SynSAR only & 81.2 & 72.4 & 45.2 & 41.7 & 60.1 \\
\quad Real only & 75.5 & 69.3 & 31.0 & 23.0 & 49.7 \\
\quad Fine-tuned & 75.8 & 71.9 & 44.0 & 34.3 & 56.5 \\
\quad Mixed & 81.2 & 70.3 & 52.4 & 44.1 & 62.0 \\
\addlinespace[3pt]
\multicolumn{6}{@{}l}{\emph{SynSAR only $-$ Real only, \Sten\,$\uparrow$}} \\
\quad MINIMA-RoMa & $+$5.6 & $+$3.1 & $+$14.3 & $+$18.6 & $+$10.4 \\
\quad MatchAnything-RoMa & $+$5.6 & $+$10.9 & $+$9.5 & $+$1.0 & $+$6.8 \\
\quad RoMa & $+$9.9 & $+$1.0 & $+$16.7 & $+$15.2 & $+$10.7 \\
\quad XoFTR & $+$49.5 & $+$32.3 & $+$2.4 & $+$3.9 & $+$22.0 \\
\quad SuperPoint+LightGlue & $+$24.7 & $+$10.9 & $+$14.3 & $+$8.3 & $+$14.6 \\
\bottomrule
\end{tabular}

\end{table*}

\begin{figure*}[tp]
\centering
\includegraphics[width=\textwidth]{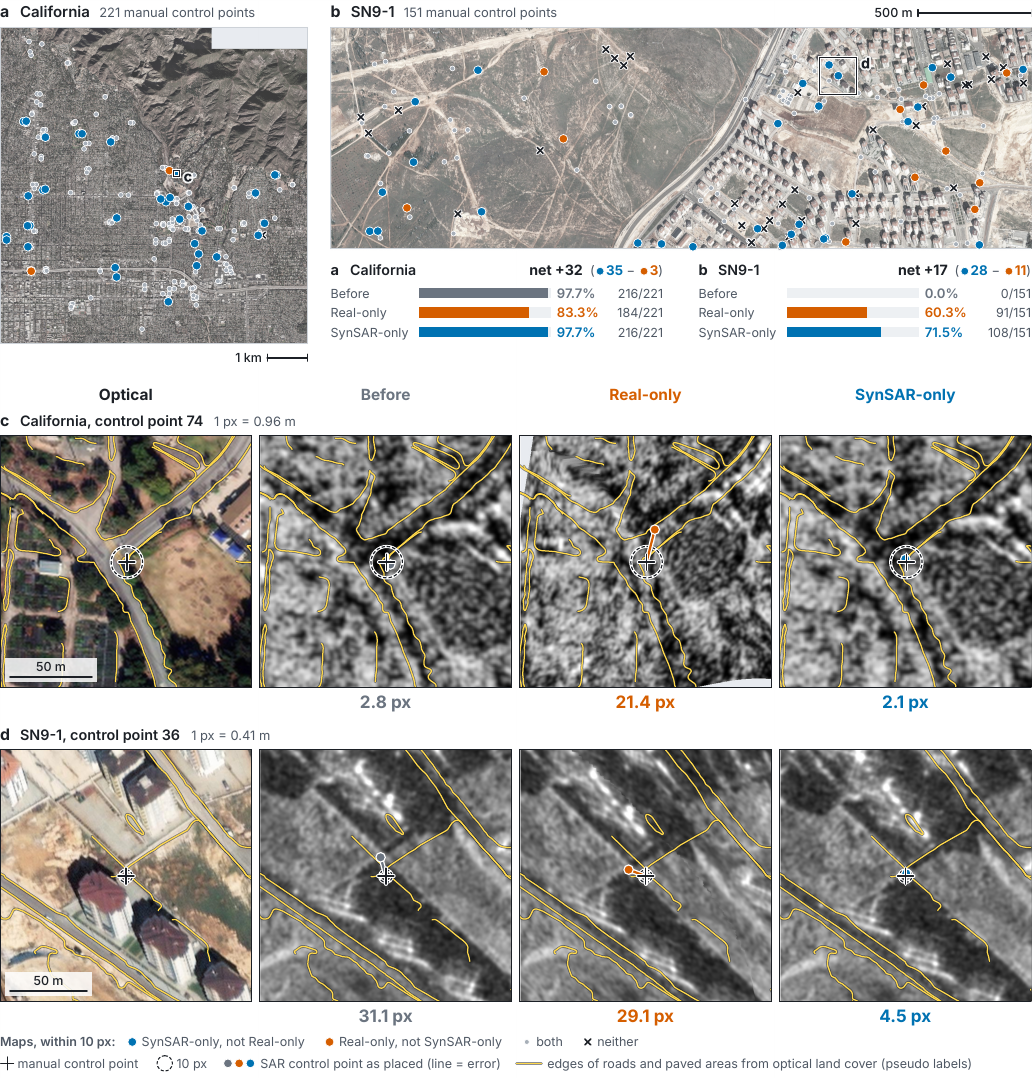}
\caption{\textbf{Real SAR--optical registration before and after matching} (MINIMA-RoMa, seed~1 of each model), on the two evaluation scenes with the largest gains; all six scenes are in \cref{tab:s-reg-scenes}.
(a,~b)~Manual control points, marked by the models whose error is within 10 native SAR pixels; the bars give the share of points within 10 pixels before registration (the products' georeference) and after each model.
(c,~d)~One control point per scene: the SAR image placed in the optical window by the georeference (Before) and by each model's local estimate, with the distance between the placed and the manual control point in native SAR pixels.
Yellow lines: road and paved-area edges from an optical land cover map, at the same positions in all panels, as a visual reference. The SAR images are smoothed with a 1\,m Gaussian for display; light gray: no image data.}
\label{fig:s-reg-qual}
\end{figure*}

\paragraph{Cross-matcher transfer.}
Each matcher is trained for 12,000 steps either on the real pairs or on SynSAR (\cref{tab:s-reg-data}); \cref{tab:s-reg-matchers} extends Tab.~3 of the main paper with \Sfive, S@5\,m, the mean error, and the interval of each difference.
The gain holds for all five architectures: SynSAR-only training raises \Sten, \Sfive, and S@5\,m for every matcher and lowers the mean error for all but MatchAnything-RoMa, whose error is unchanged (11.2 pixels).
It is 3.5--5.0 \Sten points for the three dense matchers and 27.5 and 35.6 points for SuperPoint+LightGlue and XoFTR, which reach only 21.2 and 18.7 with the real pairs.
After SynSAR training, RoMa, pretrained on optical pairs only, reaches the level of the two matchers with cross-modal pretraining (67.9 against 68.7 and 69.6).

\paragraph{Scene-level analysis.}
The gains vary across scenes (\cref{tab:s-reg-scenes}).
SynSAR-only training leads on four of the six scenes for MINIMA-RoMa and RoMa, on five for MatchAnything-RoMa, and on all six for XoFTR and SuperPoint+LightGlue; all three dense matchers trail on Spain ($-0.9$ to $-6.5$), and two of them on Morocco.
The gains of MINIMA-RoMa and RoMa are largest on California and SN9-1.
On California, where the georeference alone places 97.7\% of the control points within 10 pixels, MINIMA-RoMa keeps this alignment with SynSAR-only training (97.6), against 82.5 with real-only training.
On the four validation scenes, where the georeference places 9.7\% of the control points within 10 pixels on average, SynSAR-only training leads on every scene for all five matchers, by 6.8--22.0 points on average (\cref{tab:s-reg-scenes}, bottom).

\paragraph{Real pairs without building supervision.}
SynSAR supervises only correspondences whose matched optical pixels have a map-grid height of at most 1\,m (\cref{tab:s-reg-data}), whereas the real pairs supervise every valid pixel except damaged buildings.
To test whether this difference explains the gain, we also train real-only MINIMA-RoMa with all annotated buildings and a 10\,m margin excluded from the loss, for 12,000 steps with three seeds.
This only approximates the ground-level supervision of SynSAR, because the layover of tall buildings can reach beyond the margin, and it removes 32.6\% instead of 3.4\% of the valid area of the four training scenes, or about 27\% of the supervised pixels of the training windows.
The model reaches 62.5 \Sten, 3.1 points below real-only training ([$-$6.0, $-$0.1]), and SynSAR-only training stays 7.1 points ahead ([4.6, 9.6]).
Excluding buildings from the real supervision thus does not close the gap, although masking them also removes a substantial share of that supervision.

\paragraph{Qualitative results.}
\Cref{fig:s-reg-qual} compares the georeference with the two MINIMA-RoMa models (seed~1) on California and SN9-1.
With this seed, 35 control points of California are within 10 pixels only for the SynSAR-only model and 3 only for the real-only model; on SN9-1 the counts are 28 and 11.
Most of these points are near the threshold: where only one model is within 10 pixels, the other has a median error of 14.8 (California) and 13.8 pixels (SN9-1).
The local views show the two cases: on California, the real-only estimate shifts an intersection that the georeference had already aligned (21.4 pixels from the manual point, against 2.8 before registration and 2.1 for the SynSAR-only model); on SN9-1, the SynSAR-only estimate reduces the initial offset from 31.1 to 4.5 pixels, whereas the real-only estimate stays at 29.1.

\begin{table*}[t]
\caption{\textbf{Land cover mapping on OEM-SAR by class.} \miou (mean{\color{sdgray}\,$\pm$\,SD} over three seeds) and the IoU of each class on the 490 test images. Abbreviations: bare land (Bare), rangeland (Range), developed areas (Devel.), agriculture (Agri.), and building (Bldg.).}
\label{tab:s-landcover}
\centering\small
\setlength{\tabcolsep}{5pt}
\begin{tabular}{@{}lccccccccc@{}}
\toprule
& & \multicolumn{8}{c}{IoU per class\,$\uparrow$} \\
\cmidrule(l){3-10}
Training & \miou\,$\uparrow$ & Bare & Range & Devel. & Road & Tree & Water & Agri. & Bldg. \\
\midrule
\multicolumn{10}{@{}l}{\emph{DINOv3-L with DPT head}} \\
\quad SynSAR & 23.4{\scriptsize\color{sdgray}$\,\pm$0.2} & 0.0 & 16.4 & 20.5 & 20.9 & 38.0 & 28.5 & 29.0 & 33.8 \\
\quad SynSAR, no SAR-specific augmentation & 23.7{\scriptsize\color{sdgray}$\,\pm$0.4} & 0.2 & 16.4 & 21.6 & 20.1 & 39.1 & 31.9 & 28.4 & 32.3 \\
\quad SynSAR + UDA & 27.5{\scriptsize\color{sdgray}$\,\pm$1.9} & 3.6 & 15.4 & 25.6 & 21.9 & 40.4 & 42.4 & 39.6 & 31.0 \\
\addlinespace[2pt]
\quad BRIGHT (optical pseudo) & 31.4{\scriptsize\color{sdgray}$\,\pm$0.6} & 0.3 & 24.9 & 20.3 & 19.6 & 41.0 & 59.4 & 43.1 & 42.3 \\
\quad OEM-SAR train (optical pseudo) & 36.7{\scriptsize\color{sdgray}$\,\pm$0.7} & 0.0 & 26.6 & 26.4 & 25.9 & 41.6 & 70.8 & 58.7 & 43.6 \\
\quad OEM-SAR val (human) & 37.3{\scriptsize\color{sdgray}$\,\pm$1.0} & 10.4 & 25.4 & 25.4 & 26.1 & 46.7 & 60.3 & 62.0 & 42.3 \\
\addlinespace[4pt]
\multicolumn{10}{@{}l}{\emph{U-Net}} \\
\quad SynSAR & 20.2{\scriptsize\color{sdgray}$\,\pm$0.6} & 0.4 & 15.2 & 19.2 & 15.8 & 32.9 & 30.3 & 20.0 & 27.3 \\
\quad SynSAR, no SAR-specific augmentation & 19.5{\scriptsize\color{sdgray}$\,\pm$0.8} & 0.1 & 14.8 & 18.7 & 15.6 & 33.4 & 29.5 & 17.1 & 26.6 \\
\quad SynSAR + UDA & 26.9{\scriptsize\color{sdgray}$\,\pm$1.3} & 0.0 & 14.2 & 23.2 & 22.0 & 41.4 & 43.2 & 37.0 & 33.9 \\
\addlinespace[2pt]
\quad BRIGHT (optical pseudo) & 24.0{\scriptsize\color{sdgray}$\,\pm$0.7} & 1.5 & 18.6 & 17.1 & 15.9 & 33.0 & 36.8 & 33.1 & 35.8 \\
\quad OEM-SAR train (optical pseudo) & 33.8{\scriptsize\color{sdgray}$\,\pm$0.6} & 0.0 & 26.2 & 25.2 & 17.9 & 38.5 & 66.8 & 55.2 & 40.4 \\
\quad OEM-SAR val (human) & 33.7{\scriptsize\color{sdgray}$\,\pm$0.3} & 10.4 & 24.1 & 23.6 & 22.1 & 43.0 & 53.1 & 55.8 & 37.7 \\
\bottomrule
\end{tabular}

\end{table*}

\begin{figure*}[t]
\centering
\includegraphics[width=\textwidth]{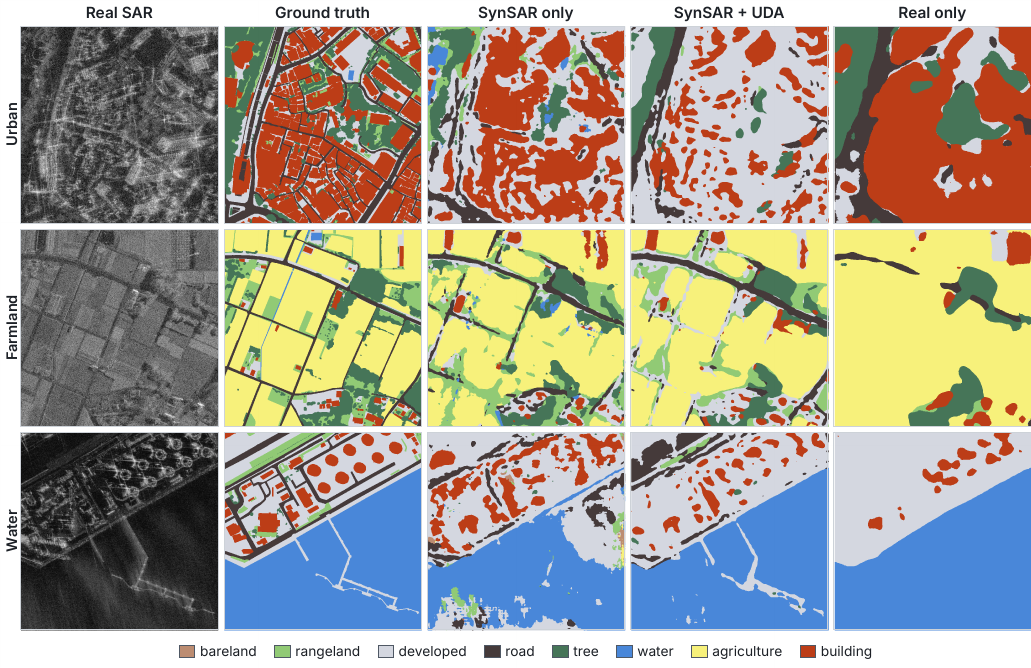}
\caption{\textbf{Land cover mapping on real OEM-SAR test images} (DINOv3, seed~1). Rows: an urban, a farmland, and a waterfront scene, each the test image whose SynSAR-only accuracy is closest to the median of its scene type. Ground truth: manual labels; SynSAR + UDA: self-training on unlabeled real SAR; Real only: OEM-SAR training images with optical pseudo labels.}
\label{fig:s-landcover-qual}
\end{figure*}

\begin{table*}[t]
\caption{\textbf{Building height estimation on SN6 and AMS.} Building MAE in m (mean{\color{sdgray}\,$\pm$\,SD} over three seeds) and the bias (mean signed error on building pixels; negative is underestimation). Real SAR, SN6 or AMS: trained on that city, hence in-domain on its own test set and cross-city on the other.}
\label{tab:s-height}
\centering\small
\setlength{\tabcolsep}{5pt}
\begin{tabular}{@{}lcccc@{}}
\toprule
& \multicolumn{2}{c}{SN6} & \multicolumn{2}{c}{AMS} \\
\cmidrule(lr){2-3}\cmidrule(l){4-5}
Training & MAE\,$\downarrow$ & Bias & MAE\,$\downarrow$ & Bias \\
\midrule
\multicolumn{5}{@{}l}{\emph{DINOv3-L with DPT head}} \\
\quad SynSAR & 7.73{\scriptsize\color{sdgray}$\,\pm$0.21} & $-$6.84 & 7.50{\scriptsize\color{sdgray}$\,\pm$0.09} & $-$5.69 \\
\quad SynSAR, no SAR-specific augmentation & 7.71{\scriptsize\color{sdgray}$\,\pm$0.07} & $-$6.93 & 7.50{\scriptsize\color{sdgray}$\,\pm$0.03} & $-$5.77 \\
\quad SynSAR + UDA & 7.51{\scriptsize\color{sdgray}$\,\pm$0.15} & $-$6.92 & 7.26{\scriptsize\color{sdgray}$\,\pm$0.11} & $-$6.01 \\
\addlinespace[2pt]
\quad Real SAR, SN6 & 6.66{\scriptsize\color{sdgray}$\,\pm$0.02} & $-$5.92 & 8.82{\scriptsize\color{sdgray}$\,\pm$0.02} & $-$7.87 \\
\quad Real SAR, AMS & 8.47{\scriptsize\color{sdgray}$\,\pm$0.11} & $-$7.42 & 5.38{\scriptsize\color{sdgray}$\,\pm$0.07} & $-$3.28 \\
\addlinespace[3pt]
\multicolumn{5}{@{}l}{\emph{U-Net}} \\
\quad SynSAR & 9.19{\scriptsize\color{sdgray}$\,\pm$0.07} & $-$8.76 & 8.83{\scriptsize\color{sdgray}$\,\pm$0.08} & $-$7.79 \\
\quad SynSAR, no SAR-specific augmentation & 8.91{\scriptsize\color{sdgray}$\,\pm$0.21} & $-$8.38 & 8.82{\scriptsize\color{sdgray}$\,\pm$0.24} & $-$7.76 \\
\quad SynSAR + UDA & 7.83{\scriptsize\color{sdgray}$\,\pm$0.13} & $-$7.42 & 7.52{\scriptsize\color{sdgray}$\,\pm$0.12} & $-$6.53 \\
\addlinespace[2pt]
\quad Real SAR, SN6 & 7.38{\scriptsize\color{sdgray}$\,\pm$0.04} & $-$6.75 & 9.62{\scriptsize\color{sdgray}$\,\pm$0.08} & $-$8.96 \\
\quad Real SAR, AMS & 9.21{\scriptsize\color{sdgray}$\,\pm$0.06} & $-$8.27 & 5.85{\scriptsize\color{sdgray}$\,\pm$0.02} & $-$3.88 \\
\bottomrule
\end{tabular}

\end{table*}

\begin{table*}[t]
\caption{\textbf{Building height errors by height range} (DINOv3, mean of three seeds). Building MAE and bias in m per range of true building height, with the share of building pixels in each range for the 9,251 SynSAR training worlds and for each test set. Cross-city and in-domain real: real SAR training in the other and in the same city.}
\label{tab:s-height-groups}
\centering\small
\setlength{\tabcolsep}{4.2pt}
\begin{tabular}{@{}lcccccccccc@{}}
\toprule
& \multicolumn{2}{c}{0--3\,m} & \multicolumn{2}{c}{3--10\,m} & \multicolumn{2}{c}{10--20\,m} & \multicolumn{2}{c}{20--40\,m} & \multicolumn{2}{c}{$>$40\,m} \\
\cmidrule(lr){2-3}\cmidrule(lr){4-5}\cmidrule(lr){6-7}\cmidrule(lr){8-9}\cmidrule(l){10-11}
Training & MAE\,$\downarrow$ & Bias & MAE\,$\downarrow$ & Bias & MAE\,$\downarrow$ & Bias & MAE\,$\downarrow$ & Bias & MAE\,$\downarrow$ & Bias \\
\midrule
\emph{SynSAR, share of building pixels} & \multicolumn{2}{c}{\emph{9.7\%}} & \multicolumn{2}{c}{\emph{61.7\%}} & \multicolumn{2}{c}{\emph{18.0\%}} & \multicolumn{2}{c}{\emph{7.5\%}} & \multicolumn{2}{c}{\emph{3.1\%}} \\
\addlinespace[4pt]
\emph{SN6, share of building pixels} & \multicolumn{2}{c}{\emph{5.2\%}} & \multicolumn{2}{c}{\emph{47.2\%}} & \multicolumn{2}{c}{\emph{36.4\%}} & \multicolumn{2}{c}{\emph{8.7\%}} & \multicolumn{2}{c}{\emph{2.6\%}} \\
\quad SynSAR & 2.58 & $+$0.71 & 3.97 & $-$2.77 & 7.44 & $-$7.07 & 17.25 & $-$17.11 & 59.49 & $-$59.49 \\
\quad SynSAR + UDA & 2.25 & $+$0.60 & 3.56 & $-$2.73 & 7.18 & $-$7.05 & 17.79 & $-$17.78 & 61.08 & $-$61.08 \\
\quad Cross-city real (AMS) & 2.74 & $+$1.47 & 4.07 & $-$2.53 & 8.26 & $-$7.99 & 20.83 & $-$20.82 & 62.50 & $-$62.50 \\
\quad In-domain real (SN6) & 2.03 & $+$0.01 & 2.97 & $-$1.86 & 5.97 & $-$5.69 & 17.21 & $-$17.13 & 58.31 & $-$58.31 \\
\addlinespace[4pt]
\emph{AMS, share of building pixels} & \multicolumn{2}{c}{\emph{8.5\%}} & \multicolumn{2}{c}{\emph{34.7\%}} & \multicolumn{2}{c}{\emph{48.2\%}} & \multicolumn{2}{c}{\emph{7.8\%}} & \multicolumn{2}{c}{\emph{0.8\%}} \\
\quad SynSAR & 5.02 & $+$4.27 & 4.51 & $-$2.08 & 7.86 & $-$7.52 & 17.61 & $-$17.50 & 44.78 & $-$44.78 \\
\quad SynSAR + UDA & 4.50 & $+$3.87 & 3.79 & $-$2.36 & 7.88 & $-$7.77 & 18.10 & $-$18.10 & 46.46 & $-$46.46 \\
\quad Cross-city real (SN6) & 3.61 & $+$2.78 & 4.27 & $-$3.17 & 10.29 & $-$10.24 & 21.56 & $-$21.56 & 50.67 & $-$50.67 \\
\quad In-domain real (AMS) & 5.61 & $+$4.99 & 3.40 & $-$1.03 & 4.67 & $-$3.91 & 14.61 & $-$14.51 & 42.61 & $-$42.61 \\
\bottomrule
\end{tabular}

\end{table*}

\begin{figure*}[t]
\centering
\includegraphics[width=\textwidth]{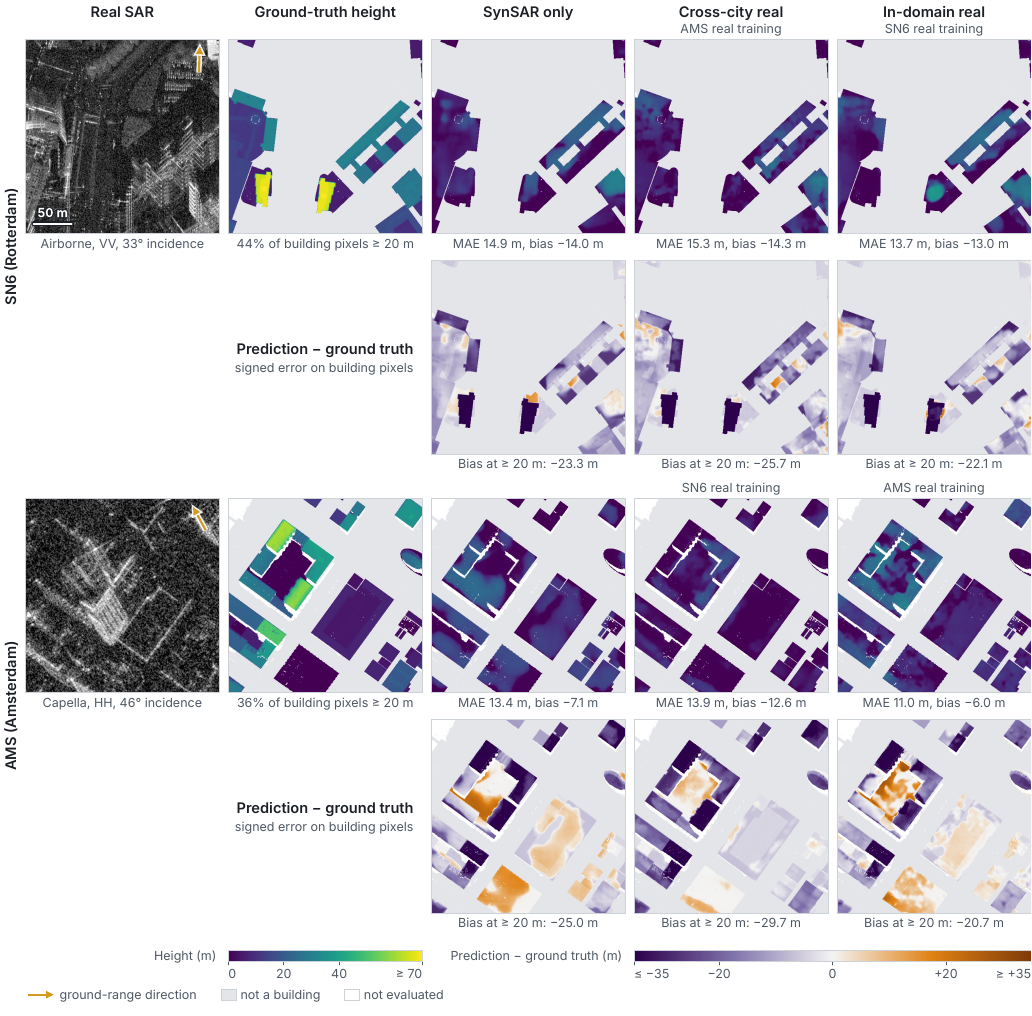}
\caption{\textbf{Building height estimation on real SAR} (DINOv3, seed~1). For SN6 and AMS, the test tile ($256\times256$\,m) whose SynSAR-only error is closest to the median of the tiles with a large share of buildings of at least 20\,m. Upper rows: the SAR image, the ground-truth height, and the predictions, with the building MAE and bias on the tile; cross-city and in-domain: real SAR training in the other and in the same city. Lower rows: prediction minus ground truth on building pixels, with the bias on buildings of at least 20\,m.}
\label{fig:s-height-qual}
\end{figure*}

\subsection{Land cover mapping}
\label{sec:s-landcover}

\paragraph{Overall transfer.}
SynSAR-only training transfers to real OEM-SAR images with both architectures (Tab.~4 of the main paper).
Self-training on unlabeled real SAR adds 4.1 \miou for DINOv3 ([$-$0.6, 8.8]) and 6.7 for the U-Net, whose students learn from the pseudo labels of the DINOv3 teacher.
The three DINOv3 models trained on real SAR in \cref{tab:s-landcover} remain stronger (31.4--37.3).

\paragraph{Class-wise behavior.}
With DINOv3, SynSAR-only training comes within 3.6--5.9 IoU points of OEM-SAR pseudo-label training on trees, roads, and developed areas and matches BRIGHT pseudo-label training on roads and developed areas (\cref{tab:s-landcover}).
The largest gaps are on agriculture (29.0 against 58.7) and water (28.5 against 70.8); self-training narrows them by 10.5 and 13.9 points but lowers building IoU by 2.8.
The two classes are confused with each other: SynSAR-only training assigns 24\% of the water pixels to agriculture and 15\% of the agriculture pixels to water, against 1\% each for OEM-SAR pseudo-label training, and self-training lowers these shares to 18\% and 9\%.
Both are homogeneous surfaces with low backscatter, and real fields and water bodies vary with crop type, growth stage, moisture, and wind, which the synthetic parcels may cover only in part.
Bare land is recognized only by the models trained on human labels (IoU 10.4); all others stay below 4.
SAR-specific augmentation has no consistent effect (23.4 with it and 23.7 without for DINOv3; 20.2 and 19.5 for the U-Net).

\paragraph{Qualitative results.}
In \cref{fig:s-landcover-qual}, the SynSAR-trained models delineate roads, field boundaries, and groups of buildings, whereas the model trained on real pseudo labels predicts a few large regions that fit the dominant classes better.
SynSAR-only training labels much of the dense urban block as developed areas and part of the sea as land; self-training restores the sea surface and the jetty but turns more urban buildings into developed areas.

\subsection{Building height estimation}
\label{sec:s-height}

\paragraph{Cross-city transfer.}
For DINOv3, SynSAR-only training is more accurate than real SAR training in the other city, by 0.74\,m [0.31, 1.17] on SN6 and 1.32\,m [1.12, 1.52] on AMS, while in-domain real training remains best (\cref{tab:s-height}).
Self-training lowers the error by 0.24\,m [0.004, 0.47] on AMS and by 0.22\,m [$-$0.21, 0.65] on SN6.
The U-Net trained on SynSAR alone is less accurate, on par with cross-city real training on SN6 (9.19 against 9.21\,m) and ahead of it on AMS (8.83 against 9.62\,m).
SAR-specific augmentation changes the DINOv3 error by at most 0.02\,m, and the U-Net is 0.28\,m more accurate without it on SN6.

\paragraph{Errors by building height.}
For every model, the bias is positive or zero below 3\,m and negative above, and above 20\,m the error is almost entirely underestimation (\cref{tab:s-height-groups}); buildings above 40\,m (2.6\% and 0.8\% of the building pixels) are underestimated by 43--63\,m, even with in-domain real training.
The cross-city advantage of SynSAR is concentrated in buildings of 10\,m and taller: on SN6 it grows to 3.6\,m at 20--40\,m, and on AMS SynSAR is 2.4--5.9\,m more accurate than SN6 training from 10\,m upward but less accurate below 10\,m.
Its predictions also follow the true height more closely than those of the cross-city models: from the lowest to the highest range, the mean prediction rises from 2.6 to 9.5\,m on SN6 and from 5.1 to 10.2\,m on AMS, against 3.4 to 6.5\,m and 3.6 to 4.4\,m for cross-city training and 1.9 to 10.6\,m and 5.8 to 12.4\,m for in-domain training.
Self-training trades the two ends: it lowers the error below 10\,m by 0.3--0.7\,m and raises it above 20\,m by 0.5--1.7\,m.

\paragraph{Qualitative results.}
On the tiles of \cref{fig:s-height-qual}, no model predicts more than 39\,m, so towers of up to 74 and 63\,m are underestimated whatever the training data.
On the AMS tile, the model trained on SN6 underestimates almost everywhere (bias $-12.6$\,m), whereas the SynSAR-only and in-domain models reach higher on the building wings but overestimate the low parts of the blocks, matching the positive low-building bias of \cref{tab:s-height-groups}.

\subsection{Learning with limited real supervision}
\label{sec:s-limited}

\begin{table}[t]
\caption{\textbf{Results with limited real-domain supervision} (DINOv3 and MINIMA-RoMa; mean{\color{sdgray}\,$\pm$\,SD} over three seeds). Real-domain supervision: share of the pseudo-labeled OEM-SAR training images (land cover mapping), share of the SN6 training tiles (height), or number of BRIGHT training scenes (registration). Fine-tuned: SynSAR model fine-tuned on the real data; mixed: half synthetic and half real samples. $^\dagger$Four runs, one per training scene, so the SD includes the choice of scene.}
\label{tab:s-limited}
\centering\footnotesize
\setlength{\tabcolsep}{4pt}
\begin{tabular}{@{}lccc@{}}
\toprule
Real-domain supervision & Real only & Fine-tuned & Mixed \\
\midrule
\multicolumn{4}{@{}l}{\emph{Land cover mapping, OEM-SAR: \miou\,$\uparrow$}} \\
\quad None (SynSAR only) & \multicolumn{3}{c}{\makebox[108pt]{{\color{sdgray}\leaders\hrule height 2.5pt depth -2.1pt\hfill}\kern5pt 23.4{\scriptsize\color{sdgray}$\,\pm$0.2}\kern5pt {\color{sdgray}\leaders\hrule height 2.5pt depth -2.1pt\hfill}}} \\
\quad 1\% & 16.8{\scriptsize\color{sdgray}$\,\pm$0.7} & 19.9{\scriptsize\color{sdgray}$\,\pm$1.4} & 23.6{\scriptsize\color{sdgray}$\,\pm$0.8} \\
\quad 5\% & 30.1{\scriptsize\color{sdgray}$\,\pm$0.4} & 30.9{\scriptsize\color{sdgray}$\,\pm$0.7} & 32.6{\scriptsize\color{sdgray}$\,\pm$0.5} \\
\quad 25\% & 34.9{\scriptsize\color{sdgray}$\,\pm$0.2} & 34.9{\scriptsize\color{sdgray}$\,\pm$0.3} & 34.7{\scriptsize\color{sdgray}$\,\pm$0.5} \\
\quad 100\% & 36.7{\scriptsize\color{sdgray}$\,\pm$0.7} & 36.3{\scriptsize\color{sdgray}$\,\pm$0.6} & 36.5{\scriptsize\color{sdgray}$\,\pm$0.2} \\
\addlinespace[3pt]
\multicolumn{4}{@{}l}{\emph{Building height, SN6: MAE (m)\,$\downarrow$}} \\
\quad None (SynSAR only) & \multicolumn{3}{c}{\makebox[108pt]{{\color{sdgray}\leaders\hrule height 2.5pt depth -2.1pt\hfill}\kern5pt 7.73{\scriptsize\color{sdgray}$\,\pm$0.21}\kern5pt {\color{sdgray}\leaders\hrule height 2.5pt depth -2.1pt\hfill}}} \\
\quad 1\% & 8.73{\scriptsize\color{sdgray}$\,\pm$0.05} & 7.78{\scriptsize\color{sdgray}$\,\pm$0.09} & 7.67{\scriptsize\color{sdgray}$\,\pm$0.01} \\
\quad 5\% & 7.32{\scriptsize\color{sdgray}$\,\pm$0.03} & 6.92{\scriptsize\color{sdgray}$\,\pm$0.05} & 6.80{\scriptsize\color{sdgray}$\,\pm$0.07} \\
\quad 25\% & 6.43{\scriptsize\color{sdgray}$\,\pm$0.04} & 6.32{\scriptsize\color{sdgray}$\,\pm$0.01} & 6.22{\scriptsize\color{sdgray}$\,\pm$0.03} \\
\quad 100\% & 6.66{\scriptsize\color{sdgray}$\,\pm$0.02} & 6.43{\scriptsize\color{sdgray}$\,\pm$0.06} & 6.24{\scriptsize\color{sdgray}$\,\pm$0.02} \\
\addlinespace[3pt]
\multicolumn{4}{@{}l}{\emph{Building height, AMS (cross-city): MAE (m)\,$\downarrow$}} \\
\quad None (SynSAR only) & \multicolumn{3}{c}{\makebox[108pt]{{\color{sdgray}\leaders\hrule height 2.5pt depth -2.1pt\hfill}\kern5pt 7.50{\scriptsize\color{sdgray}$\,\pm$0.09}\kern5pt {\color{sdgray}\leaders\hrule height 2.5pt depth -2.1pt\hfill}}} \\
\quad 1\% & 9.71{\scriptsize\color{sdgray}$\,\pm$0.02} & 8.45{\scriptsize\color{sdgray}$\,\pm$0.09} & 7.77{\scriptsize\color{sdgray}$\,\pm$0.08} \\
\quad 5\% & 9.00{\scriptsize\color{sdgray}$\,\pm$0.04} & 8.12{\scriptsize\color{sdgray}$\,\pm$0.01} & 7.44{\scriptsize\color{sdgray}$\,\pm$0.05} \\
\quad 25\% & 8.61{\scriptsize\color{sdgray}$\,\pm$0.03} & 8.14{\scriptsize\color{sdgray}$\,\pm$0.04} & 7.27{\scriptsize\color{sdgray}$\,\pm$0.06} \\
\quad 100\% & 8.82{\scriptsize\color{sdgray}$\,\pm$0.02} & 8.26{\scriptsize\color{sdgray}$\,\pm$0.08} & 7.31{\scriptsize\color{sdgray}$\,\pm$0.02} \\
\addlinespace[3pt]
\multicolumn{4}{@{}l}{\emph{Registration, six scenes: \Sten\,$\uparrow$}} \\
\quad None (SynSAR only) & \multicolumn{3}{c}{\makebox[108pt]{{\color{sdgray}\leaders\hrule height 2.5pt depth -2.1pt\hfill}\kern5pt 69.6{\scriptsize\color{sdgray}$\,\pm$0.4}\kern5pt {\color{sdgray}\leaders\hrule height 2.5pt depth -2.1pt\hfill}}} \\
\quad One scene$^\dagger$ & 55.4{\scriptsize\color{sdgray}$\,\pm$11.7} & 63.5{\scriptsize\color{sdgray}$\,\pm$6.4} & 68.4{\scriptsize\color{sdgray}$\,\pm$3.4} \\
\quad Four scenes & 65.5{\scriptsize\color{sdgray}$\,\pm$1.4} & 67.5{\scriptsize\color{sdgray}$\,\pm$0.5} & 70.5{\scriptsize\color{sdgray}$\,\pm$0.4} \\
\quad Eight scenes & 66.9{\scriptsize\color{sdgray}$\,\pm$0.9} & 69.9{\scriptsize\color{sdgray}$\,\pm$1.0} & 69.3{\scriptsize\color{sdgray}$\,\pm$0.6} \\
\bottomrule
\end{tabular}

\end{table}

\begin{table}[t]
\caption{\textbf{Gains of fine-tuning and mixing over real-only training} with the same real-domain data: difference of means with its 95\% Welch interval {\color{sdgray}[in brackets]}; for height, the reduction of building MAE (positive is better). $^\dagger$Four runs per setting, one per training scene, so the interval includes the choice of scene.}
\label{tab:s-limited-cmp}
\centering\footnotesize
\setlength{\tabcolsep}{3pt}
\begin{tabular}{@{}lcc@{}}
\toprule
Real-domain supervision & Fine-tuned vs.\ Real only & Mixed vs.\ Real only \\
\midrule
\multicolumn{3}{@{}l}{\emph{Land cover mapping, OEM-SAR: \miou gain\,$\uparrow$}} \\
\quad 1\% & $+$3.1\,{\color{sdgray}[0.3, 5.9]} & $+$6.8\,{\color{sdgray}[5.1, 8.6]} \\
\quad 5\% & $+$0.8\,{\color{sdgray}[$-$0.6, 2.3]} & $+$2.5\,{\color{sdgray}[1.5, 3.5]} \\
\quad 25\% & 0.0\,{\color{sdgray}[$-$0.7, 0.6]} & $-$0.3\,{\color{sdgray}[$-$1.3, 0.8]} \\
\quad 100\% & $-$0.4\,{\color{sdgray}[$-$1.9, 1.2]} & $-$0.2\,{\color{sdgray}[$-$1.9, 1.5]} \\
\addlinespace[3pt]
\multicolumn{3}{@{}l}{\emph{Building height, SN6: MAE reduction (m)\,$\uparrow$}} \\
\quad 1\% & $+$0.95\,{\color{sdgray}[0.76, 1.14]} & $+$1.05\,{\color{sdgray}[0.94, 1.17]} \\
\quad 5\% & $+$0.40\,{\color{sdgray}[0.31, 0.50]} & $+$0.53\,{\color{sdgray}[0.37, 0.69]} \\
\quad 25\% & $+$0.11\,{\color{sdgray}[0.02, 0.20]} & $+$0.21\,{\color{sdgray}[0.12, 0.29]} \\
\quad 100\% & $+$0.23\,{\color{sdgray}[0.09, 0.37]} & $+$0.42\,{\color{sdgray}[0.37, 0.47]} \\
\addlinespace[3pt]
\multicolumn{3}{@{}l}{\emph{Building height, AMS (cross-city): MAE reduction (m)\,$\uparrow$}} \\
\quad 1\% & $+$1.26\,{\color{sdgray}[1.04, 1.47]} & $+$1.94\,{\color{sdgray}[1.76, 2.11]} \\
\quad 5\% & $+$0.88\,{\color{sdgray}[0.80, 0.96]} & $+$1.56\,{\color{sdgray}[1.46, 1.66]} \\
\quad 25\% & $+$0.47\,{\color{sdgray}[0.39, 0.54]} & $+$1.34\,{\color{sdgray}[1.22, 1.46]} \\
\quad 100\% & $+$0.56\,{\color{sdgray}[0.38, 0.74]} & $+$1.51\,{\color{sdgray}[1.46, 1.56]} \\
\addlinespace[3pt]
\multicolumn{3}{@{}l}{\emph{Registration, six scenes: \Sten gain\,$\uparrow$}} \\
\quad One scene$^\dagger$ & $+$8.0\,{\color{sdgray}[$-$9.5, 25.6]} & $+$13.0\,{\color{sdgray}[$-$5.0, 30.9]} \\
\quad Four scenes & $+$2.0\,{\color{sdgray}[$-$1.1, 5.1]} & $+$5.0\,{\color{sdgray}[1.8, 8.1]} \\
\quad Eight scenes & $+$3.0\,{\color{sdgray}[0.7, 5.2]} & $+$2.3\,{\color{sdgray}[0.4, 4.3]} \\
\bottomrule
\end{tabular}

\end{table}

\Cref{tab:s-limited,tab:s-limited-cmp} give the results behind Fig.~4 of the main paper.

\paragraph{Land cover mapping.}
The benefit of SynSAR is largest when real-domain supervision is scarce.
Mixed training improves \miou by 6.8 points with 1\% of the pseudo-labeled OEM-SAR training images and by 2.5 with 5\%; with 25\% or more, both fine-tuning and mixing stay within 0.4 points of real-only training.
Fine-tuning follows the same trend with smaller gains ($+3.1$ at 1\% and $+0.8$ at 5\%).
The fractions count images with pseudo labels derived from optical imagery, not manual annotations.

\paragraph{Building height estimation.}
Mixed training reduces building MAE at every fraction of the SN6 training tiles on both test sets, by 0.21--1.05\,m on SN6 and 1.34--1.94\,m on cross-city AMS, including 0.42 and 1.51\,m with all SN6 tiles.
Fine-tuning also helps at every fraction, most with 1\% and 5\%; at 25\% and 100\% its gains are 0.11 and 0.23\,m on SN6 and 0.47--0.56\,m on AMS.
The fractions count tiles rather than ground area: the 5\% subset covers 58\% of the 724 SN6 ground cells and the 25\% subset all of them (\cref{sec:s-real-data}), so larger fractions mainly add repeated SAR views of the same buildings, and the results are not monotonic beyond 25\% (e.g., 6.43 against 6.66\,m for real-only training on SN6).

\begin{figure*}[t]
\centering
\includegraphics[width=\textwidth]{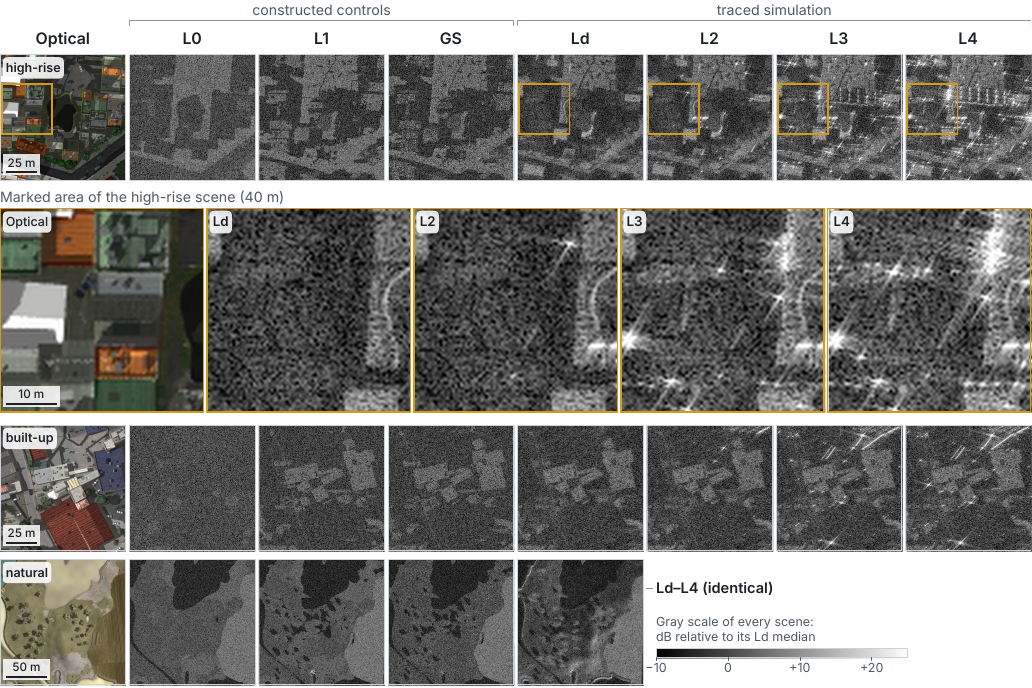}
\caption{\textbf{SAR appearance across simulation levels.} Two built-up worlds, one with high-rise buildings, and a natural world; all levels of a world share the scene and the acquisition. Built-up worlds are cropped to 100\,m; the natural world is shown whole (154\,m). Each world uses one dB display range relative to its median intensity at Ld ($-10$ to $+25$\,dB). Second row: the marked 40\,m region (optical and Ld--L4). For the natural world, Ld--L4 are the same image.}
\label{fig:s-levels}
\end{figure*}

\begin{table}[!tb]
\caption{\textbf{Controlled SAR simulation levels.} L0, L1, and GS are built from the class statistics of Ld; higher-order thin-cylinder contributions are already included at L2. Scene content and dense-prediction supervision are the same at all levels; for registration, L1--L4 share the same correspondence supervision and L0 uses map-geometry correspondences.}
\label{tab:s-levels}
\centering\footnotesize
\setlength{\tabcolsep}{3.5pt}
\begin{tabularx}{\linewidth}{@{}llX@{}}
\toprule
Level & Geometry & SAR image \\
\midrule
\multicolumn{3}{@{}l}{\emph{Constructed controls}} \\
\quad L0 & map & class-mean intensity $\times$ speckle \\
\quad L1 & radar & dominant-class mean intensity $\times$ speckle \\
\quad GS & radar & L1 class map, intensities randomized within classes \\
\addlinespace[2pt]
\multicolumn{3}{@{}l}{\emph{Traced simulation}} \\
\quad Ld & radar & single bounce, distributed scattering only \\
\quad L2 & radar & $+$ coherent facet and thin-cylinder returns \\
\quad L3 & radar & $+$ second-bounce surface reflections \\
\quad L4 & radar & up to five bounces (full \synsar) \\
\bottomrule
\end{tabularx}
\end{table}

\begin{table}[!tb]
\caption{\textbf{Concentration of simulated SAR returns.} Medians over the 1,732 high-rise worlds (a building of at least 30\,m) and the 6,848 other worlds rendered at every level. Power share: fraction of the total image power in the brightest 0.1\% of pixels. Power change: total image power relative to the preceding level.}
\label{tab:s-mech-images}
\centering\footnotesize
\setlength{\tabcolsep}{6pt}
\begin{tabular}{@{}lcccc@{}}
\toprule
& \multicolumn{2}{c}{Power share (\%)} & \multicolumn{2}{c}{Power change (dB)} \\
\cmidrule(lr){2-3}\cmidrule(l){4-5}
Level & High-rise & Other & High-rise & Other \\
\midrule
Ld & 2.1 & 1.9 & --- & --- \\
L2 & 10.7 & 3.9 & $+$0.83 & $+$0.18 \\
L3 & 29.6 & 10.4 & $+$3.32 & $+$0.41 \\
L4 & 40.8 & 14.0 & $+$2.55 & $+$0.09 \\
\bottomrule
\end{tabular}

\end{table}

\begin{table*}[t]
\caption{\textbf{Real-world transfer at every simulation level} (DINOv3 and MINIMA-RoMa; mean{\color{sdgray}\,$\pm$\,SD} over three seeds). Land cover mapping on OEM-SAR. The L4 models are the SynSAR-only models of the main paper. Constant height: the mean training height predicted everywhere.}
\label{tab:s-mech-all}
\centering\small
\setlength{\tabcolsep}{4.5pt}
\begin{tabular}{@{}lcccc@{}}
\toprule
& Land cover mapping & \multicolumn{2}{c}{Building MAE (m)\,$\downarrow$} & Registration (six scenes) \\
\cmidrule(lr){2-2}\cmidrule(lr){3-4}\cmidrule(l){5-5}
Level & \miou\,$\uparrow$ & SN6 & AMS & \Sten\,$\uparrow$ \\
\midrule
L0 & 12.0{\scriptsize\color{sdgray}$\,\pm$0.3} & 11.33{\scriptsize\color{sdgray}$\,\pm$0.13} & 10.93{\scriptsize\color{sdgray}$\,\pm$0.08} & 66.2{\scriptsize\color{sdgray}$\,\pm$0.2} \\
L1 & 17.5{\scriptsize\color{sdgray}$\,\pm$0.6} & 10.64{\scriptsize\color{sdgray}$\,\pm$0.09} & 10.13{\scriptsize\color{sdgray}$\,\pm$0.09} & 68.9{\scriptsize\color{sdgray}$\,\pm$0.5} \\
GS & 18.5{\scriptsize\color{sdgray}$\,\pm$0.5} & 10.36{\scriptsize\color{sdgray}$\,\pm$0.17} & 9.84{\scriptsize\color{sdgray}$\,\pm$0.11} & 68.8{\scriptsize\color{sdgray}$\,\pm$0.3} \\
\addlinespace[2pt]
Ld & 19.1{\scriptsize\color{sdgray}$\,\pm$0.7} & 9.33{\scriptsize\color{sdgray}$\,\pm$0.14} & 9.32{\scriptsize\color{sdgray}$\,\pm$0.12} & 68.8{\scriptsize\color{sdgray}$\,\pm$0.7} \\
L2 & 21.0{\scriptsize\color{sdgray}$\,\pm$0.4} & 8.26{\scriptsize\color{sdgray}$\,\pm$0.05} & 8.25{\scriptsize\color{sdgray}$\,\pm$0.09} & 68.8{\scriptsize\color{sdgray}$\,\pm$0.3} \\
L3 & 23.4{\scriptsize\color{sdgray}$\,\pm$0.4} & 7.77{\scriptsize\color{sdgray}$\,\pm$0.12} & 7.49{\scriptsize\color{sdgray}$\,\pm$0.10} & 69.1{\scriptsize\color{sdgray}$\,\pm$0.6} \\
L4 & 23.4{\scriptsize\color{sdgray}$\,\pm$0.2} & 7.73{\scriptsize\color{sdgray}$\,\pm$0.21} & 7.50{\scriptsize\color{sdgray}$\,\pm$0.09} & 69.6{\scriptsize\color{sdgray}$\,\pm$0.4} \\
\midrule
Constant height & --- & 9.30 & 9.09 & --- \\
\bottomrule
\end{tabular}

\end{table*}

\begin{table*}[t]
\caption{\textbf{Comparisons between simulation levels.} Difference of seed means with its 95\% Welch interval {\color{sdgray}[in brackets]}; positive is better. For registration (six scenes), L1 vs.\ L0 also changes the correspondence supervision. The first three comparisons correspond to the brackets in Fig.~5 of the main paper.}
\label{tab:s-mech-steps}\label{tab:s-register-c}
\centering\small
\setlength{\tabcolsep}{4pt}
\begin{tabular}{@{}llcccc@{}}
\toprule
& & Land cover mapping & \multicolumn{2}{c}{Building MAE reduction (m)\,$\uparrow$} & Registration \\
\cmidrule(lr){3-3}\cmidrule(lr){4-5}\cmidrule(l){6-6}
Comparison & Added & $\Delta$\miou\,$\uparrow$ & SN6 & AMS & $\Delta$\Sten\,$\uparrow$ \\
\midrule
L1 vs.\ L0 & radar geometry & $+$5.5\,{\color{sdgray}[4.3, 6.8]} & $+$0.70\,{\color{sdgray}[0.43, 0.96]} & $+$0.80\,{\color{sdgray}[0.60, 1.00]} & $+$2.6\,{\color{sdgray}[1.5, 3.8]} \\
L4 vs.\ L1 & simulated scattering & $+$5.9\,{\color{sdgray}[4.6, 7.2]} & $+$2.91\,{\color{sdgray}[2.47, 3.35]} & $+$2.62\,{\color{sdgray}[2.43, 2.82]} & $+$0.7\,{\color{sdgray}[$-$0.5, 1.8]} \\
L4 vs.\ L2 & multiple bounces & $+$2.4\,{\color{sdgray}[1.6, 3.2]} & $+$0.53\,{\color{sdgray}[0.05, 1.02]} & $+$0.75\,{\color{sdgray}[0.55, 0.94]} & $+$0.8\,{\color{sdgray}[$-$0.1, 1.7]} \\
\addlinespace[2pt]
L2 vs.\ Ld & coherent structural returns & $+$1.9\,{\color{sdgray}[0.6, 3.3]} & $+$1.07\,{\color{sdgray}[0.77, 1.37]} & $+$1.07\,{\color{sdgray}[0.83, 1.31]} & 0.0\,{\color{sdgray}[$-$1.5, 1.4]} \\
\bottomrule
\end{tabular}

\end{table*}

\begin{table}[t]
\caption{\textbf{Per-class IoU on OEM-SAR} at every simulation level (mean of three seeds).}
\label{tab:s-mech-classes}
\centering\footnotesize
\setlength{\tabcolsep}{2.6pt}
\begin{tabular}{@{}lcccccccc@{}}
\toprule
& \multicolumn{8}{c}{IoU per class\,$\uparrow$} \\
\cmidrule(l){2-9}
Level & Bare & Range & Devel. & Road & Tree & Water & Agri. & Bldg. \\
\midrule
L0 & 0.0 & 8.8 & 17.6 & 19.4 & 11.2 & 11.7 & 14.3 & 13.0 \\
L1 & 0.2 & 12.0 & 21.1 & 18.2 & 19.8 & 28.0 & 21.1 & 19.9 \\
GS & 1.1 & 10.2 & 21.5 & 18.0 & 17.5 & 33.6 & 23.9 & 22.4 \\
\addlinespace[2pt]
Ld & 0.0 & 10.4 & 21.6 & 18.3 & 30.4 & 23.7 & 27.5 & 20.7 \\
L2 & 0.0 & 12.0 & 21.7 & 18.4 & 33.6 & 25.0 & 28.6 & 28.7 \\
L3 & 0.1 & 15.9 & 21.3 & 20.4 & 38.0 & 29.7 & 29.9 & 32.2 \\
L4 & 0.0 & 16.4 & 20.5 & 20.9 & 38.0 & 28.5 & 29.0 & 33.8 \\
\bottomrule
\end{tabular}

\end{table}

\paragraph{SAR--optical registration.}
With one real training scene, real-only training depends strongly on which scene is used (38.4--63.4 \Sten over the four runs, one per scene), whereas fine-tuning and mixing reach 63.5 and 68.4 on average (54.4--69.5 and 63.4--70.8 over the four runs) and exceed real-only training with each of the four scenes.
Over real-only training, fine-tuning and mixing gain $+2.0$ [$-$1.1, 5.1] and $+5.0$ [1.8, 8.1] with four scenes, and $+3.0$ [0.7, 5.2] and $+2.3$ [0.4, 4.3] with eight.
From one to four to eight scenes, real-only training rises by 10.1 and then 1.4 points, whereas mixing varies by 2.1, so with SynSAR the result depends much less on the number and choice of real training scenes.
With eight scenes, real-only training (66.9) remains below SynSAR only (69.6; $+2.6$ [0.6, 4.6]).

\section{Controlled study of simulation mechanisms}
\label{sec:s-mech}

This section supplements Sec.~5.5 of the main paper, which compares seven SAR simulation levels on a common set of procedural worlds, with additional results and details.

\subsection{Controlled simulation levels}
\label{sec:s-levels}

\paragraph{Controlled worlds.}
Each task uses its SynSAR training set (\cref{sec:s-training}): the 9,251 small worlds for land cover mapping and height estimation, each observed once, and the 191 large training worlds for registration.
For dense prediction, all levels share the scene, the land cover and height targets, the validity masks, the training crops, and the augmentation draws, so that only the SAR image changes.
Models are trained with the SynSAR-only settings of each task and three seeds (\cref{sec:s-training}).
The images of all seven levels are part of the release.

\paragraph{Simulation levels.}
\Cref{tab:s-levels} summarizes the seven levels.
L0, L1, and GS are constructed controls rather than renderings, built from the class statistics of the Ld image.
L0 uses class-mean intensities in map geometry and therefore has no layover, foreshortening, or shadow.
L1 introduces the radar projection: it uses the dominant radar-return class of each pixel, including a separate no-return class, and assigns the mean intensity of that class.
GS retains this radar-geometry class map and the intensity distribution of each class, while randomizing the intensity values within each class.
The traced levels comprise distributed scattering only (Ld), then add coherent structural returns from facets and thin cylinders (L2), second-bounce surface reflections (L3), and up to five bounces (L4, the full simulator).
Speckle is present at every level, so the step from Ld to L2 adds these structural returns rather than coherence as such.
This progression lets us examine how imaging geometry, scattering, and multiple reflections affect real-world transfer.

\paragraph{Scope of controlled comparisons.}
For registration, L1 through L4, including GS, share the optical image and the correspondence supervision derived from the scene geometry (\cref{sec:s-supervision}), so that only the SAR image changes.
L0, an image in map geometry, is instead supervised with the map-to-optical correspondences consistent with its own geometry, on 95.8\% as many pixels.
The step from L0 to L1 therefore introduces radar geometry together with its aligned supervision, rather than changing image appearance alone.
Of the 9,251 small worlds, 671 contain no buildings and predominantly natural land cover; in these worlds, the levels Ld through L4 share one image, so the differences among these levels arise from the other 8,580 worlds.

\subsection{SAR appearance across simulation levels}
\label{sec:s-appearance}

\Cref{fig:s-levels} illustrates how the simulation levels change SAR appearance, and \cref{tab:s-mech-images} quantifies the concentration of backscattered power over the 8,580 worlds rendered at every level.
Coherent structural returns introduce localized bright features: in the 1,732 worlds that contain a building at least 30\,m tall, the brightest 0.1\% of pixels hold a median of 2.1\% of the image power at Ld and 10.7\% at L2.
With additional bounces, this share reaches 40.8\% at L4, compared with 14.0\% in the other worlds.

Multiple bounces also raise the total power, primarily in high-rise worlds: the median increases from L2 to L3 and from L3 to L4 are 3.32 and 2.55\,dB there, compared with 0.41 and 0.09\,dB elsewhere.
Coherent structural returns and multiple reflections thus alter the spatial concentration of simulated backscatter, most strongly in worlds with tall buildings.
\Cref{sec:s-twin} compares the levels with observed SAR on digital twins (\cref{tab:s-twin-levels}).

\subsection{Real-world transfer across simulation levels}
\label{sec:s-mech-transfer}

\Cref{tab:s-mech-all} reports real-world transfer at the seven simulation levels, and \cref{tab:s-mech-steps} gives selected comparisons with their 95\% intervals.
Moving from L0 to L1 improves the evaluated scores in all three tasks, whereas the differences between GS and L1 are small (1.0 \miou, 0.28 and 0.29\,m in building MAE, and 0.1 \Sten).
For registration, L0 is already on par with real-only training (66.2 against 65.5 \Sten in \cref{tab:s-reg-matchers}; $+0.7$ [$-$2.7, 4.1]), and the step to L1, where image geometry and correspondence supervision both change, adds 2.6 \Sten [1.5, 3.8].

From L1 to L4, land cover mapping improves by 5.9 \miou and building MAE decreases by 2.91 and 2.62\,m on SN6 and AMS, while registration changes by $+0.7$ \Sten [$-$0.5, 1.8].
Within this range, the step from Ld to L2 reduces building MAE by 1.07\,m on both test sets and improves land cover mapping by 1.9 \miou [0.6, 3.3], while the registration difference is 0.0 \Sten [$-$1.5, 1.4] (\cref{tab:s-mech-steps}).
From L2 onward, building MAE is below that of a constant prediction of the mean training height (\cref{tab:s-mech-all}).
Class-wise results also vary across the simulation levels (\cref{tab:s-mech-classes}).

\begin{figure}[!t]
\centering
\includegraphics[width=\linewidth]{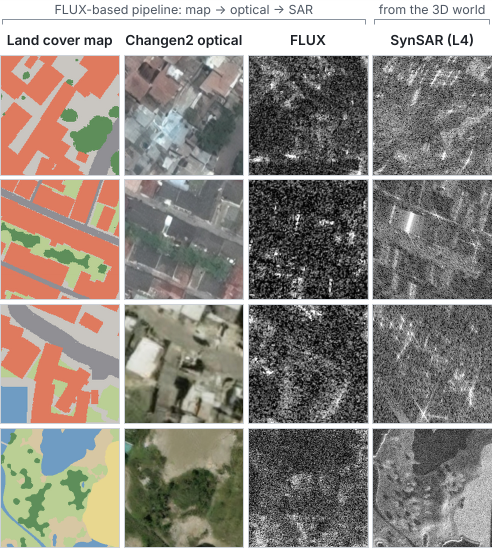}
\caption{\textbf{FLUX-based translation on synthetic worlds.} Four training worlds: Changen2 converts the land cover map into an optical-like image, which FLUX translates to SAR, while \synsar (L4) renders SAR from the 3D world (both used for training). The built-up worlds are cropped to 76\,m; the natural world is shown whole (154\,m). Each SAR image is stretched between its own 1st and 99.7th percentiles in dB.}
\label{fig:s-flux}
\end{figure}

\begin{table*}[t]
\caption{\textbf{Real-world transfer of the FLUX-based pipeline and of \synsar} (DINOv3 and MINIMA-RoMa; mean{\color{sdgray}\,$\pm$\,SD} over three seeds; zero-shot: one run). On the synthetic worlds, land cover and height use the worlds' targets; for registration, the FLUX rows use the training pairs and map-geometry correspondences of L0. In the translation controls, FLUX translations of real optical images replace the real SAR under the same labels or pairs (land cover mapping: 1,837 BRIGHT tiles; registration: four training scenes).}
\label{tab:s-flux-all}
\centering\small
\setlength{\tabcolsep}{4pt}
\begin{tabular}{@{}llcccc@{}}
\toprule
& & Land cover mapping & \multicolumn{2}{c}{Building MAE (m)\,$\downarrow$} & Registration (six scenes) \\
\cmidrule(lr){3-3}\cmidrule(lr){4-5}\cmidrule(l){6-6}
Training images & Optical input of FLUX & \miou\,$\uparrow$ & SN6 & AMS & \Sten\,$\uparrow$ \\
\midrule
\multicolumn{6}{@{}l}{\emph{Training on the synthetic worlds}} \\
\quad SynSAR L0 &  & 12.0{\scriptsize\color{sdgray}$\,\pm$0.3} & 11.33{\scriptsize\color{sdgray}$\,\pm$0.13} & 10.93{\scriptsize\color{sdgray}$\,\pm$0.08} & 66.2{\scriptsize\color{sdgray}$\,\pm$0.2} \\
\quad SynSAR L1 &  & 17.5{\scriptsize\color{sdgray}$\,\pm$0.6} & 10.64{\scriptsize\color{sdgray}$\,\pm$0.09} & 10.13{\scriptsize\color{sdgray}$\,\pm$0.09} & 68.9{\scriptsize\color{sdgray}$\,\pm$0.5} \\
\quad SynSAR L4 &  & 23.4{\scriptsize\color{sdgray}$\,\pm$0.2} & 7.73{\scriptsize\color{sdgray}$\,\pm$0.21} & 7.50{\scriptsize\color{sdgray}$\,\pm$0.09} & 69.6{\scriptsize\color{sdgray}$\,\pm$0.4} \\
\quad FLUX & generated (Changen2) & 19.4{\scriptsize\color{sdgray}$\,\pm$0.4} & 10.06{\scriptsize\color{sdgray}$\,\pm$0.07} & 9.34{\scriptsize\color{sdgray}$\,\pm$0.10} & 50.0{\scriptsize\color{sdgray}$\,\pm$2.2} \\
\quad FLUX & rendered & 18.1{\scriptsize\color{sdgray}$\,\pm$0.2} & 10.75{\scriptsize\color{sdgray}$\,\pm$0.07} & 10.10{\scriptsize\color{sdgray}$\,\pm$0.01} & 63.2{\scriptsize\color{sdgray}$\,\pm$1.3} \\
\addlinespace[2pt]
\multicolumn{6}{@{}l}{\emph{Training on real scenes: translation controls}} \\
\quad Real SAR &  & 31.4{\scriptsize\color{sdgray}$\,\pm$0.6} & --- & --- & 65.5{\scriptsize\color{sdgray}$\,\pm$1.4} \\
\quad FLUX & real & 21.7{\scriptsize\color{sdgray}$\,\pm$0.2} & --- & --- & 24.8{\scriptsize\color{sdgray}$\,\pm$4.7} \\
\addlinespace[2pt]
\multicolumn{6}{@{}l}{\emph{No training}} \\
\quad Zero-shot &  & --- & --- & --- & 51.7 \\
\bottomrule
\end{tabular}

\end{table*}

\begin{table}[t]
\caption{\textbf{Per-class IoU on OEM-SAR} for the FLUX-based pipeline and its references (mean of three seeds). Top: training on the synthetic worlds; bottom: BRIGHT tiles with their real SAR or with FLUX translations. In parentheses: the optical input of FLUX.}
\label{tab:s-flux-classes}
\centering\footnotesize
\setlength{\tabcolsep}{2.2pt}
\begin{tabular}{@{}lcccccccc@{}}
\toprule
& \multicolumn{8}{c}{IoU per class\,$\uparrow$} \\
\cmidrule(l){2-9}
Training images & Bare & Range & Devel. & Road & Tree & Water & Agri. & Bldg. \\
\midrule
SynSAR L0 & 0.0 & 8.8 & 17.6 & 19.4 & 11.2 & 11.7 & 14.3 & 13.0 \\
SynSAR L4 & 0.0 & 16.4 & 20.5 & 20.9 & 38.0 & 28.5 & 29.0 & 33.8 \\
FLUX (generated) & 0.2 & 14.6 & 17.0 & 17.9 & 25.9 & 28.7 & 17.0 & 33.7 \\
FLUX (rendered) & 0.8 & 15.3 & 17.6 & 18.4 & 27.4 & 26.0 & 13.5 & 25.8 \\
\midrule
Real SAR & 0.3 & 24.9 & 20.3 & 19.6 & 41.0 & 59.4 & 43.1 & 42.3 \\
FLUX (real) & 0.0 & 19.0 & 18.6 & 17.0 & 23.6 & 24.4 & 38.2 & 32.8 \\
\bottomrule
\end{tabular}

\end{table}

\begin{figure*}[t]
\centering
\includegraphics[width=\textwidth]{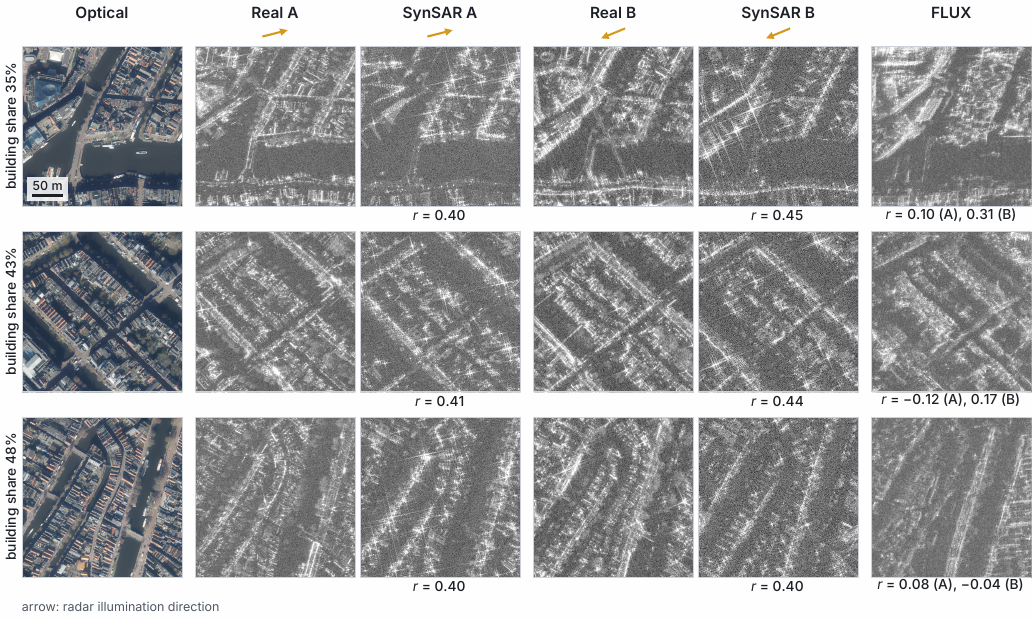}
\caption{\textbf{Digital-twin renderings compared with observed SAR.} Three of the 60 Amsterdam tiles, selected near the 25th, 50th, and 75th percentiles of building share; FLUX translates the orthophoto~\cite{beeldmateriaal_ortho} once. Images are shown at 0.5\,m; each SAR image is in dB relative to its median ($\pm$15\,dB range). Numbers: Pearson $r$ with the real image of the same acquisition on 2\,m blocks (FLUX: with A and with B). The central part of the middle tile appears in Fig.~6 of the main paper. Orthophotos: Beeldmateriaal.nl (CC BY 4.0).}
\label{fig:s-twin}
\end{figure*}

\begin{figure}[t]
\centering
\includegraphics[width=\linewidth]{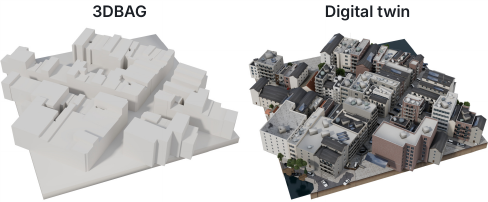}
\caption{\textbf{From 3DBAG to a digital twin.} A 128\,m block of the middle tile of \cref{fig:s-twin}. Left: 3DBAG~\cite{peters2022bag3d}, one flat-topped prism per building. Right: the twin that \synsar renders, with procedurally generated roofs, facades, and smaller objects and with trees from AHN4~\cite{ahn_ahn4}.}
\label{fig:s-twinbuild}
\end{figure}

\begin{table*}[t]
\caption{\textbf{Agreement with observed SAR on 60 Amsterdam digital twins.} Means over the tiles of the correlations defined in \cref{sec:s-twin} (2\,m blocks), with 95\% bootstrap intervals over tiles {\color{sdgray}[in brackets]}. A and B are two Umbra acquisitions with opposite look directions; columns give the acquisition of the real image, and FLUX translates the orthophoto of a tile once for both.}
\label{tab:s-twin}
\centering\small
\setlength{\tabcolsep}{5pt}
\begin{tabular}{@{}lcccccc@{}}
\toprule
& \multicolumn{3}{c}{Acquisition A} & \multicolumn{3}{c}{Acquisition B} \\
\cmidrule(lr){2-4}\cmidrule(l){5-7}
& SynSAR & FLUX & SynSAR $-$ FLUX & SynSAR & FLUX & SynSAR $-$ FLUX \\
\midrule
\multicolumn{7}{@{}l}{\emph{Agreement with the real image of the same acquisition}} \\
\quad Pearson $r$\,$\uparrow$ & 0.37 & $-$0.03 & $+$0.40\,{\color{sdgray}[0.37, 0.43]} & 0.37 & 0.06 & $+$0.31\,{\color{sdgray}[0.28, 0.34]} \\
\quad Building-zone $r$\,$\uparrow$ & 0.38 & $-$0.06 & $+$0.44\,{\color{sdgray}[0.41, 0.47]} & 0.39 & 0.05 & $+$0.34\,{\color{sdgray}[0.31, 0.37]} \\
\quad Ground-zone $r$\,$\uparrow$ & 0.25 & $-$0.02 & $+$0.27\,{\color{sdgray}[0.23, 0.31]} & 0.26 & 0.06 & $+$0.20\,{\color{sdgray}[0.17, 0.24]} \\
\bottomrule
\end{tabular}

\vspace{13.8pt}

\begin{tabular}[t]{@{}lcc@{}}
\toprule
& Acquisition A & Acquisition B \\
\midrule
\multicolumn{3}{@{}l}{\emph{SynSAR rendered with the geometry of the other acquisition, Pearson $r$}} \\
\quad Other geometry & $-$0.12 & $-$0.13 \\
\quad Matching $-$ other geometry & $+$0.49\,{\color{sdgray}[0.45, 0.53]} & $+$0.50\,{\color{sdgray}[0.46, 0.53]} \\
\bottomrule
\end{tabular}\hspace{16pt}\begin{tabular}[t]{@{}lc@{}}
\toprule
& A against B \\
\midrule
\multicolumn{2}{@{}l}{\emph{Between the two acquisitions of a tile, Pearson $r$}} \\
\quad Real images & $-$0.07\,{\color{sdgray}[$-$0.10, $-$0.04]} \\
\quad SynSAR renderings & $-$0.09\,{\color{sdgray}[$-$0.11, $-$0.06]} \\
\bottomrule
\end{tabular}

\end{table*}

\begin{table}[t]
\caption{\textbf{Agreement with observed SAR across simulation levels} on the 60 digital twins. Mean Pearson $r$ over the tiles (2\,m blocks, real image of the same acquisition) at the levels of \cref{sec:s-levels}, and the paired difference to L4 with its 95\% bootstrap interval over tiles {\color{sdgray}[in brackets]}. L0, L1, and GS are constructed from the Ld image.}
\label{tab:s-twin-levels}
\centering\footnotesize
\setlength{\tabcolsep}{4pt}
\begin{tabular}{@{}lcccc@{}}
\toprule
& \multicolumn{2}{c}{Acquisition A} & \multicolumn{2}{c}{Acquisition B} \\
\cmidrule(lr){2-3}\cmidrule(l){4-5}
Level & $r$\,$\uparrow$ & L4 $-$ level & $r$\,$\uparrow$ & L4 $-$ level \\
\midrule
L0 & 0.02 & $+$0.35\,{\color{sdgray}[0.32, 0.38]} & 0.04 & $+$0.33\,{\color{sdgray}[0.30, 0.35]} \\
L1 & 0.13 & $+$0.23\,{\color{sdgray}[0.21, 0.25]} & 0.14 & $+$0.23\,{\color{sdgray}[0.21, 0.24]} \\
GS & 0.13 & $+$0.24\,{\color{sdgray}[0.22, 0.26]} & 0.13 & $+$0.23\,{\color{sdgray}[0.21, 0.25]} \\
Ld & 0.20 & $+$0.17\,{\color{sdgray}[0.15, 0.19]} & 0.20 & $+$0.17\,{\color{sdgray}[0.15, 0.18]} \\
L2 & 0.20 & $+$0.17\,{\color{sdgray}[0.15, 0.18]} & 0.20 & $+$0.16\,{\color{sdgray}[0.15, 0.17]} \\
L3 & 0.34 & $+$0.03\,{\color{sdgray}[0.02, 0.03]} & 0.34 & $+$0.03\,{\color{sdgray}[0.02, 0.03]} \\
L4 & 0.37 &  & 0.37 &  \\
\bottomrule
\end{tabular}

\end{table}

The differences between L3 and L4 are small in all tasks: below 0.1 \miou, at most 0.05\,m in building MAE, and 0.5 \Sten (\cref{tab:s-mech-all}).
Together, under the evaluated settings, the benefits of the tested imaging mechanisms vary by task: the traced scattering levels improve dense prediction, whereas the registration change is small.

\section{Physics-based simulation vs.\ learned optical-to-SAR translation}
\label{sec:s-cmp}

This section supplements Sec.~5.6 of the main paper, which evaluates FLUX-based optical-to-SAR translation as an alternative pipeline for generating synthetic training data, with additional results and details.

\subsection{Translation baseline and comparison protocol}
\label{sec:s-flux}

The two pipelines start from the same synthetic worlds and use the same downstream training settings, but differ in the inputs of image generation and, for registration, in the geometric supervision.
The comparison therefore assesses the downstream transfer of complete pipelines.

\paragraph{Translation model.}
FLUX.1-dev~\cite{bfl2024flux} is adapted to RGB-to-SAR translation: the transformer is conditioned on the optical image, whose VAE latent is concatenated with the noisy SAR latent, together with a fixed text prompt; rank-128 LoRA~\cite{hu2022lora} adapters on its attention, feed-forward, and projection layers and the widened input projection are trained with the flow-matching objective, while all remaining pretrained weights are kept frozen.
The adaptation data are 4,543 paired optical and SAR images of OEM-SAR, all outside its 490-image test set.
Training uses AdamW with a batch size of 32, for 200 epochs at 512 pixels with a learning rate of $10^{-4}$ and then 6 epochs at 1,024 pixels with a learning rate of $2\times10^{-5}$.
The model has no input that describes the acquisition geometry.

\paragraph{Land cover mapping and building height estimation.}
For each small world of \cref{sec:s-levels}, Changen2~\cite{zheng2025changen2} (RSDiT-B/2 with its OpenEarthMap~\cite{xia2023openearthmap} weights, 50 DDIM~\cite{song2021ddim} steps) generates an optical-like RGB image from the land cover map of the world.
FLUX translates it into a SAR-like image (30 Euler steps, at the ground scale of the world on a mirror-padded 1,024-pixel canvas), which is converted to amplitude and normalized by its median (\cref{fig:s-flux}).
The land cover and height targets are those of the world and are thus shared with the \synsar levels, as are the training settings.
The generated image follows the land cover layout, but its individual objects need not match those of the 3D world.
\Cref{tab:s-flux-all} also reports both tasks with the rendered optical image of each world as the input of FLUX; the translation is placed on the map grid, and the targets and training settings are unchanged.

\paragraph{SAR--optical registration.}
FLUX translates the rendered optical image of each of the 191 large worlds.
The translation is placed on the map grid, and the training pairs and supervision are those of L0, i.e., map-geometry correspondences, whereas the \synsar images at L1--L4 are supervised with radar-geometry correspondences (\cref{sec:s-levels}).
\Cref{tab:s-flux-all} also reports registration with generated optical input, where the translation shares only the land cover layout with the rendered optical image of the pair.

\paragraph{Real optical input.}
As controls, FLUX translates real optical images, and the translations replace the real SAR in training: the pre-event optical images of the 1,837 BRIGHT tiles for land cover mapping and the optical images of the four real training scenes for registration, with labels, pairs, and training settings unchanged (\cref{sec:s-training}).
For the digital twins of \cref{sec:s-twin}, FLUX translates the aerial orthophoto of each tile.

\subsection{Real-world transfer of the two pipelines}
\label{sec:s-flux-transfer}

Under the evaluated setup, training on \synsar (L4) yields higher overall transfer than training on the FLUX-based pipeline in all three tasks (\cref{tab:s-flux-all}): 23.4 against 19.4 \miou in land cover mapping (difference 4.0 [3.2, 4.8]), a building MAE of 7.73 against 10.06\,m on SN6 and 7.50 against 9.34\,m on AMS (2.33 [1.87, 2.79] and 1.84 [1.62, 2.06]\,m), and 69.6 against 63.2 \Sten in registration (6.4 [3.6, 9.2]).
As described in \cref{sec:s-flux}, FLUX takes generated optical input for land cover mapping and height and rendered optical input for registration; these are differences between complete pipelines, and for registration they include the difference in correspondence supervision.

Relative to the map-geometry control L0, FLUX gains 7.4 \miou [6.5, 8.2] in land cover mapping and lowers building MAE by 1.27\,m [1.01, 1.54] on SN6 and 1.59\,m [1.37, 1.80] on AMS, but is 3.1 \Sten lower in registration [$-$6.1, $-$0.1].
With the optical inputs exchanged, FLUX reaches 18.1 \miou and building MAEs of 10.75 and 10.10\,m with rendered input ($-1.3$ [$-$2.0, $-$0.5]; 0.69 [0.54, 0.84] and 0.76 [0.50, 1.01]\,m higher), and 50.0 \Sten in registration with generated input ($-13.1$ [$-$17.6, $-$8.7]).
With real optical images as input, the models reach 21.7 \miou and 24.8 \Sten, against 31.4 and 65.5 with the real SAR of the same tiles and scenes ($-9.7$ [$-$10.9, $-$8.4] and $-40.7$ [$-$51.3, $-$30.1]).
Per-class IoU is listed in \cref{tab:s-flux-classes}.

\subsection{Agreement with observed SAR}
\label{sec:s-twin}

We construct digital twins of 60 Amsterdam tiles ($256\times256$\,m) observed by two Umbra acquisitions~\cite{umbra_opendata}, A and B, that look from opposite directions at similar incidence (about 46$^\circ$).
The twins take their buildings from 3DBAG~\cite{peters2022bag3d}, their land cover from the Dutch large-scale topographic map~\cite{bgt_pdok}, and their trees from AHN4~\cite{ahn_ahn4}, on flat terrain with the generator's default materials.
From 3DBAG, each building takes its footprint, a single height, and a flat-or-slanted roof label; roof geometry, facades, and smaller objects are sampled procedurally with the same rules as in the synthetic worlds (\cref{fig:s-twinbuild}).
\synsar renders each twin under the geometry of each acquisition; the renderings are part of the release.
The evaluated FLUX model translates the aerial orthophoto of the tile~\cite{beeldmateriaal_ortho} once; having no acquisition-geometry input, it yields a single image, which is compared with both acquisitions (\cref{fig:s-twin}).

Each image is compared with the real image of an acquisition on 2\,m blocks.
For an image with amplitude $a$, the intensity is averaged over each block $b$, converted to dB, centered on the median over the set $V$ of blocks valid in both images, and clipped:
\begin{align*}
d_b&=10\log_{10}\Bigl(\frac{1}{|b|}\sum_{p\in b}a_p^2\Bigr),\\
\tilde d_b&=\operatorname{clip}\bigl(d_b-\operatorname*{median}_{b'\in V}d_{b'},\,-15,\,15\bigr).
\end{align*}
The Pearson correlation $r$ between the resulting values of a generated and a real image is computed over $V$; the building-zone and ground-zone $r$ restrict $V$ to the blocks inside and outside the building zone, which extends each building footprint toward the sensor by its layover length $h\cot\theta$ and away from it by its shadow length $h\tan\theta$ (height $h$, incidence angle $\theta$ of the acquisition).
The correlations are computed per tile; \cref{tab:s-twin} reports their means over the 60 tiles, with 95\% percentile bootstrap intervals over tiles.

Under the matching acquisition geometry, the \synsar renderings have a mean Pearson correlation of 0.37 with the real image, for each acquisition.
Rendered with the geometry of the other acquisition, the same twins give $-0.12$ and $-0.13$, and the FLUX translations give $-0.03$ and 0.06.
The two real images of a tile have a mean correlation of $-0.07$ with each other, and the two renderings of its twin $-0.09$.
The correlation of the matching rendering is higher in the building zones (0.38 and 0.39) than in the ground zones (0.25 and 0.26).

Formed at the lower simulation levels of \cref{sec:s-levels}, the same twins give a mean correlation of 0.02 and 0.04 at L0, 0.13 and 0.14 at L1, 0.20 at L2, and 0.34 at L3 for the two acquisitions, against 0.37 at L4 (\cref{tab:s-twin-levels}).
The difference between L4 and L3 is 0.03 [0.02, 0.03] for each acquisition.

These results show moderate acquisition-dependent spatial agreement for the evaluated digital twins.
\synsar uses 3D scene geometry with approximate materials, whereas the evaluated FLUX model translates real orthophotos without acquisition-geometry conditioning.
The comparison covers these pipelines on 60 tiles from one city.

\section{Scene and observation diversity}
\label{sec:s-div}

This section supplements Sec.~3.1 of the main paper, which notes that a world can be imaged again, with an experiment on repeated observations.

\subsection{Repeated observations and equal-budget setup}
\label{sec:s-div-protocol}

\begin{figure}[t]
\centering
\includegraphics[width=0.82\linewidth]{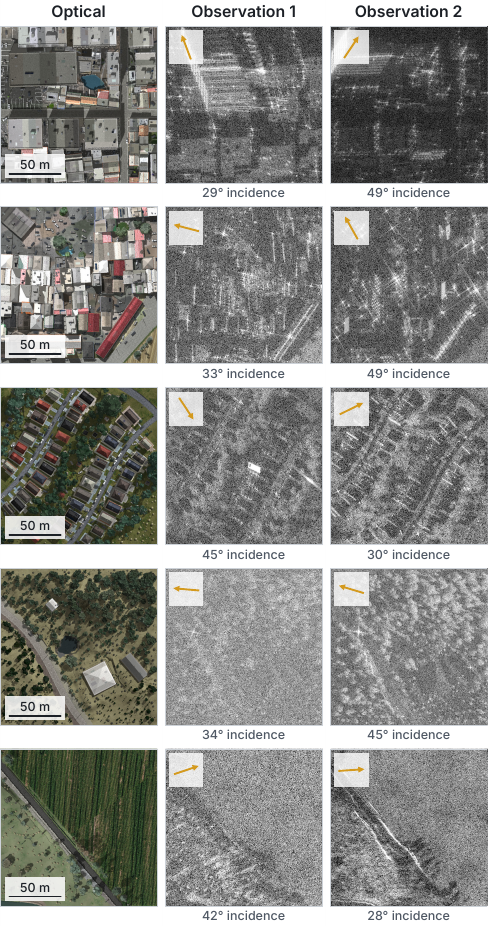}
\caption{\textbf{One world under two observations.} Five worlds of set B, from dense built-up to natural, each with its optical image and its two SAR observations. Arrows point along ground range, away from the radar. Each SAR image is stretched between its own 1st and 99.6th percentiles in dB.}
\label{fig:s-twoobs}
\end{figure}

\begin{table}[!t]
\caption{\textbf{Scene and observation diversity at equal image budgets} (DINOv3 trained on SynSAR only; means of three seeds). B: 1,800 worlds $\times$ 2 observations; C: 3,600 worlds $\times$ 1 observation. Land cover mapping on OEM-SAR; height: building MAE, overall and by true building height, and bias (mean signed error). B$-$C with its 95\% Welch interval {\color{sdgray}[in brackets]}.}
\label{tab:s-obs}
\centering\footnotesize
\setlength{\tabcolsep}{4pt}\renewcommand{\arraystretch}{0.92}
\begin{tabular}{@{}llccc@{}}
\toprule
Test set & Readout & B & C & B $-$ C \\
\midrule
OEM-SAR & \miou\,$\uparrow$ & 21.9 & 22.5 & $-$0.5\,{\color{sdgray}[$-$1.8, 0.7]} \\
 & Bare IoU\,$\uparrow$ & 0.1 & 0.1 & 0.0\,{\color{sdgray}[$-$0.2, 0.3]} \\
 & Range IoU\,$\uparrow$ & 14.1 & 15.3 & $-$1.2\,{\color{sdgray}[$-$3.1, 0.8]} \\
 & Devel. IoU\,$\uparrow$ & 21.3 & 21.4 & $-$0.1\,{\color{sdgray}[$-$0.8, 0.7]} \\
 & Road IoU\,$\uparrow$ & 19.4 & 19.8 & $-$0.4\,{\color{sdgray}[$-$1.9, 1.1]} \\
 & Tree IoU\,$\uparrow$ & 33.6 & 35.1 & $-$1.5\,{\color{sdgray}[$-$4.3, 1.2]} \\
 & Water IoU\,$\uparrow$ & 29.6 & 31.1 & $-$1.5\,{\color{sdgray}[$-$7.3, 4.2]} \\
 & Agri. IoU\,$\uparrow$ & 29.0 & 27.6 & $+$1.4\,{\color{sdgray}[$-$6.5, 9.2]} \\
 & Bldg. IoU\,$\uparrow$ & 28.5 & 29.4 & $-$0.9\,{\color{sdgray}[$-$1.8, 0.01]} \\
\addlinespace[3pt]
SN6 & MAE (m)\,$\downarrow$ & 7.75 & 8.04 & $-$0.29\,{\color{sdgray}[$-$0.61, 0.03]} \\
 & MAE, 0--3\,m\,$\downarrow$ & 2.82 & 2.65 & $+$0.17\,{\color{sdgray}[0.05, 0.29]} \\
 & MAE, 3--10\,m\,$\downarrow$ & 3.88 & 4.05 & $-$0.17\,{\color{sdgray}[$-$0.37, 0.03]} \\
 & MAE, 10--20\,m\,$\downarrow$ & 7.50 & 8.04 & $-$0.54\,{\color{sdgray}[$-$0.89, $-$0.19]} \\
 & MAE, 20--40\,m\,$\downarrow$ & 17.67 & 17.80 & $-$0.13\,{\color{sdgray}[$-$0.94, 0.68]} \\
 & MAE, $>$40\,m\,$\downarrow$ & 59.43 & 59.88 & $-$0.45\,{\color{sdgray}[$-$2.18, 1.28]} \\
 & Bias (m) & $-$6.80 & $-$7.27 & $+$0.47\,{\color{sdgray}[0.07, 0.86]} \\
\addlinespace[3pt]
AMS & MAE (m)\,$\downarrow$ & 7.53 & 7.96 & $-$0.43\,{\color{sdgray}[$-$0.51, $-$0.36]} \\
 & MAE, 0--3\,m\,$\downarrow$ & 4.96 & 4.68 & $+$0.28\,{\color{sdgray}[0.03, 0.54]} \\
 & MAE, 3--10\,m\,$\downarrow$ & 4.60 & 4.58 & $+$0.02\,{\color{sdgray}[$-$0.06, 0.11]} \\
 & MAE, 10--20\,m\,$\downarrow$ & 7.90 & 8.72 & $-$0.81\,{\color{sdgray}[$-$0.96, $-$0.67]} \\
 & MAE, 20--40\,m\,$\downarrow$ & 17.49 & 18.28 & $-$0.79\,{\color{sdgray}[$-$1.21, $-$0.37]} \\
 & MAE, $>$40\,m\,$\downarrow$ & 43.84 & 45.41 & $-$1.57\,{\color{sdgray}[$-$2.43, $-$0.72]} \\
 & Bias (m) & $-$5.67 & $-$6.40 & $+$0.72\,{\color{sdgray}[0.52, 0.92]} \\
\bottomrule
\end{tabular}

\end{table}

We compare two ways of allocating a fixed synthetic-image budget for land cover mapping and building height estimation: additional observations of existing worlds and additional independently generated worlds.
Set B contains two observations of each of 1,800 worlds; set C contains the same worlds and 1,800 further worlds, each observed once.
B and C thus have the same number of images (3,600).

The second observation of a world in B keeps the scene, its land cover labels, and its height targets and changes only the acquisition: another VV spotlight acquisition with the same pixel spacing and an incidence angle that differs from the first by 10--28$^\circ$ (\cref{fig:s-twoobs}).
These second observations are part of the release; the dataset otherwise contains one image per world (\cref{tab:s-dataset}).
The additional worlds of C are selected to match the second observations of B by world family, pixel spacing, and 3$^\circ$ incidence-angle bins, so the acquisition distributions of B and C are closely matched but not identical (mean incidence angles of 37.0$^\circ$ and 36.6$^\circ$).
Both sets train DINOv3 with the SynSAR-only settings of each task and three seeds (\cref{sec:s-training}) and are evaluated identically.

\subsection{Real-world transfer under equal image budgets}
\label{sec:s-div-results}

Under equal image budgets, B and C reach 21.9 and 22.5 \miou in land cover mapping (B$-$C $-0.5$ [$-$1.8, 0.7]; \cref{tab:s-obs}).
For building height estimation, B and C reach 7.53 and 7.96\,m on AMS ($-0.43$\,m [$-$0.51, $-$0.36]) and 7.75 and 8.04\,m on SN6 ($-0.29$\,m [$-$0.61, 0.03]).
\Cref{tab:s-obs} also lists the per-class IoU, the MAE by building height, and the bias.

\end{document}